\documentclass[10pt,twocolumn,letterpaper]{article}

\usepackage{iccv}
\usepackage{times}
\usepackage{epsfig}
\usepackage{graphicx}
\usepackage{amsmath}
\usepackage{amssymb}
\usepackage{multirow}
\usepackage{colortbl}
\usepackage{url}
\usepackage{caption}
\usepackage{enumitem}
\usepackage{mathrsfs}
\usepackage{algorithm}
\usepackage{algpseudocode}
\usepackage{dsfont}
\usepackage{color}
\usepackage{booktabs}
\usepackage{array}

\usepackage[breaklinks=true,bookmarks=false]{hyperref}

\newcommand{\tabincell}[2]{\begin{tabular}{@{}#1@{}}#2\end{tabular}}

\newcommand{\rf}[1]{\textcolor[rgb]{0.0,0.0,0.0}{#1}}

\iccvfinalcopy 

\def\iccvPaperID{5844} 

\ificcvfinal\fi

\begin{document}

\title{Where Is My Mirror?}

\author{
Xin Yang\textsuperscript{1,$\star$}, \ \
Haiyang Mei\textsuperscript{1,$\star$}, \ \
Ke Xu\textsuperscript{1,3}, \ \
Xiaopeng Wei\textsuperscript{1}, \ \
Baocai Yin\textsuperscript{1,2}, \ \
Rynson W.H. Lau\textsuperscript{3,$\dagger$}
\\
\textsuperscript{1} Dalian University of Technology, \ \
\textsuperscript{2} Peng Cheng Laboratory, \ \
\textsuperscript{3} City University of Hong Kong
\\
}
\maketitle
\renewcommand{\thefootnote}{}

\footnotetext{\textsuperscript{$\star$}Joint first authors. \textsuperscript{$\dagger$}Rynson Lau is the corresponding author, and he led this project. Project page: \url{https://mhaiyang.github.io/ICCV2019_MirrorNet/index}}

\ificcvfinal\thispagestyle{empty}\fi

\begin{abstract}
Mirrors are everywhere in our daily lives. Existing computer vision systems do not consider mirrors, and hence may get confused by the reflected content inside a mirror, resulting in a severe performance degradation. However, separating the real content outside a mirror from the reflected content inside it is non-trivial.  The key challenge is that mirrors typically reflect contents similar to their surroundings, making it very difficult to differentiate the two.
In this paper, we present a novel method to segment mirrors from an input image. To the best of our knowledge, this is the first work to address the mirror segmentation problem with a computational approach.
We make the following contributions.
First, we construct a large-scale mirror dataset that contains mirror images with corresponding manually annotated masks. This dataset covers a variety of daily life scenes, and will be made publicly available for future research.
Second, we propose a novel network, called MirrorNet, for mirror segmentation, by modeling both semantical and low-level color/texture discontinuities between the contents inside and outside of the mirrors.
%
%
Third, we conduct extensive experiments to evaluate the proposed method, and show that it outperforms the carefully chosen baselines from the state-of-the-art detection and segmentation methods.
\end{abstract}
\vspace{-0.2in}

\section{Introduction}
Mirrors are very common and important in our daily lives. The presence of mirrors may, however, severely degrades the performance of  existing computer vision tasks, \eg, by producing wrong depth predictions (Figure \ref{fig:example}(b)) or falsely detecting the reflected objects as real ones (Figure \ref{fig:example}(c)). Hence, it is essential to these systems to be able to detect and segment mirrors from the input images.

\def\wthr{0.3\linewidth}
\def\hthr{.6in}
\begin{figure}
\setlength{\tabcolsep}{.10pt}
  \centering
  \begin{tabular}{ccc}
    & \includegraphics[width=\wthr, height=2.5mm]{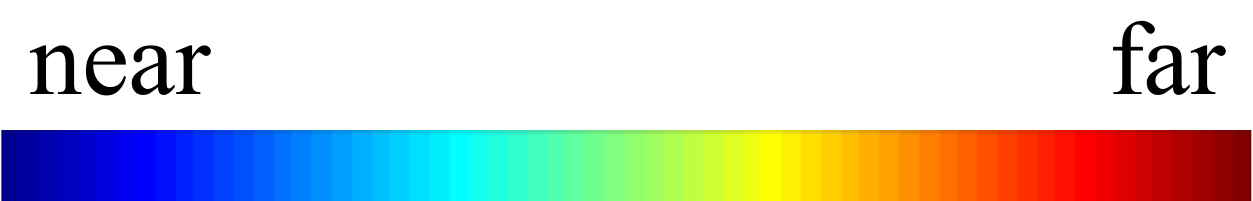}  & \\

    \includegraphics[width=\wthr, height=\hthr]{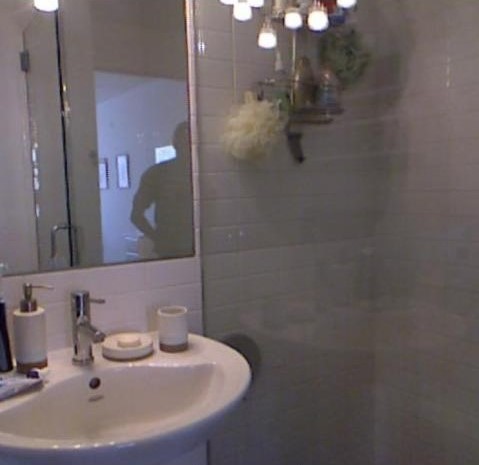}&
    \includegraphics[width=\wthr, height=\hthr]{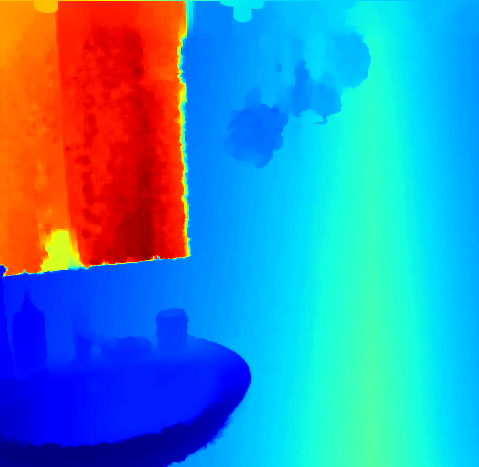}&
    \includegraphics[width=\wthr, height=\hthr]{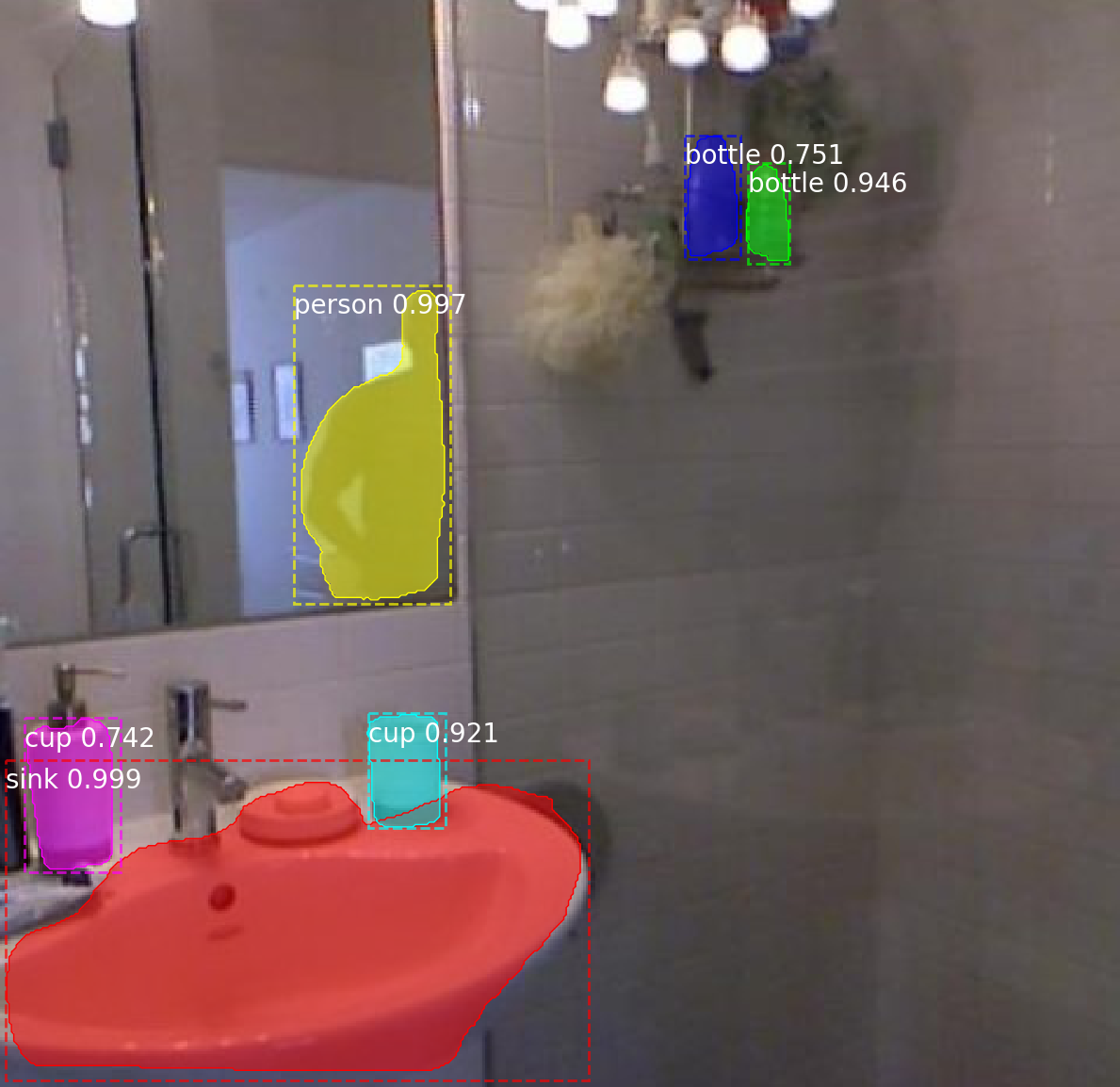}\\
    \includegraphics[width=\wthr,height=\hthr]{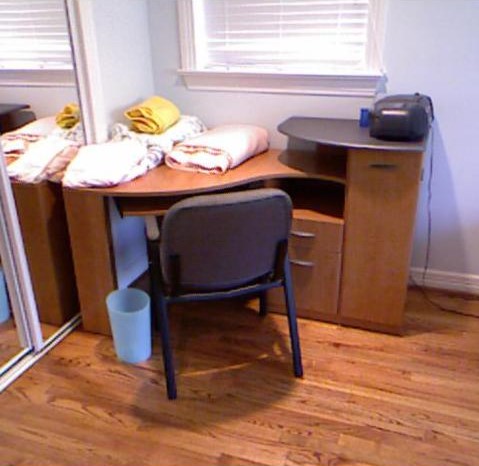}&
    \includegraphics[width=\wthr, height=\hthr]{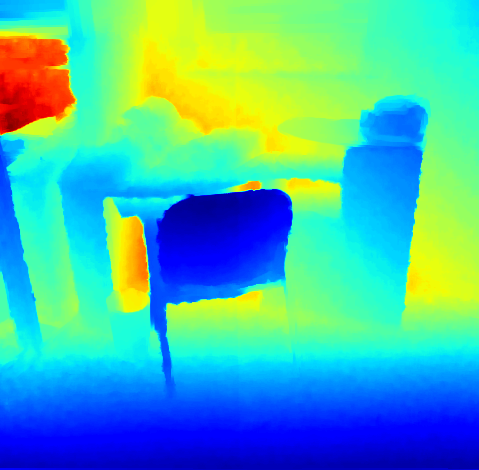}&
    \includegraphics[width=\wthr, height=\hthr]{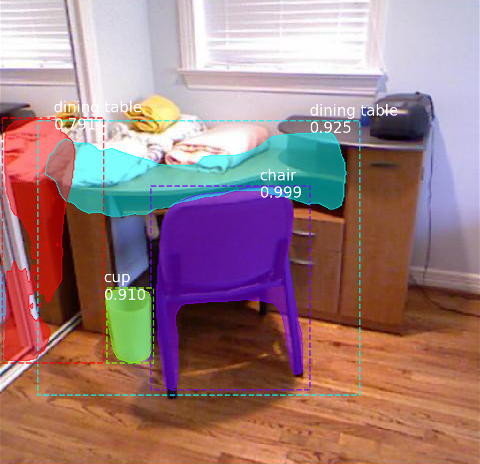} \\
    \footnotesize{(a) image} & \footnotesize{(b) depth} & \footnotesize{(c) segmentation} \\

    \includegraphics[width=\wthr, height=\hthr]{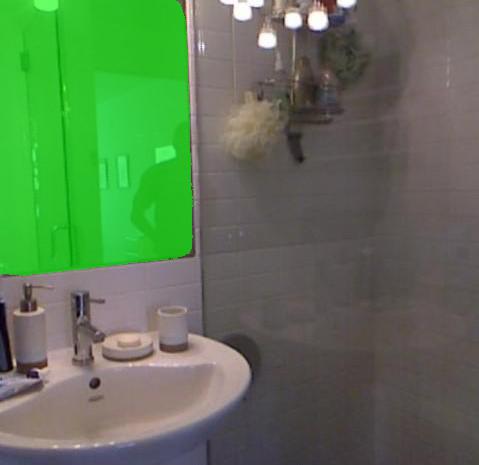}&
    \includegraphics[width=\wthr, height=\hthr]{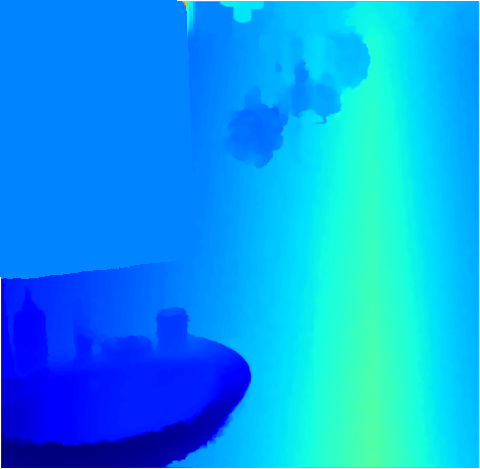}&
    \includegraphics[width=\wthr, height=\hthr]{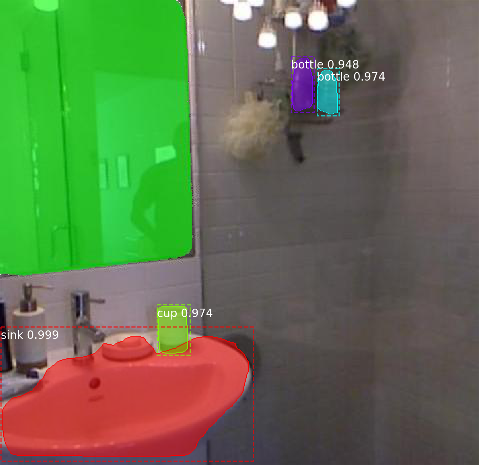}\\
    \includegraphics[width=\wthr, height=\hthr]{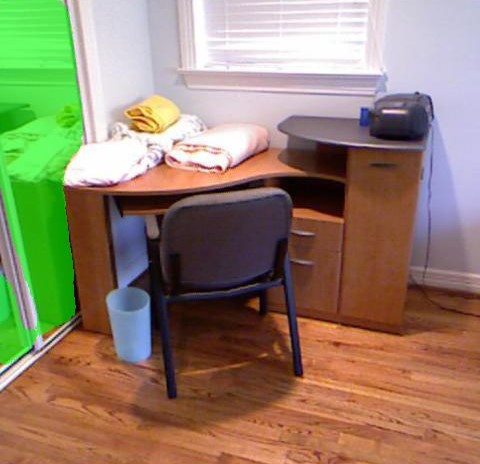}&
    \includegraphics[width=\wthr, height=\hthr]{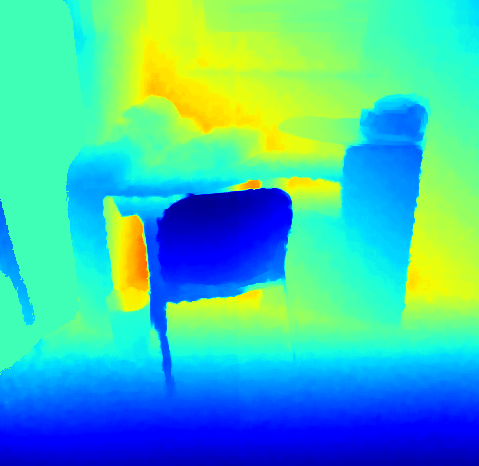}&
    \includegraphics[width=\wthr, height=\hthr]{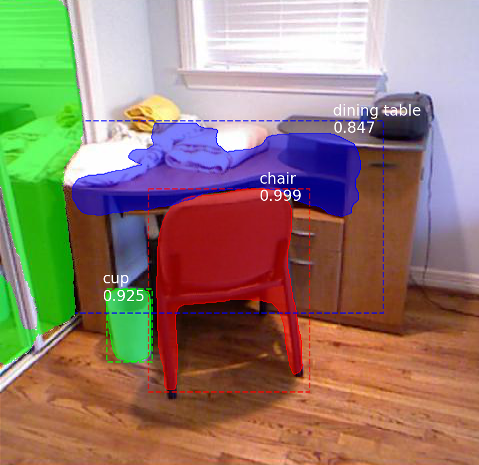}\\
   \footnotesize{(d) MirrorNet} & \footnotesize{(e) our depth} & \footnotesize{(f) our segmentation}\\
    \end{tabular}\vspace{-2mm}
  \caption{Problems with mirrors in existing vision tasks. In depth prediction, NYU-v2 dataset~\cite{silberman2012indoor} uses a Kinect to capture depth as ground truth. It wrongly predicts the depths of the reflected contents, instead of the mirror depths (b). In instance semantic segmentation, Mask RCNN~\cite{he2017mask} wrongly detects objects inside the mirrors (c). With MirrorNet, we first detect and mask out the mirrors (d). We then obtain the correct depths (e), by interpolating the depths from surrounding pixels of the mirrors, and segmentation maps (f).}
  \label{fig:example}\vspace{-5mm}
\end{figure}

Automatically segmenting mirrors from the background is extremely challenging, due to the fact that the contents reflected by the mirrors are very similar to those outside them (\ie, their surroundings). This makes them fundamentally different from other objects that have been addressed well by the state-of-the-art segmentation methods \cite{Zhao_2018_ECCV, he2017mask}. Meanwhile, as the contents reflected by the mirrors may not necessarily be salient, directly applying state-of-the-art saliency detection methods \cite{Chen_2018_ECCV, liu2018picanet} for detecting mirrors is also not appropriate.

In this work, we aim to address the mirror segmentation problem. We note that humans can generally detect the existence of mirrors well. To do this, we observe that humans typically try to identify content discontinuity at the mirror boundaries in order to differentiate if some content belong to the reflection of a mirror. Hence, a straightforward solution to this problem is to apply low-level features to detect mirror boundaries. Unfortunately, this may fail if an object partially appears in front of a mirror, \eg, the second example in Figure \ref{fig:example}. In this case, separating the reflection of the object from the object itself may not be straightforward. Instead, this discontinuity includes both low-level color/texture changes as well as high-level semantics. This observation inspires us to leverage the contextual contrasted information for mirror segmentation.

In this paper, we address the mirror segmentation problem in two ways. First, we have constructed a large-scale mirror segmentation dataset (MSD), which contains $4,018$ pairs of images with mirrors and their corresponding segmentation masks, covering a variety of daily life scenes. Second, we propose a novel network, called MirrorNet, with a Contextual Contrasted Feature Extraction (CCFE) module, to segment mirrors of different sizes, by learning the contextual contrast inside and outside of the mirrors.

We have the following main contributions:
\begin{itemize}
  \item We construct the first large-scale mirror dataset, which consists of $4,018$ images containing mirrors and their corresponding manually annotated mirror masks, taken from diverse daily life scenes.
  \item We propose a novel network that incorporates a novel contextual contrasted feature extraction module for mirror segmentation, by learning to model the contextual contrast inside and outside of the mirrors.
  \item Through extensive experiments, we show that the proposed network outperforms many baselines derived from state-of-the-art segmentation/detection methods.
\end{itemize}

\section{Related Work}
In this section, we briefly review state-of-the-art methods from relevant fields, including semantic/instance segmentation, saliency/shadow detection, as well as mirror detection works from the 3D community.

{\bf Semantic segmentation.} It aims to assign per-pixel predictions of object categories to the input image.
Based on the fully convolutional encoder-decoder structure~\cite{long2015fully}, state-of-the-art semantic segmentation approaches typically leverage multi-scale (level) context aggregation to learn discriminative features for recognizing the objects and delineating their boundaries.
Specifically, low-level encoder features are combined with their corresponding decoder features by feeding recorded pooling indices~\cite{Badrinarayanan2017SegNet} or concatenation~\cite{ronneberger2015u}. Dilated convolutions are used in~\cite{chen2015semantic,yu2015multi} to expand the receptive fields to compensate for the lost details in the encoder part. PSPNet~\cite{zhao2017pyramid} leverages pyramid pooling to obtain multi-scale representations in order to differentiate objects of similar appearances.
Zhang~\etal~\cite{Zhang_2018_ECCV} propose to fuse the low-/high-level features so as to take advantages of both high resolution spatial and rich semantic information in the encoder part.
Zhang~\etal~\cite{zhang2018context} propose to explicitly predict the objects in the scene and use this prediction to selectively highlight the semantic features.
Ding~\etal~\cite{ding2018context} propose to learn contextual contrasted features to boost the segmentation performance of small objects.

However, applying existing segmentation methods for mirror segmentation (\ie, treating mirrors as one of the object categories) cannot solve the fundamental problem of mirror segmentation, which is that the reflected content of a mirror
can also be segmented too. In this paper, we focus on the mirror segmentation problem and formulate it as a binary classification problem (\ie, mirror or non-mirror).

\def\wsample{0.095\linewidth}
\def\hsample{.73in}
\begin{figure*}[tbp]
\setlength{\tabcolsep}{1.5pt}
  \centering
  \begin{tabular}{cccccccccc}
    \includegraphics[width=\wsample, height=\hsample]{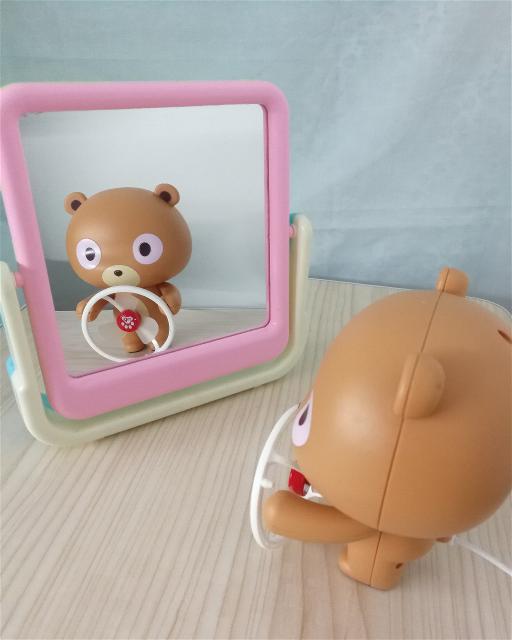}&
    \includegraphics[width=\wsample, height=\hsample]{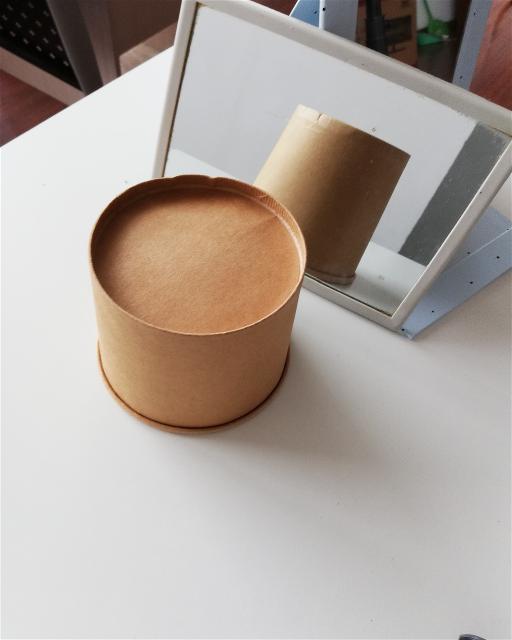}&
    \includegraphics[width=\wsample, height=\hsample]{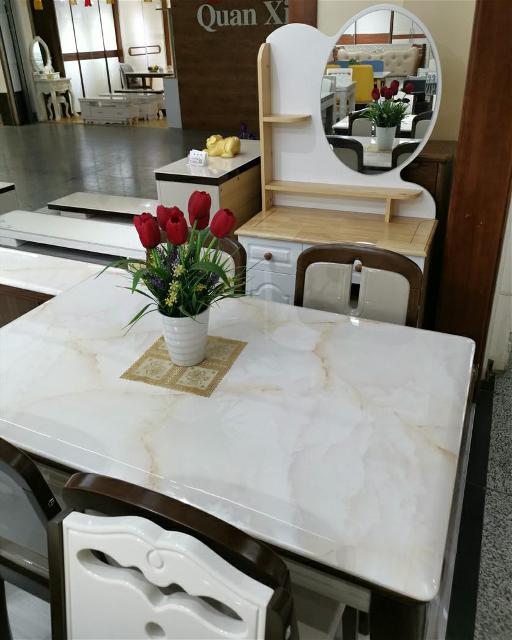}&
    \includegraphics[width=\wsample, height=\hsample]{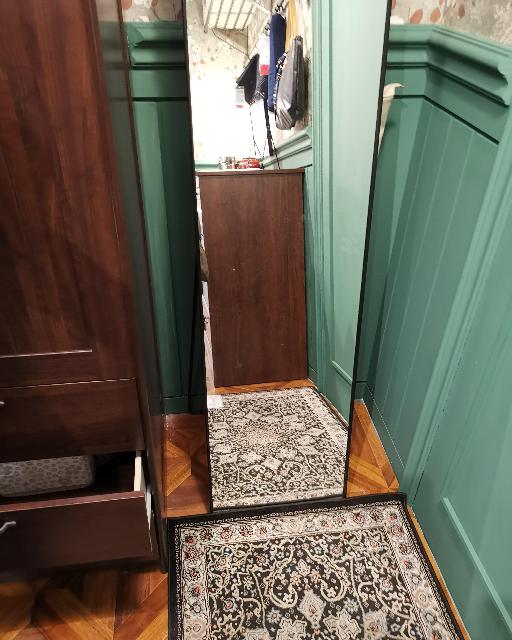}&
    \includegraphics[width=\wsample, height=\hsample]{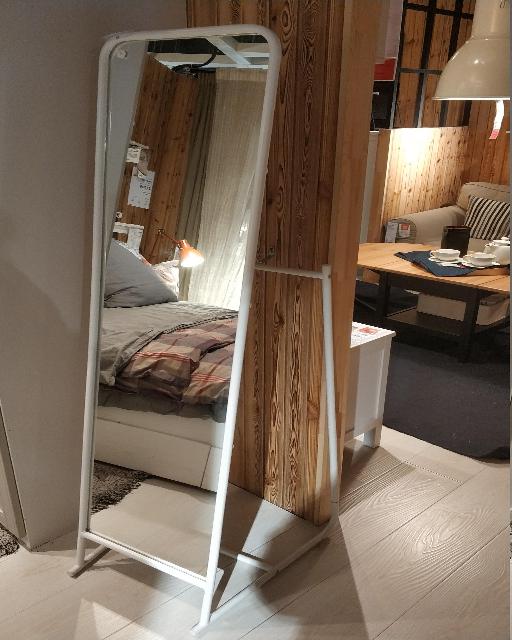}&
    \includegraphics[width=\wsample, height=\hsample]{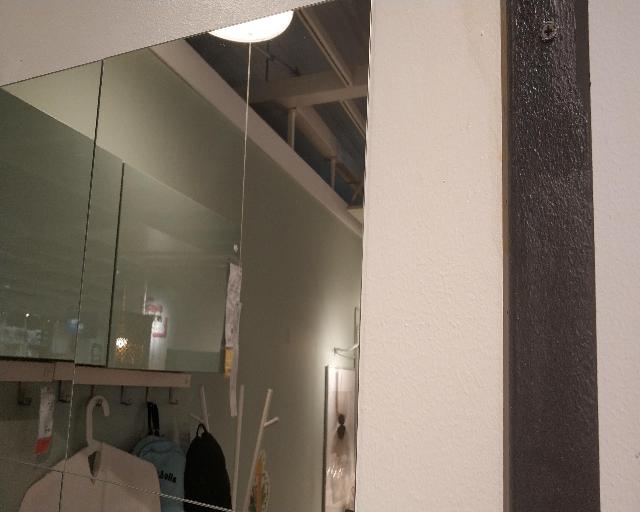}&
    \includegraphics[width=\wsample, height=\hsample]{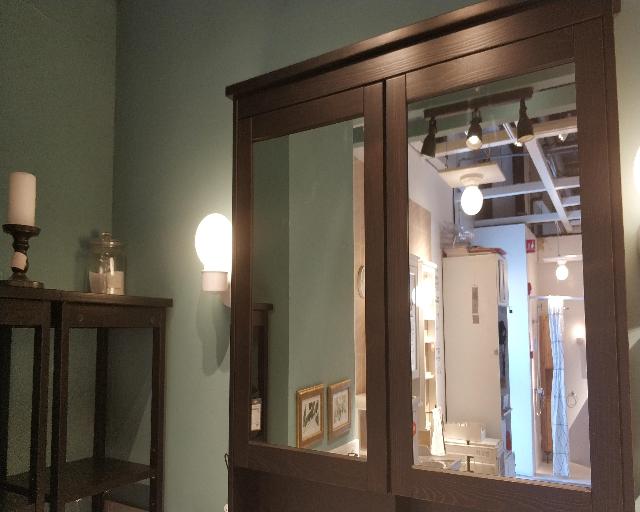}&
    \includegraphics[width=\wsample, height=\hsample]{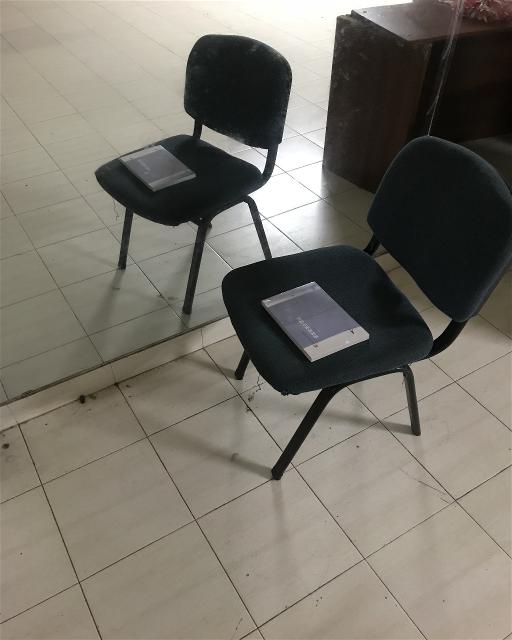}&
    \includegraphics[width=\wsample, height=\hsample]{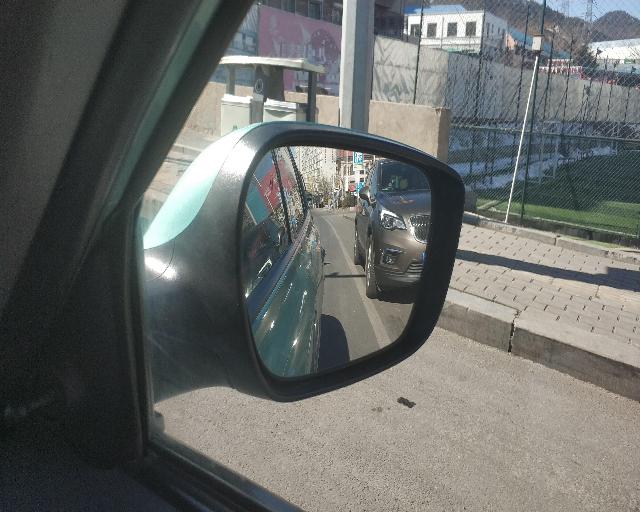}&
    \includegraphics[width=\wsample, height=\hsample]{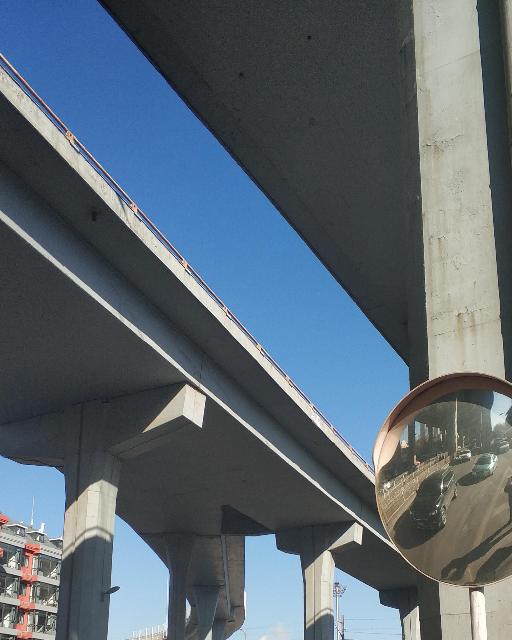} \\

    \includegraphics[width=\wsample, height=\hsample]{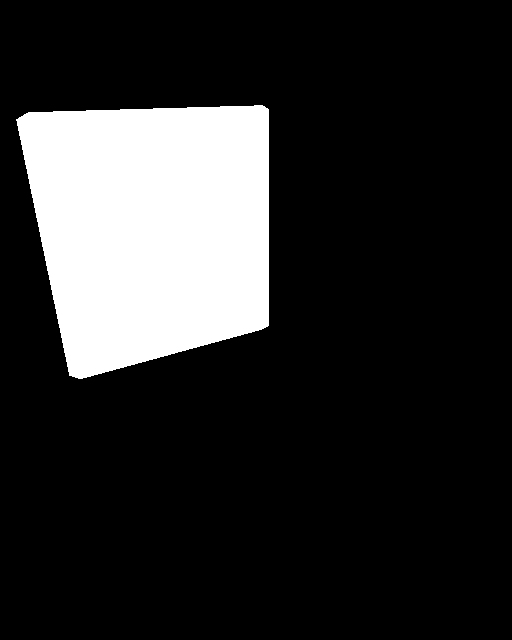}&
    \includegraphics[width=\wsample, height=\hsample]{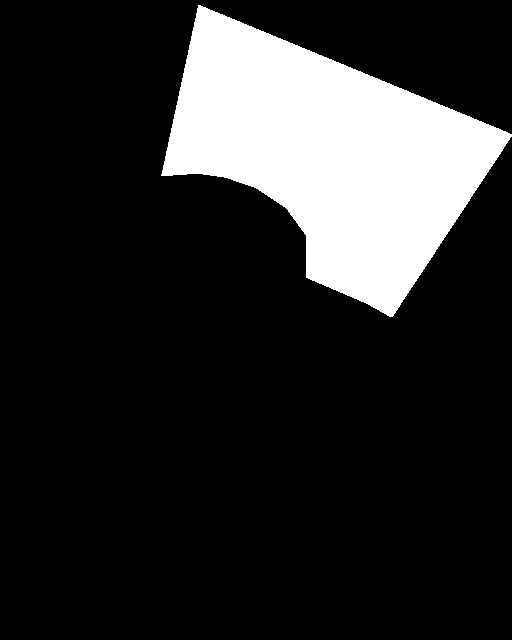}&
    \includegraphics[width=\wsample, height=\hsample]{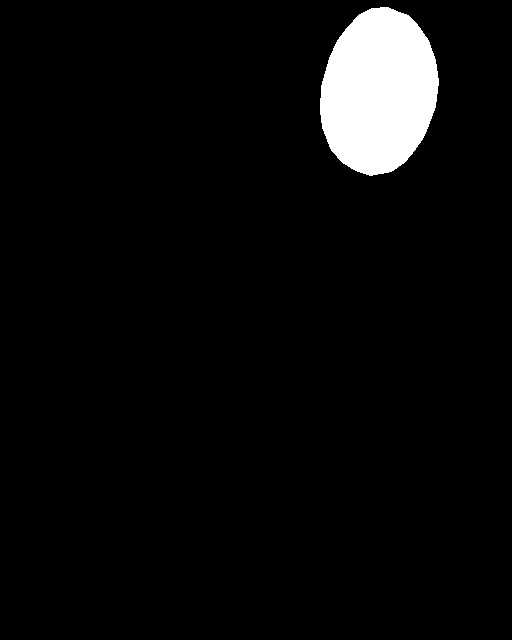}&
    \includegraphics[width=\wsample, height=\hsample]{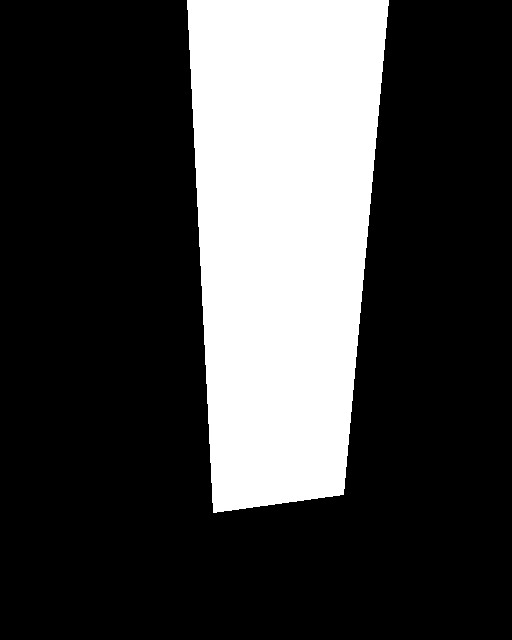}&
    \includegraphics[width=\wsample, height=\hsample]{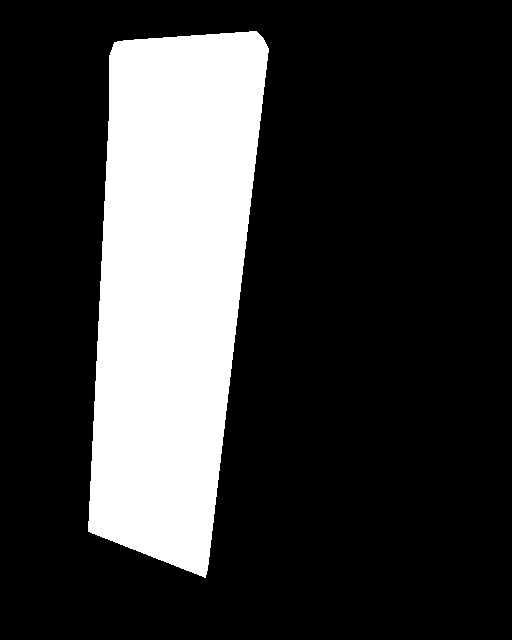}&
    \includegraphics[width=\wsample, height=\hsample]{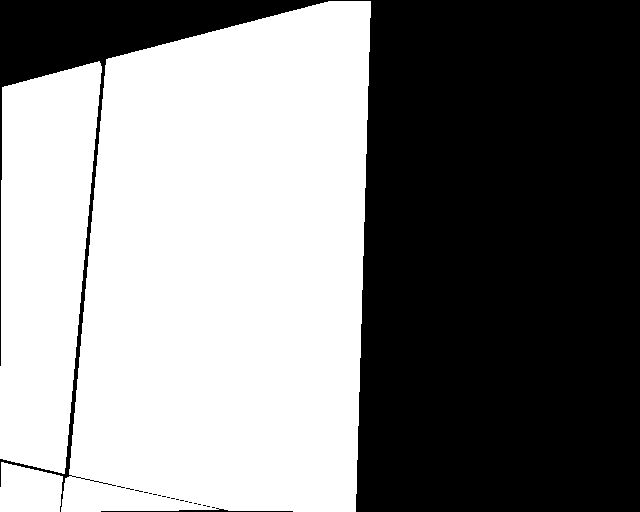}&
    \includegraphics[width=\wsample, height=\hsample]{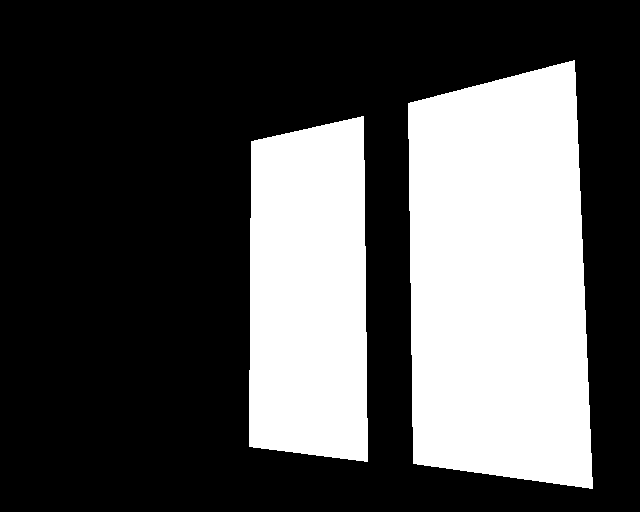}&
    \includegraphics[width=\wsample, height=\hsample]{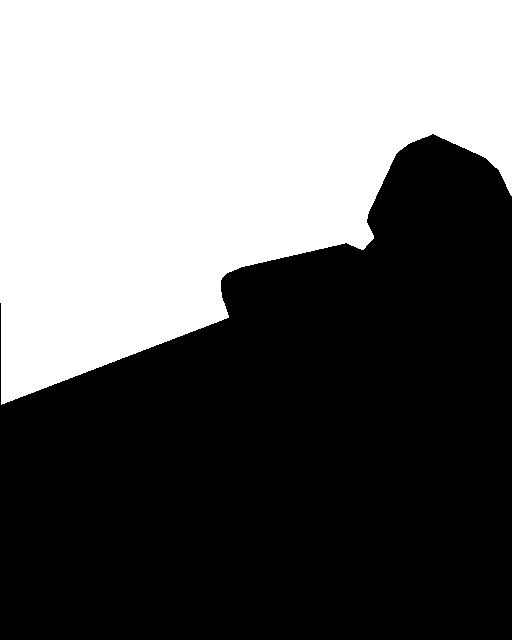}&
    \includegraphics[width=\wsample, height=\hsample]{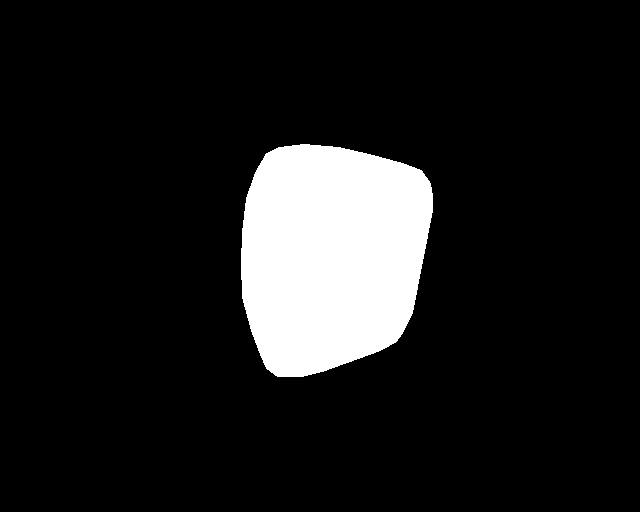}&
    \includegraphics[width=\wsample, height=\hsample]{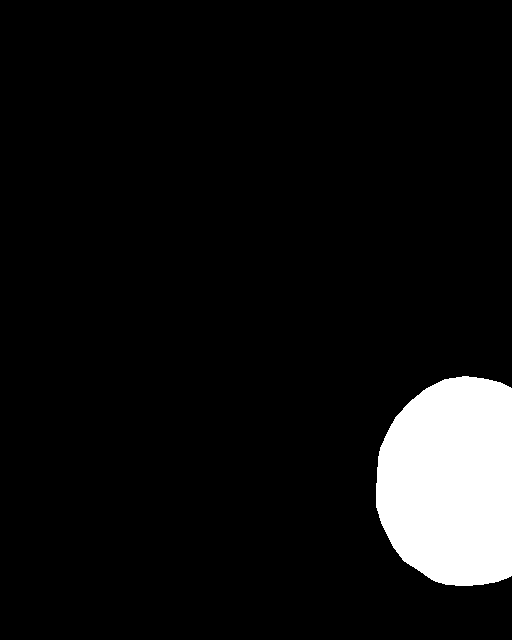} \\

    \vspace{-6mm}

    \end{tabular}
    \caption{Example mirror image/mask pairs in our mirror segmentation dataset (MSD). It shows that our MSD covers a variety of our daily life scenes that contain mirrors.}
    \label{fig:dataset}
  \vspace{-3.5mm}
\end{figure*}

{\bf Instance segmentation.} It aims to simultaneously recognize, localize and segment out objects while differentiating individual instances of the same category.
State-of-the-art detection based instance segmentation methods extend object detection methods, \eg, Faster-RCNN~\cite{ren2017faster} and FPN~\cite{lin2017feature}, to obtain instance maps. Mask RCNN~\cite{he2017mask} uses one additional branch to predict instance segmentation masks from the box predictions of Faster-RCNN~\cite{ren2017faster}. PANet~\cite{liu2018path} further proposes to add bottom-up paths to facilitate feature propagation in Mask RCNN \cite{he2017mask} and aggregates multi-level features for detection and segmentation. MaskLab \cite{Chen_2018_CVPR} adopts Faster-RCNN~\cite{ren2017faster} to locate objects and combines semantic segmentation with pixel-direction (to its instance center) prediction for instance segmentation. Another line of works first use a segmentation method to obtain per-pixel labels, and then a clustering method to group the pixels into instances, via depth estimation~\cite{Zhang_2015_ICCV}, spectral clustering~\cite{liang2018proposal}, and neural networks~\cite{wu2016bridging,Liu_2017_ICCV}.

Similar to semantic segmentation, instance segmentation methods cannot differentiate between the content inside a mirror and that of \rf{outside}. As a result, they would segment objects inside the mirror too.

{\bf Salient object detection (SOD).} It aims to identify the most conspicuous object(s) in an image. While conventional SOD methods rely on low-level hand-crafted features (\eg, color and contrast), deep learning based SOD methods consider either or both bottom-up and top-down saliency inferences. Wang~\etal~\cite{Wang2015Deep} propose to integrate local pixel-wise saliency estimation and global object proposal search for salient object detection. Multi-level feature aggregation from deep networks is also explored for detecting and refining the detection \cite{lee2016deep,Zhang2017Amulet,Hou_2017_CVPR}. Recent works apply attention mechanisms for learning global and local contexts~\cite{liu2018picanet} or learning foreground/background attention maps~\cite{Chen_2018_ECCV} to help detect salient objects and eliminate non-salient objects.

The content reflected by a mirror, however, may or may not be salient. Even though if it is salient, it is likely that only part of it is salient. Hence, applying existing SOD methods to detect mirrors may not address the mirror segmentation problem.

{\bf Shadow detection.} It aims to detect/remove shadows from the input images.
Hu~\etal~\cite{Hu_2018_CVPR} propose to use direction-aware features to analyze the contrasts between shadow/non-shadow regions for shadow detection. Le~\etal~\cite{Le_2018_ECCV} propose to train a shadow detection network with augmented adversarial shadow examples generated from a shadow attenuation network. Zhu~\etal~\cite{Zhu_2018_ECCV} propose a bidirectional feature pyramid to leverage the spatial contexts from shallow and deep CNN layers. A conditional GAN~\cite{mirza2014conditional} is also applied to model both local features and global image semantics for shadow detection~\cite{Nguyen2017Shadow} and removal~\cite{Wang_2018_CVPR}. Qu~\etal~\cite{Qu2017CVPR} propose a multi-context network, together with a new dataset, for shadow removal.

In general, shadow detection methods are largely based on detecting the intensity contrast between shadow and non-shadow regions. In \rf{contrast}, the contents inside and outside of a mirror typically have very similar intensity, making the mirror segmentation problem more difficult to address.

{\bf Mirror detection in the 3D community.} To our knowledge, there are only two works that consider mirror segmentation in 3D reconstruction. Matterport3D~\cite{chang2017matterport3d} proposes the user to manually segment the mirrors on an iPad during scanning. Whelan~\etal~\cite{whelan2018reconstructing} attach a hardware tag (based on the AprilTag~\cite{olson2011apriltag}) to the scanner. If a tag is detected in the captured image, it signals the presence of a mirror. A total variation-based segmentation method is then used to segment the mirror based on a set of hand-crafted features (\eg, depth discontinuity and intensity variance).

Instead of using any special hardware, in this paper, we propose the first automatic method for mirror segmentation and the first mirror dataset with mirror annotations.

\section{Mirror Segmentation Dataset}\label{sec:dataset}

To address the mirror segmentation problem, we construct the first large-scale mirror dataset, named MSD. It includes $4,018$ pairs of images containing mirrors and their corresponding manually annotated masks.

\renewcommand{\thefootnote}{1}
{\bf Dataset construction.} We use several latest smartphones for capturing images and Labelme\footnote{https://github.com/wkentaro/labelme} for manual labeling of mirrors. While capturing the images, we consider common types of mirrors (including cosmetic, dressing, decorative, bathroom, and road mirrors) that are often used in our daily life scenes (\eg, bedroom, living room, office, garden, street, and parking lot). Some example mirror images in our MSD dataset are shown in Figure~\ref{fig:dataset}. The dataset contains 3,677 images taken from indoor scenes and 341 images taken from outdoor scenes. The reason that we have many more indoor images than outdoor ones is that we want to focus on indoor scenes in this work. The outdoor images are mainly to provide more diverse mirror shapes and scenarios. For splitting the dataset into training and test sets in a fair way, we first divide the images into different groups based on the mirror types. Since we may have taken several images using each specific mirror with different combinations of foreground/background objects and camera orientations, to make sure that mirrors appearing in the training set do not appear in the test set, we split the images by randomly splitting the mirror types. Finally, we have 3,063 images for training and 955 images for testing.

{\bf Dataset analysis.} Figure~\ref{fig:statistic} shows statistical analysis on the mirror properties in our captured images (including mirror area, shape, location in the image, and global color contrast between inside/outside of the mirror) for a comprehensive understanding of the proposed MSD dataset.

\def\wdataset{0.5\linewidth}
\def\hdataset{1in}
\begin{figure}
\setlength{\tabcolsep}{.10pt}
  \centering
  \begin{tabular}{cc}
    \includegraphics[width=\wdataset, height=\hdataset]{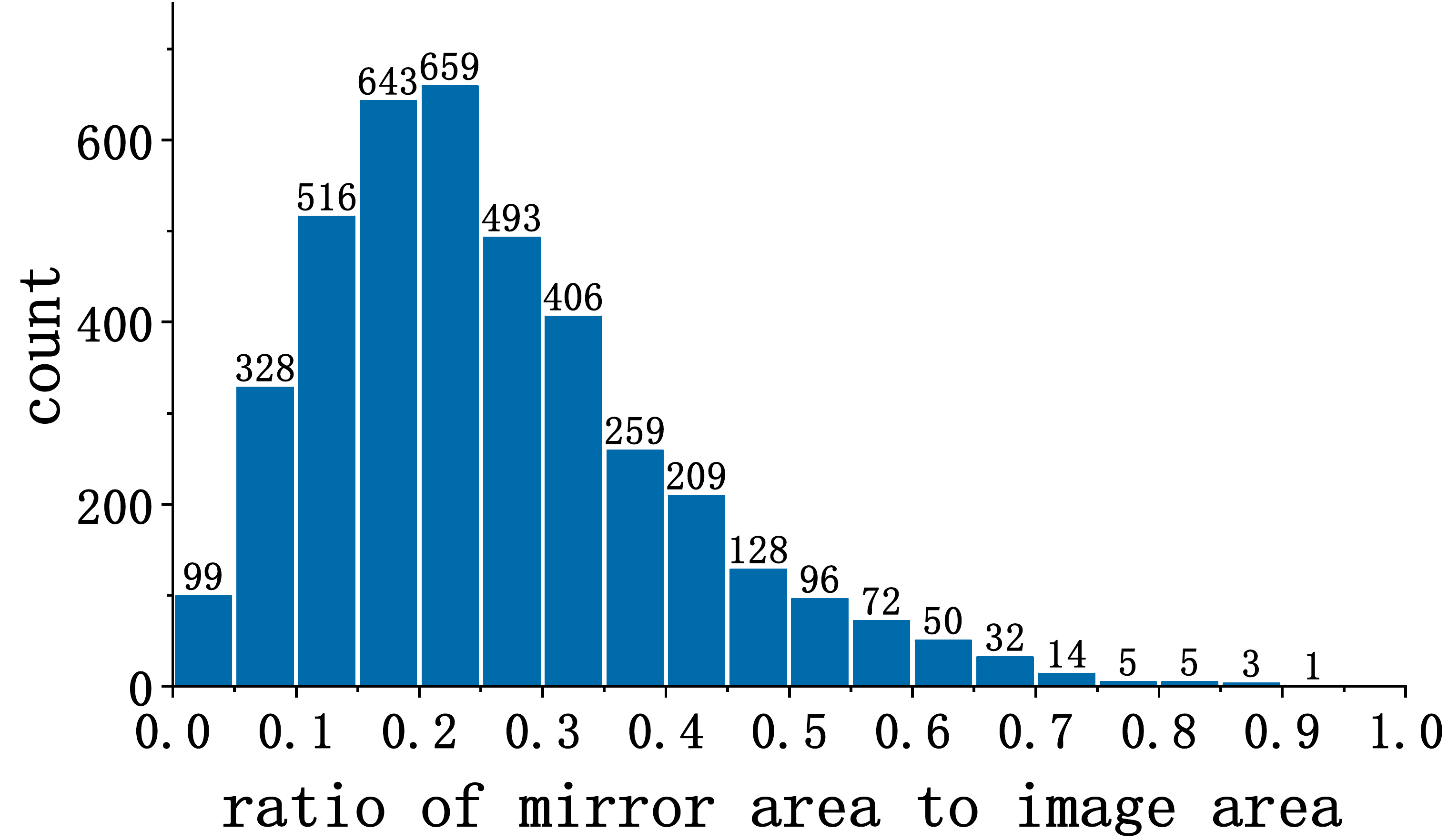}&
    \includegraphics[width=\wdataset, height=\hdataset]{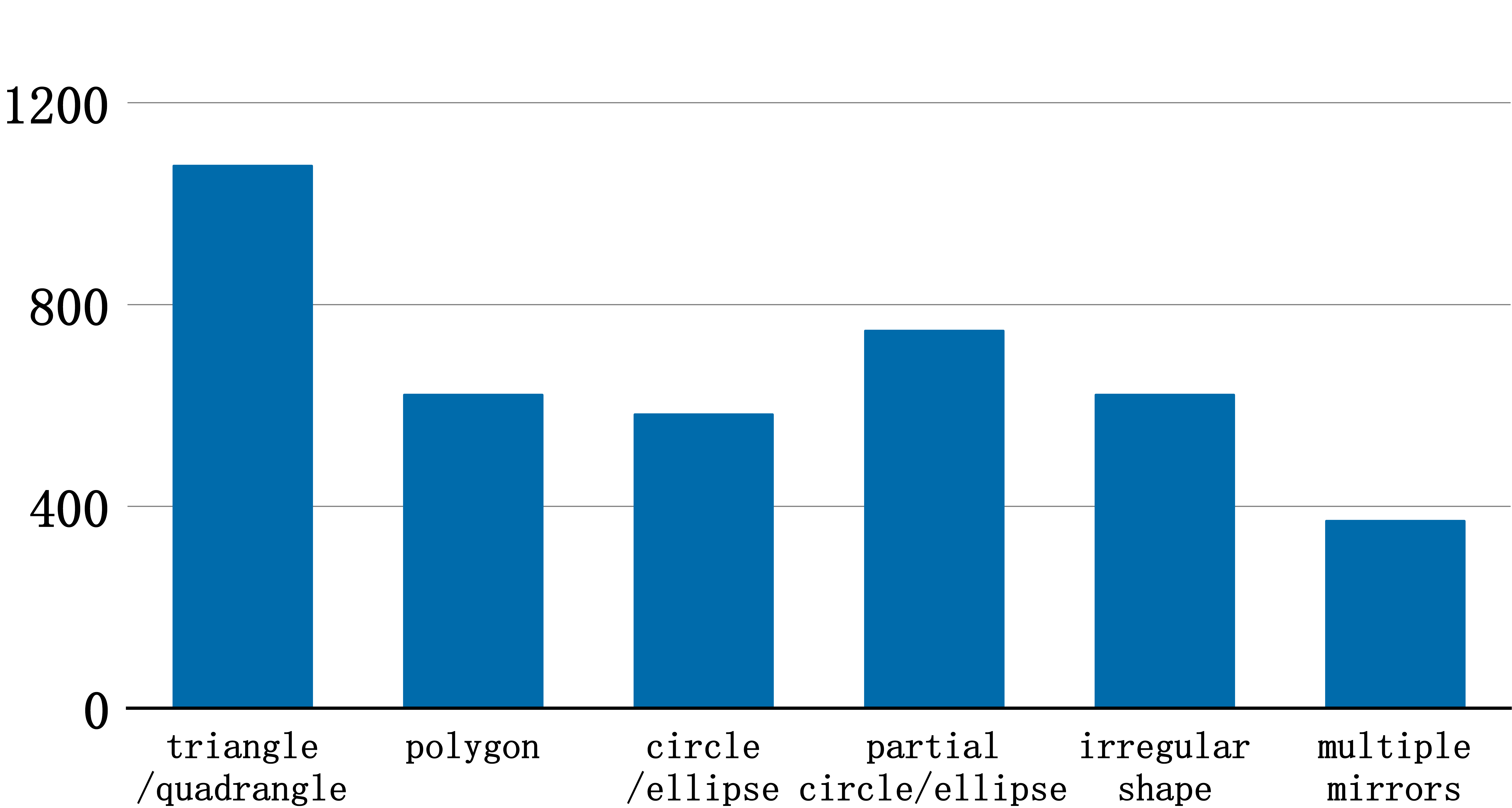} \\
    {\small (a) mirror area distribution} & {\small (b) mirror shape distribution} \vspace{0.1in} \\
    \includegraphics[width=\wdataset, height=\hdataset]{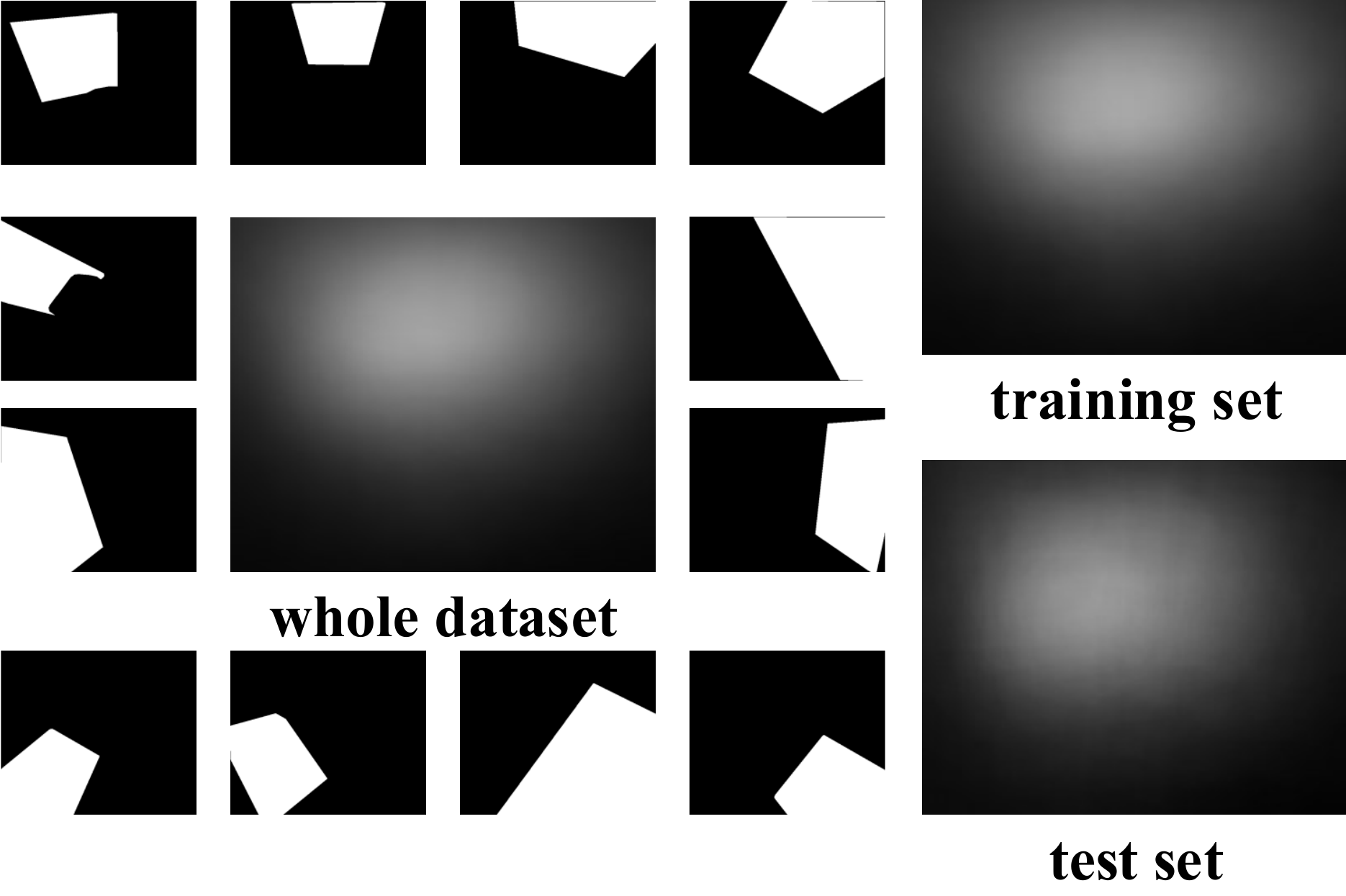}&
    \includegraphics[width=\wdataset, height=\hdataset]{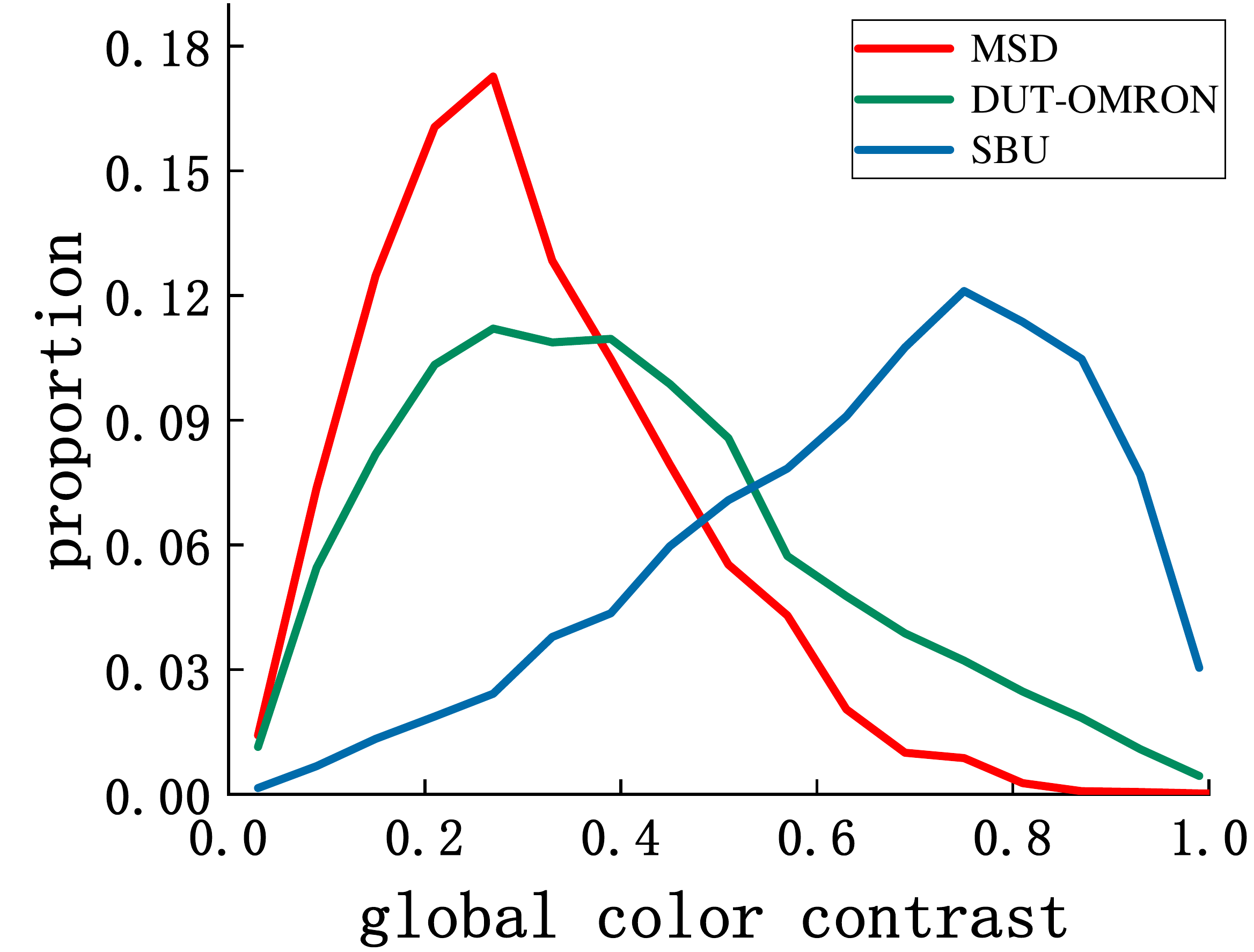} \\
    {\small (c) mirror location distribution} & {\small(d) color contrast distribution} \\
    \end{tabular}
    \vspace{-0.1in}
  \caption{Statistics of the MSD dataset. We show that MSD \rf{has} mirrors with reasonable property distributions, including mirror area, silhouette, location and color contrast.}
  \label{fig:statistic}\vspace{-4mm}
\end{figure}

\begin{figure*}
\begin{center}
\includegraphics[width = 0.9\linewidth]{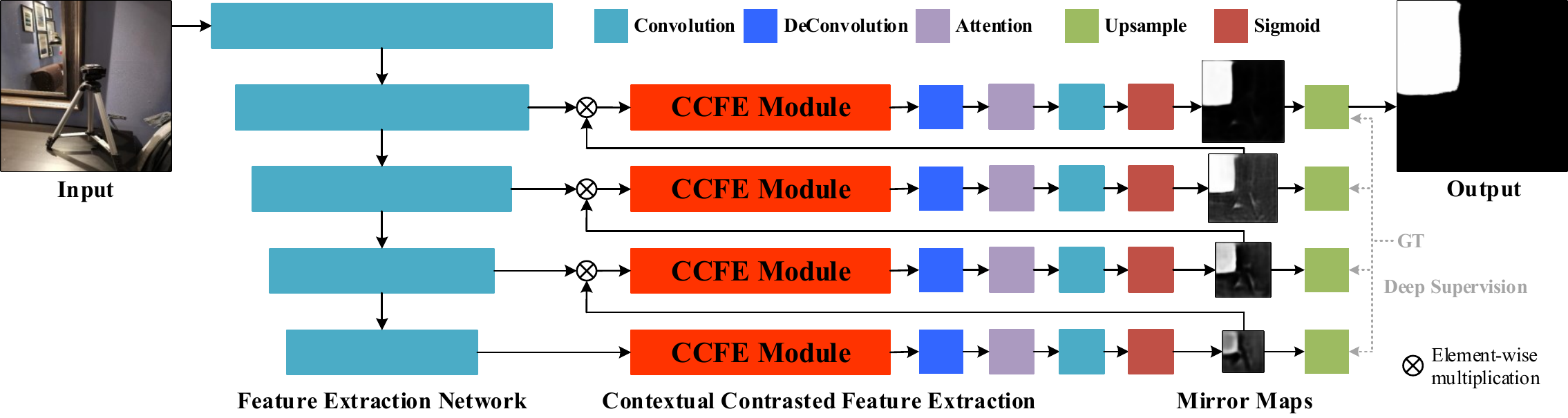}
\end{center}
\vspace{-5mm}
   \caption{Overview of MirrorNet.
   First, a pre-trained Feature Extraction Network is used to extract multi-scale feature maps.
   Second, CCFE modules are embedded to different layers of the Feature Extraction Network to learn different scales of contextual contrasted features.
   Third, MirrorNet leverages these different scales of features in a coarse to fine manner to produce mirror maps, which function as attention maps to help the upper layers focus on learning contextual contrasted features in the candidate mirror regions.
   Fourth, the coarsest mirror map is progressively refined and increased in spatial resolution as it propagates from the bottom layers up to the upper layers.}
   %
\label{fig:pipeline}
\vspace{-2mm}
\end{figure*}

\begin{itemize}
\vspace{-2mm}
  \item {\bf Mirror area distribution.} We define it as a ratio between the mirror area and image area. As shown in Figure \ref{fig:statistic}(a), majority of the mirrors fall in the range of $(0.0, 0.7]$. Mirrors falling in the range of $(0.0, 0.1]$ are small mirrors that can easily be cluttered with other background objects. Mirrors falling in the range of $[0.5, 0.95]$ are typically located close to the camera. Foreground object occlusion often happens in this situation. Mirrors falling in the range of $[0.95, 1.0]$ are not included in MSD, as the images may not provide sufficient contextual information even for \rf{humans} to determine \rf{whether} there is a mirror in them.
\vspace{-2mm}

  \item {\bf Mirror shape distribution.} There are some popular mirror shapes (\eg, elliptic and rectangular). However, if a mirror is partially occluded by an object in front of it, the resulting shape of the mirror becomes \rf{irregular}. Figure \ref{fig:statistic}(b) shows that MSD includes images of different mirror shapes and multiple mirrors.
\vspace{-2mm}

  \item {\bf Mirror location distribution.} To analyze the spatial distribution of mirrors in MSD, we compute probability maps to show how likely each pixel belongs to a mirror, as in Figure~\ref{fig:statistic}(c). Although our MSD has mirrors covering different locations, the mirrors tend to cluster around the upper part of the image. This is reasonable as mirrors are usually placed approximately around the human eyesight. We can also see that the mirror location distributions for the training/test splits are consistent to that of the whole dataset.

\vspace{-2mm}
  \item {\bf Color contrast distribution.} As mirrors can reflect unpredictable contents, we analyze the global color contrast between the contents inside/outside of the mirrors, to check if mirror contents in our dataset are salient and can easily be detected. We use $\chi^2$ distance to measure the contrasts between two RGB histograms computed separately from mirror and non-mirror regions, similar to~\cite{li2014secrets,fan2018salient}. We further compare this distribution to two existing datasets, \ie, the DUT-OMRON saliency dataset \cite{yang2013saliency} and SBU shadow dataset \cite{vicente2016large}, as shown in Figure \ref{fig:statistic}(d). We can see that MSD has the lowest global color contrast, making the mirror segmentation task more challenging.

\end{itemize}

\begin{figure*}
\begin{center}
\includegraphics[width = 1\linewidth]{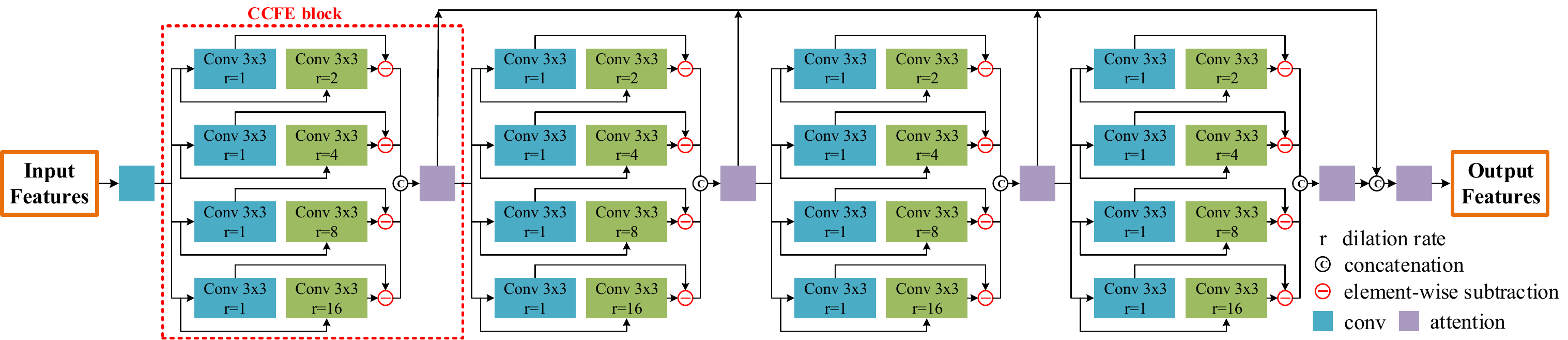}
\end{center}
\vspace{-5mm}
   \caption{The Contextual Contrasted Feature Extraction (CCFE) module. The input features are passed through four chained CCFE \rf{blocks} and the output of each CCFE block are fused via an attention module to generate multi-level contextual contrasted features. In each CCFE block (red dashed box), we first compute the contextual contrasts between local information (extracted by standard convolutions) and their surrounding contexts (extracted by dilated convolutions with different dilation rates) in parallel, and then adaptively select useful ones from these concatenated multi-scale contextual contrasted features via an attention module.}

\label{fig:ccfe}
\vspace{-5mm}
\end{figure*}

\section{Proposed Network}

We observe that in order for humans to know if we are looking at a mirror,  we typically look for content discontinuity, in terms of low-level color/texture changes as well as high-level semantic information. This inspires us to leverage the contrast between the mirror and non-mirror regions. To this end, we propose a novel Contextual Contrasted Feature Extraction (CCFE) block to extract multi-scale contextual contrasted features for mirror localization. Building upon the CCFE block, a novel CCFE module is designed to hierarchically aggregate long-range contextual contrasted information to effectively detect mirrors of different sizes.

\subsection{Overview}
Figure \ref{fig:pipeline} illustrates the proposed mirror segmentation network, called MirrorNet. It takes a single image as input and extracts multi-level features by the feature \rf{extraction} network (FEN). The deepest features, which are full of semantics, are then fed to the proposed CCFE module to learn contextual contrasted features for locating the mirrors with the coarsest mirror map, by detecting the dividing boundaries where the contrasts \rf{appear}. This mirror map functions as an attention map to suppress the feature noise of the next-upper FEN features in the non-mirror regions, so that the next-upper layer can focus on learning discriminative features in the candidate mirror regions. In this way, MirrorNet progressively leverages contextual contrasted information to refine the mirror region in a coarse-to-fine manner. Finally, we upsample the coarsest network output to obtain the original image resolutions as the output.

\subsection{Contextual Contrasted Feature Extraction}\label{section:ccfe}
Figure \ref{fig:ccfe} shows the structure of the proposed CCFE module. Given the features extracted by the Feature Extraction Network, the CCFE \rf{aims} to produce multi-scale contextual contrasted features for detecting mirrors of different sizes.

{\bf CCFE block.} To effectively detect mirror boundaries (where contents may change significantly), we design the CCFE block to learn contextual contrasted features between a local region and its surrounding, as:
\begin{equation}\label{eq:contrast}
\mathbb{CCF} = f_{local}(\textbf{F}, \Theta_{local}) - f_{context}(\textbf{F}, \Theta_{context}),
\end{equation}
where $\textbf{F}$ is the input features. $f_{local}$ represents a local convolution with a $3\times3$ kernel (dilation rate = 1). $f_{context}$ represents a context convolution with a $3\times3$ kernel (dilation rate = $x$). $\Theta_{local}$ and $\Theta_{context}$ are parameters. $\mathbb{CCF}$ is the desired contextual contrasted features.

We further propose to learn multi-scale contextual contrasted features to avoid the ambiguities caused by nearby real objects and their reflections in the mirror, by considering non-local contextual contrast. Hence, we set the dilation rate $x$ to 2, 4, 8, and 16, such that long-range spatial contextual contrast can be obtained. The multi-scale contextual contrasted features are then concatenated and refined via the attention module \cite{woo2018cbam}, to produce feature maps that highlight the dividing boundaries.

{\bf CCFE module.} A large mirror can easily cause under-segmentation, as the content inside it may exhibit high contrast within itself. To address this problem, global image contexts should be considered. Hence, we propose to leverage the global contextual contrast by cascading the CCFE blocks to form a deep CCFE module with larger receptive fields, such that the global image contexts are captured in deeper blocks of the CCFE module. We also adopt the attention module \cite{woo2018cbam} to highlight the candidate mirror regions of the concatenated multi-level features from different blocks in the CCFE module.

{\bf Discussion.} Although we have drawn some inspiration from the Context Contrast Local (CCL) block in~\cite{ding2018context} in our network design, our network is different from the CCL block in both motivations and implementations. First, while the CCL block aims to detect small objects, our CCFE modules are used to locate mirrors by detecting the dividing boundaries. They also serve as attention modules to enhance the feature responses in mirror regions and suppress the feature noise in the non-mirror regions. Second, the CCL block has only one scale \rf{of contrast} and is only embedded in the deepest layer for their purpose of small objects detection using semantical contrast. We extend the CCL block to our CCFE module by incorporating multi-scale contextual contrasted feature extraction, to provide sufficient contextual information for locating mirrors in different sizes. We also embed our CCFE modules to all side-outputs of the feature extraction network, such that our network takes advantages of both rich semantical contrasted contexts from deeper layers and low-level contrasted contexts from upper layers, for mirror segmentation.

\subsection{Loss Function}\label{section:loss}
Per-pixel cross entropy is commonly used as the loss function in semantic segmentation, salient object detection and shadow detection problems. However, it is not sensitive to small objects, and can easily be dominated by large objects. Hence, we choose the lov\'{a}sz-hinge loss~\cite{Berman_2018_CVPR} for optimizing our network. It is a surrogate for the non-differentiable \rf{intersection over union} (IoU) metric, which preserves the scale invariance property of the IoU metric. Deep supervision~\cite{xie2015holistically} is also adopted to facilitate the learning process. The loss function is:
\begin{equation}\label{final_loss}
Loss = \sum_{s=1}^{S}w_sL_s,
\end{equation}
where $w_s$ \rf{represents} the balancing parameters. $L_s$ is the lov\'{a}sz-hinge loss between the $s$-$th$ upsampled mirror map and the ground truth.

\subsection{Implementation Details}\label{section:implementation_details}
We have implemented MirrorNet on the PyTorch framework \cite{pytorch}. For training, input images are resized to a resolution of $384\times384$ and are augmented by horizontally random flipping. We use the pre-trained ResNeXt101 network \cite{xie2017aggregated} as the feature extraction network. The remaining parts of our network are randomly initialized. For loss optimization, we use the stochastic gradient descent (SGD) optimizer with momentum of $0.9$ and a weight decay of $5\times10^{-4}$. Batch size is set to $10$. The learning rate is initialized to $0.001$ and decayed by the poly strategy \cite{liu2015parsenet} with the power of $0.9$, for $160$ epochs.
There are $S=4$ loss terms in Eq.~\ref{final_loss}, and the balancing parameters $w_s$  are empirically set to $1$.
It takes about 12 hours for the network to converge on an NVIDIA Titan V graphics card. For testing, images are also resized to a resolution of $384\times384$ for network inferences. We then use the fully connected conditional random field (CRF) \cite{krahenbuhl2011efficient} to further enhance the network outputs by optimizing the spatial coherence of pixels as the final mirror segmentation results.

\section{Experiments}

\subsection{Experimental Settings}

{\bf Evaluation metrics.} For a comprehensive evaluation, we adopt five metrics that are commonly used in the related fields (\ie, semantic segmentation, salient object detection and shadow detection), for quantitatively evaluating the mirror segmentation performance. Specifically, we use the intersection \rf{over} union (IoU) and \rf{pixel accuracy} metrics from the semantic segmentation field as our first and second metrics.
We also use the F-measure and mean absolute error (MAE) metrics from the salient object detection field. F-measure is defined as the weighted harmonic mean of precision and recall:
\begin{equation}\label{F}
F_\beta=\frac{(1+\beta ^2)Precision \times Recall}{\beta ^2Precision + Recall},
\end{equation}
where $\beta^2$ is set to be 0.3 to emphasize more on precision over recall, as suggested in \cite{achanta2009frequency}.

Finally, we adopt the balance error rate (BER) from the shadow detection field, to evaluate the mirror segmentation performance. It considers the unbalanced areas of mirror and non-mirror regions, and is computed as:
\begin{equation}\label{ber}
BER = 100 \times (1 - \frac{1}{2}(\frac{TP}{N_{p}} + \frac{TN}{N_{n}})),
\end{equation}
where $TP$, $TN$, $N_p$ and $N_n$ represent the numbers of true positives, true negatives, mirror pixels, and non-mirror pixels, respectively.

{\bf Compared methods.} We select the state-of-the-art methods from the related fields for comparison. Specifically, we choose PSPNet \cite{zhao2017pyramid} and ICNet \cite{Zhao_2018_ECCV} from the semantic segmentation field, Mask RCNN~\cite{he2017mask} from the instance segmentation field, DSS \cite{Hou_2017_CVPR}, PiCANet \cite{liu2018picanet}, RAS \cite{Chen_2018_ECCV} and R\textsuperscript{3}Net \cite{deng2018r3net} from the salient object detection field, DSC \cite{Hu_2018_CVPR} and BDRAR \cite{Zhu_2018_ECCV} from the shadow detection field. We use their publicly available codes and train them on our proposed training set for a fair comparison.

\begin{table}[bp]
\begin{small}
\setlength{\tabcolsep}{2.7pt}
  \centering
    \begin{tabular}{p{2.6cm}<{\centering}|p{0.6cm}<{\centering}|
    p{0.78cm}<{\centering}|p{0.78cm}<{\centering}|p{0.72cm}<{\centering}|
    p{0.78cm}<{\centering}|p{0.78cm}<{\centering}}
    \hline
    method & CRF & IoU$\uparrow$ & Acc$\uparrow$ & $F_\beta$$\uparrow$ & MAE$\downarrow$ & BER$\downarrow$ \\
    \hline
    \hline
    Statistics & - & 30.83 & 0.595 & 0.438 & 0.358 & 32.89 \\
    \hline
    \hline
    PSPNet \cite{zhao2017pyramid} & - & 63.21 & 0.750 & 0.746 & 0.117 & 15.82 \\
    ICNet \cite{Zhao_2018_ECCV} & - & 57.25 & 0.694 & 0.710 & 0.124 & 18.75 \\
    \hline
    \hline
    {\small Mask RCNN} \cite{he2017mask} & - & 63.18 & 0.821 & 0.756 & 0.095 & 14.35 \\
    \hline
    \hline
    DSS \cite{Hou_2017_CVPR} & - & 59.11 & 0.665 & 0.743 & 0.125 & 18.81 \\
    PiCANet \cite{liu2018picanet} & - & 71.78 & 0.845 & 0.808 & 0.088 & 10.99 \\
    RAS \cite{Chen_2018_ECCV} & - & 60.48 & 0.695 & 0.758 & 0.111 & 17.60 \\
    R\textsuperscript{3}Net \cite{deng2018r3net} w/o C & - & 72.69 & 0.805 & 0.840 & 0.080 & 11.46 \\
    R\textsuperscript{3}Net \cite{deng2018r3net} & $\surd$ & 73.21 & 0.805 & 0.846 & \textcolor{red}{0.068} & 11.39 \\
    \hline
    \hline
    DSC \cite{Hu_2018_CVPR} & - & 69.71 & 0.816 & 0.812 & 0.087 & 11.77 \\
    BDRAR \cite{Zhu_2018_ECCV} w/o C & - & 67.01 & 0.822 & 0.799 & 0.099 & 12.46 \\
    BDRAR \cite{Zhu_2018_ECCV} & $\surd$ & 67.43 & 0.821 & 0.792 & 0.093 & 12.41 \\
    \hline
    \hline
    MirrorNet w/o C & - & \textcolor{red}{78.46} & \textcolor{red}{0.933} & \textcolor{red}{0.857} & 0.085 & \textcolor{red}{6.46} \\
    MirrorNet  & $\surd$ & \bf{78.95} & \bf{0.933} & \bf{0.857} & \bf{0.065} & \bf{6.39} \\
    \hline
    \end{tabular}
    \end{small}
    \vspace{-3mm}
      \caption{Comparison to state-of-the-arts on MSD test set. All methods are trained on MSD training set. ``w/o C'' is without using CRF~\cite{krahenbuhl2011efficient} for post-processing. ``Statistics" refers to thresholding mirror location statistics from our training set as a mirror mask for detection. The best and second best results are marked in {\bf bold} and \textcolor{red}{red}, respectively.}
      \vspace{-3mm}
  \label{tab:comparison}
\end{table}

\def\wvisual{0.086\linewidth}
\def\hvisualshort{.5in}
\def\hvisualtall{.75in}
\begin{figure*}
\setlength{\tabcolsep}{1.6pt}
  \centering
\begin{tabular}{ccccccccccc}
    \includegraphics[width=\wvisual, height=\hvisualshort]{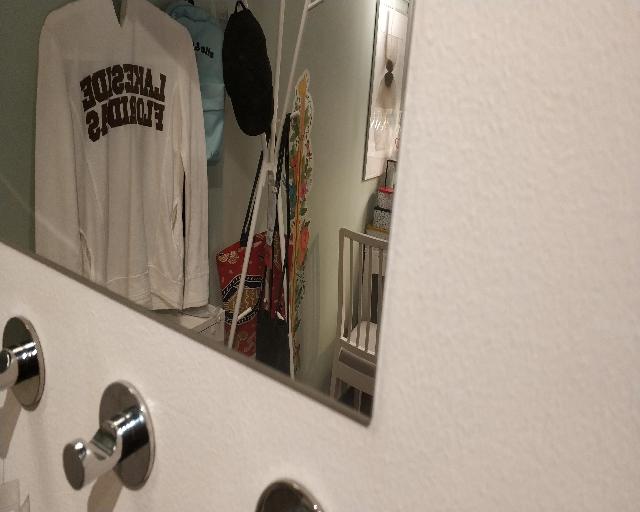}&
    \includegraphics[width=\wvisual, height=\hvisualshort]{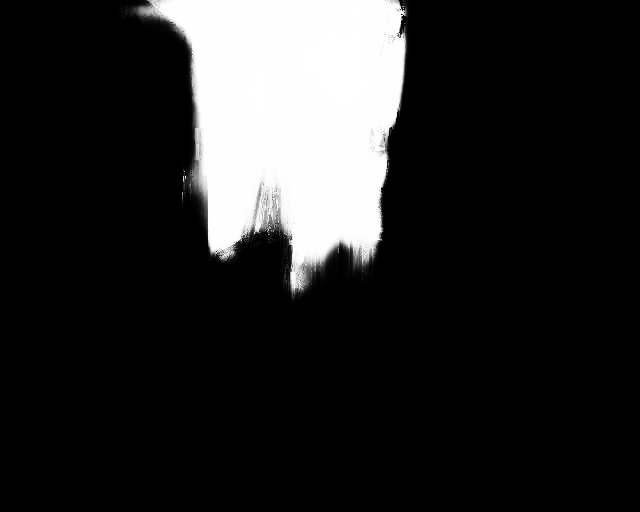}&
    \includegraphics[width=\wvisual, height=\hvisualshort]{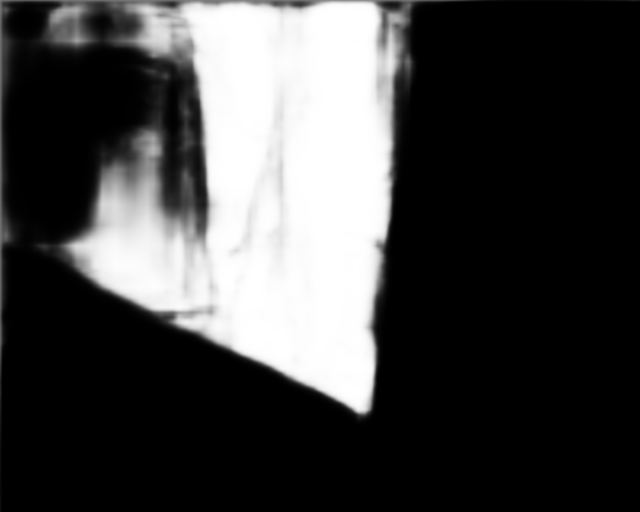}&
    \includegraphics[width=\wvisual, height=\hvisualshort]{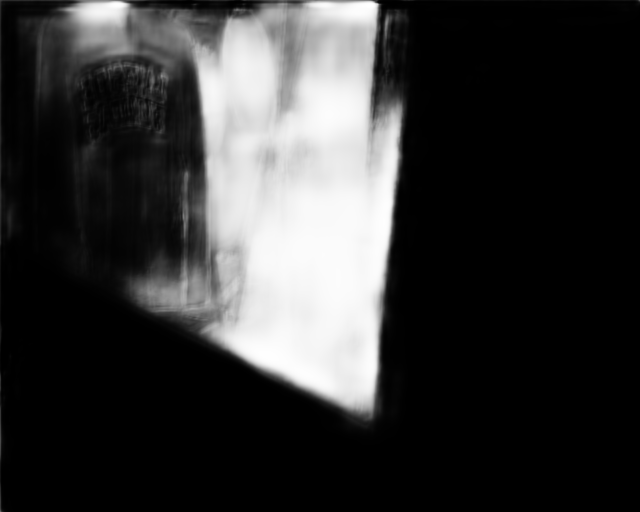}&
    \includegraphics[width=\wvisual, height=\hvisualshort]{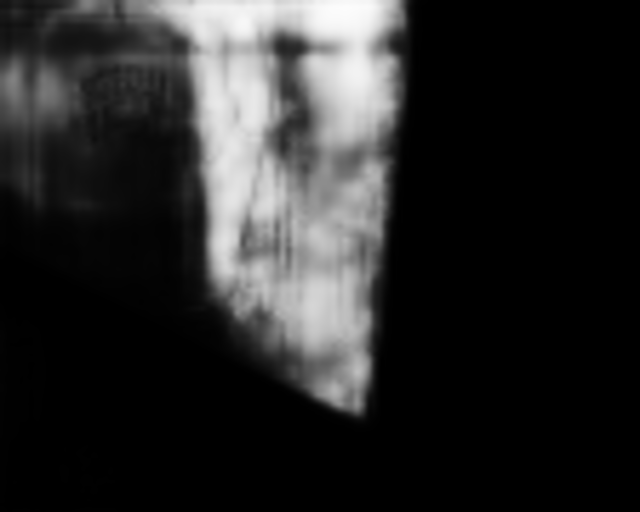}&
    \includegraphics[width=\wvisual, height=\hvisualshort]{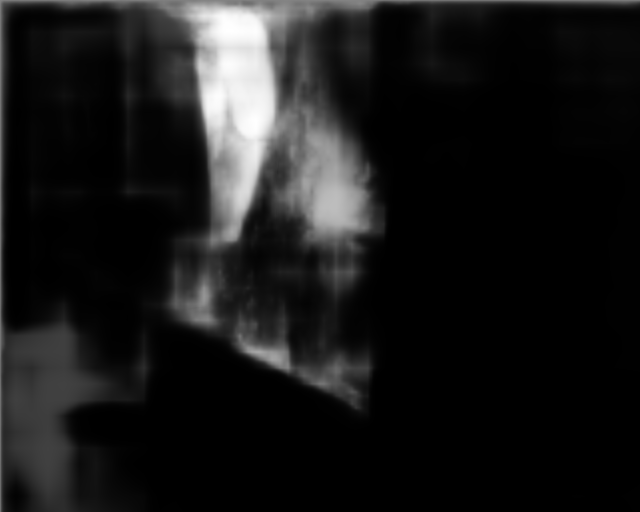}&
    \includegraphics[width=\wvisual, height=\hvisualshort]{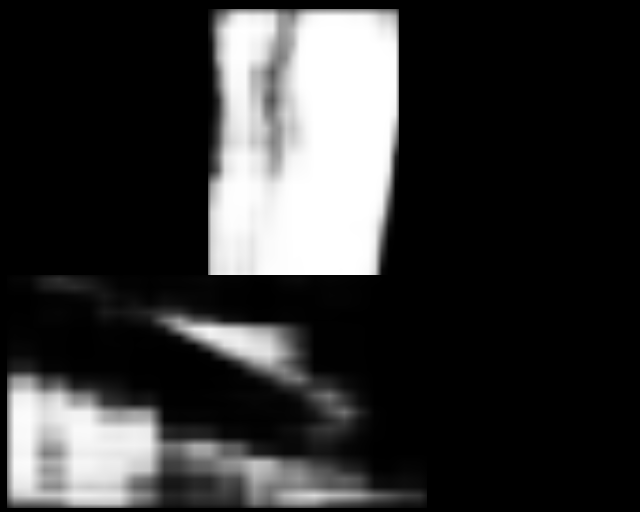}&
    \includegraphics[width=\wvisual, height=\hvisualshort]{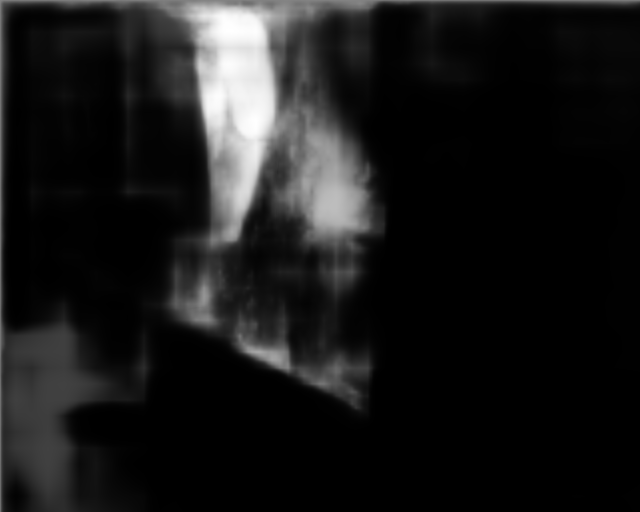}&
    \includegraphics[width=\wvisual, height=\hvisualshort]{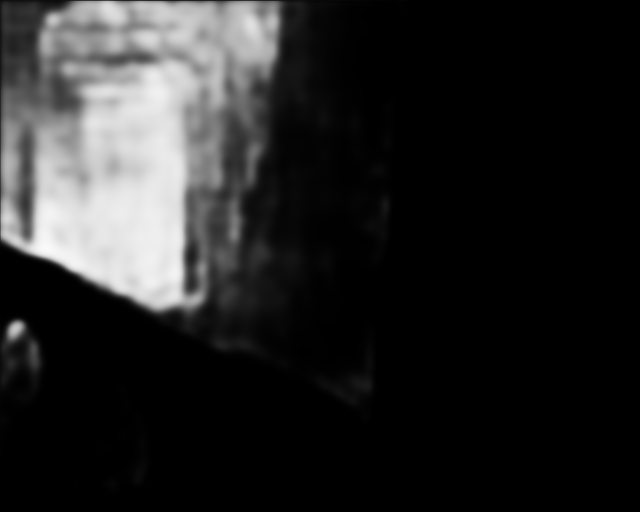}&
    \includegraphics[width=\wvisual, height=\hvisualshort]{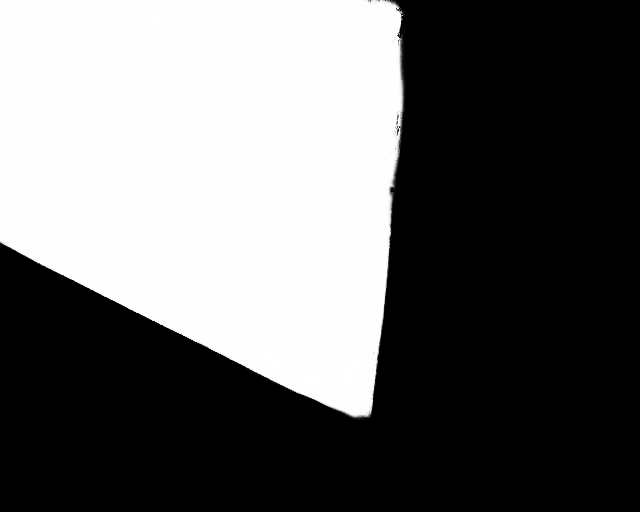}&
    \includegraphics[width=\wvisual, height=\hvisualshort]{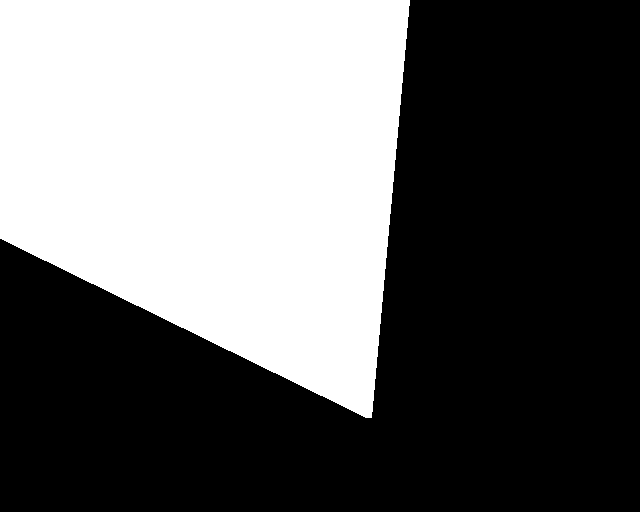} \\

    \includegraphics[width=\wvisual, height=\hvisualtall]{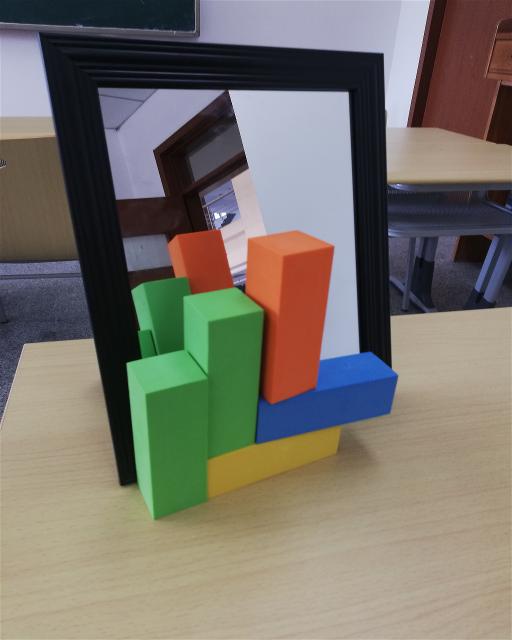}&
    \includegraphics[width=\wvisual, height=\hvisualtall]{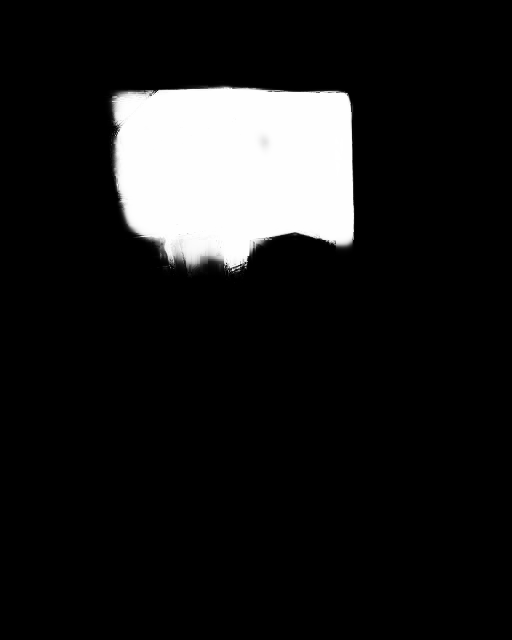}&
    \includegraphics[width=\wvisual, height=\hvisualtall]{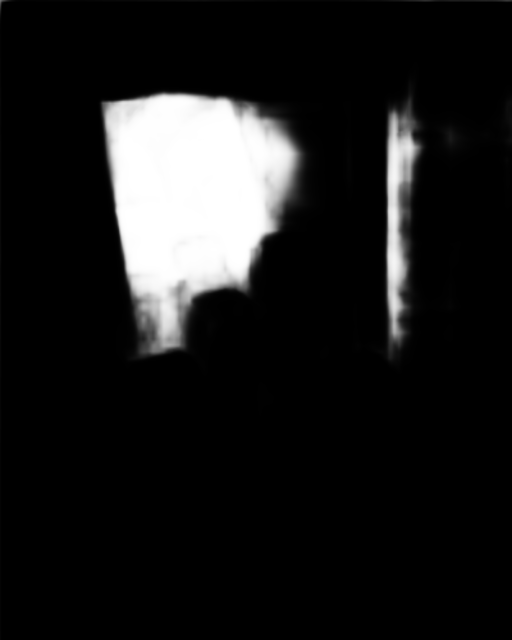}&
    \includegraphics[width=\wvisual, height=\hvisualtall]{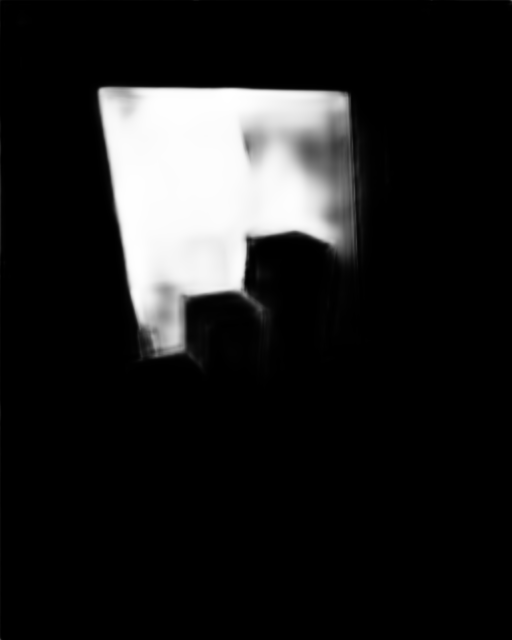}&
    \includegraphics[width=\wvisual, height=\hvisualtall]{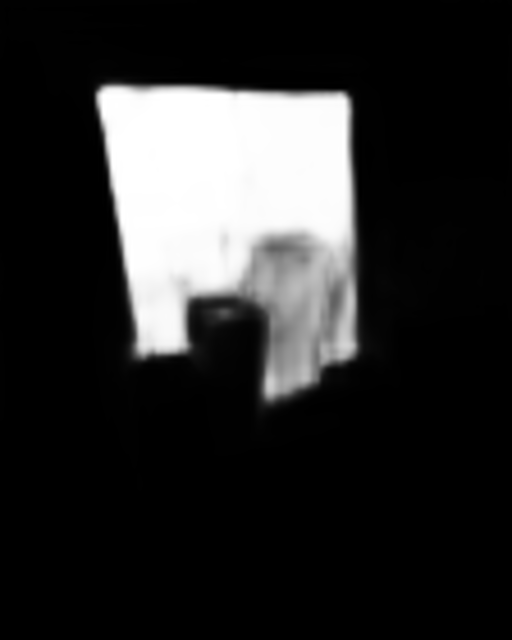}&
    \includegraphics[width=\wvisual, height=\hvisualtall]{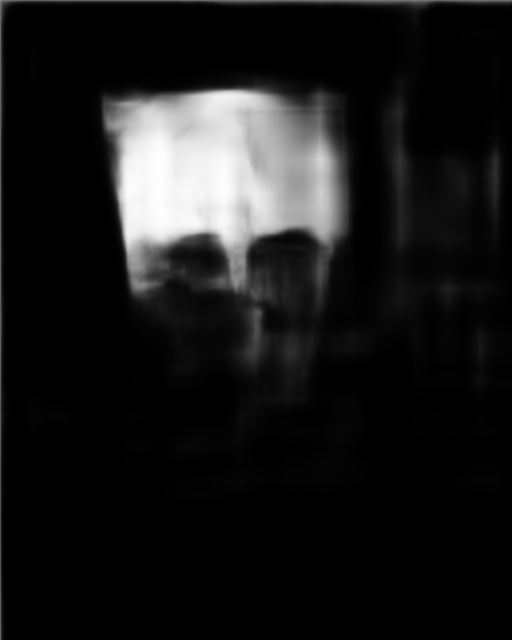}&
    \includegraphics[width=\wvisual, height=\hvisualtall]{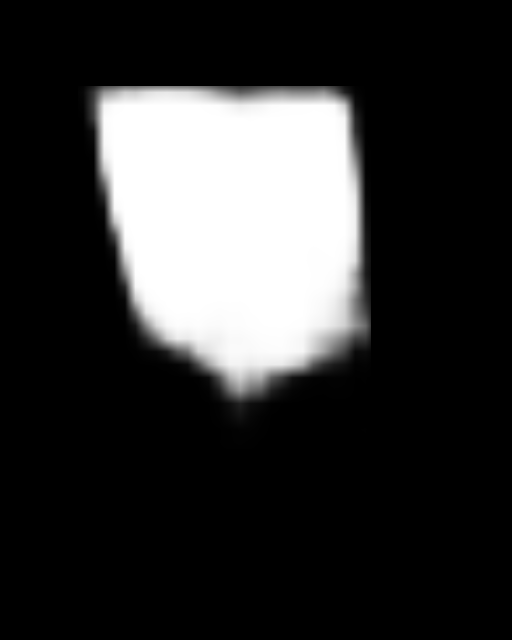}&
    \includegraphics[width=\wvisual, height=\hvisualtall]{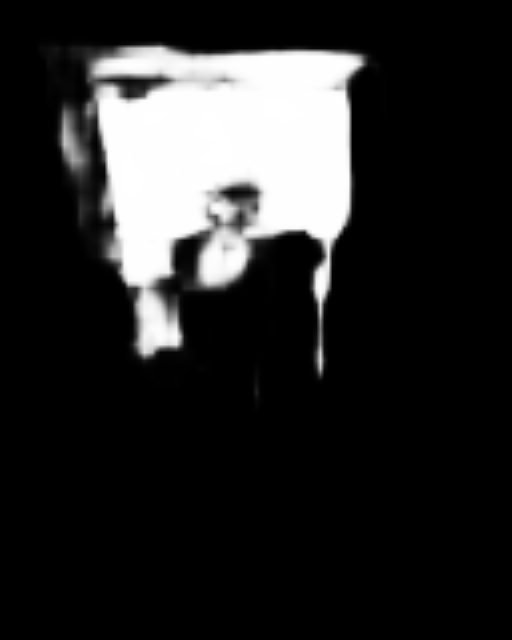}&
    \includegraphics[width=\wvisual, height=\hvisualtall]{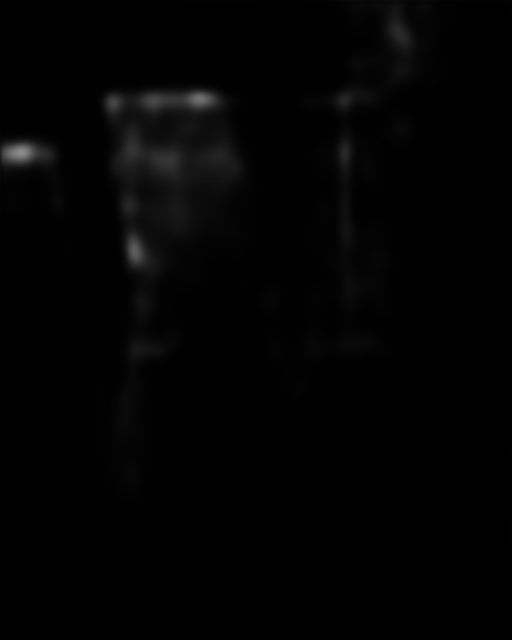}&
    \includegraphics[width=\wvisual, height=\hvisualtall]{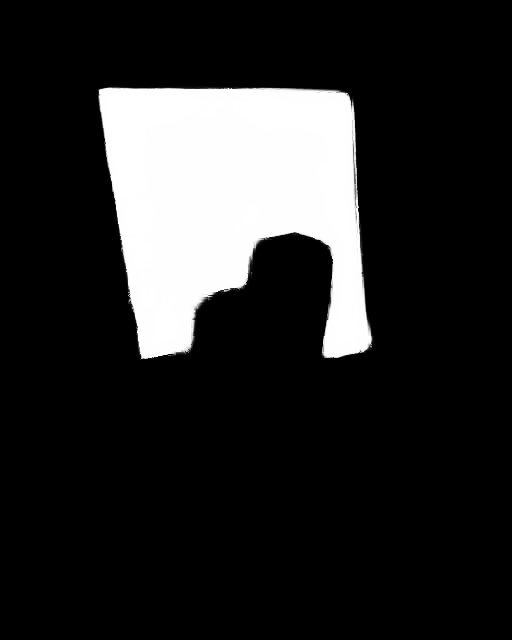}&
    \includegraphics[width=\wvisual, height=\hvisualtall]{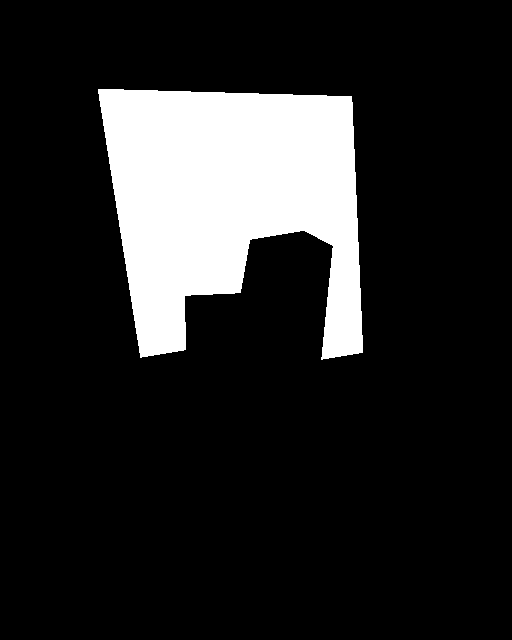} \\

    \includegraphics[width=\wvisual, height=\hvisualtall]{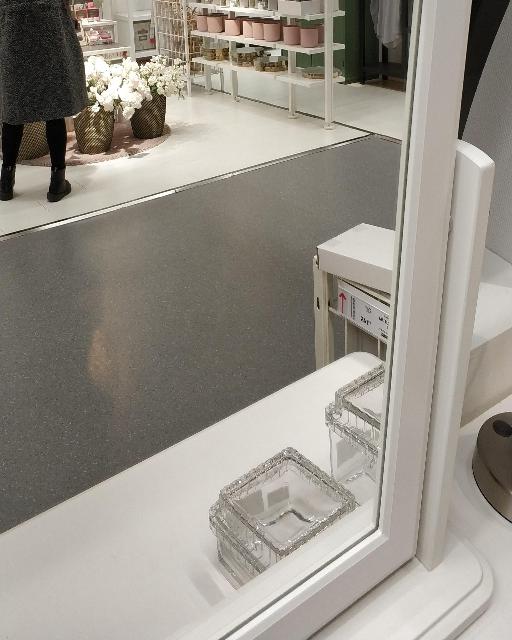}&
    \includegraphics[width=\wvisual, height=\hvisualtall]{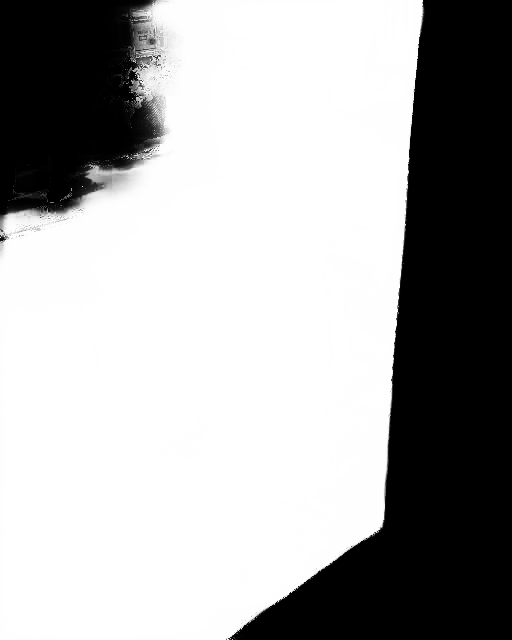}&
    \includegraphics[width=\wvisual, height=\hvisualtall]{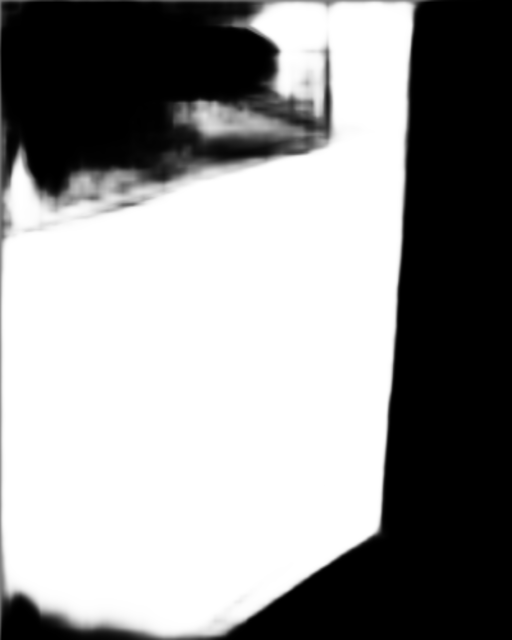}&
    \includegraphics[width=\wvisual, height=\hvisualtall]{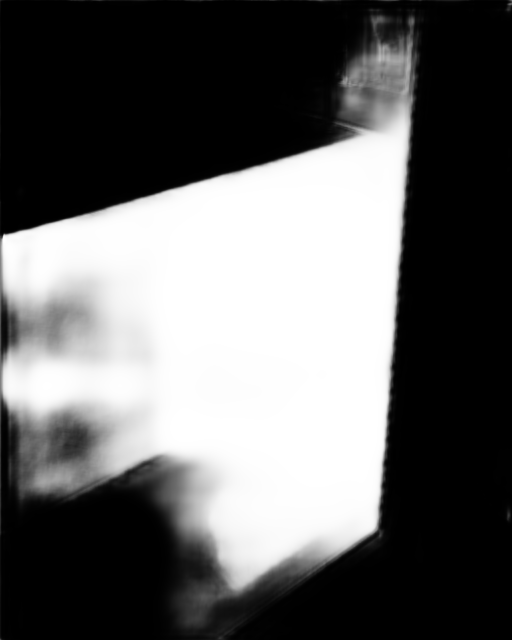}&
    \includegraphics[width=\wvisual, height=\hvisualtall]{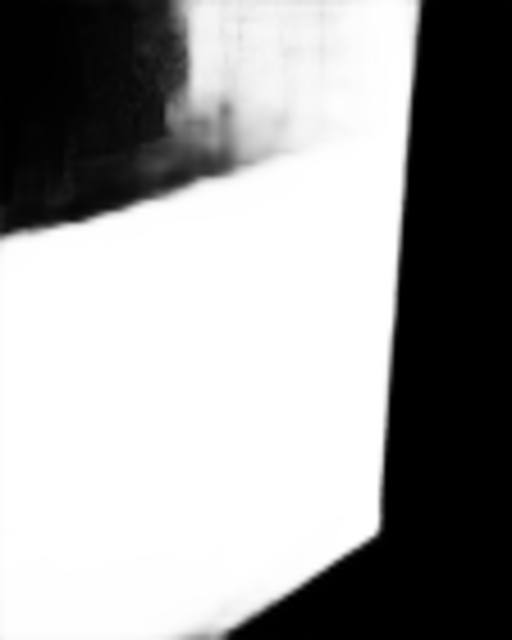}&
    \includegraphics[width=\wvisual, height=\hvisualtall]{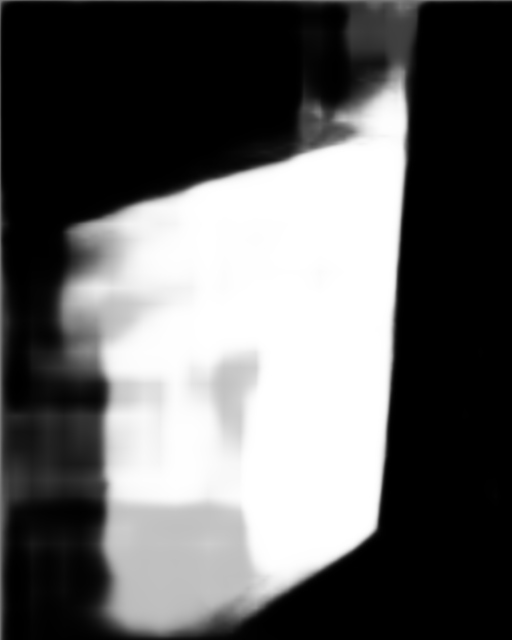}&
    \includegraphics[width=\wvisual, height=\hvisualtall]{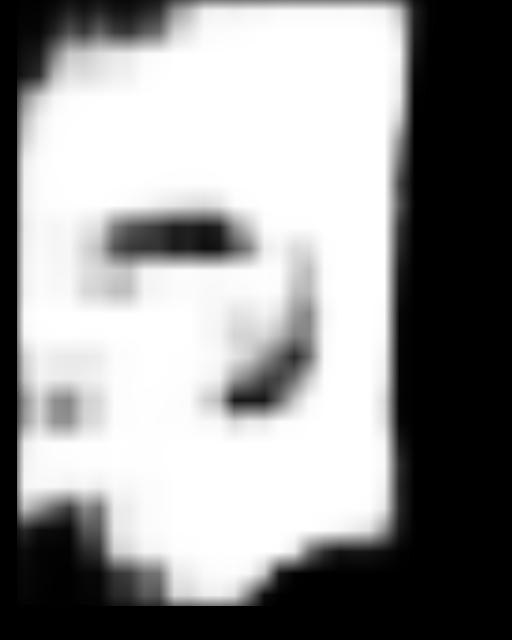}&
    \includegraphics[width=\wvisual, height=\hvisualtall]{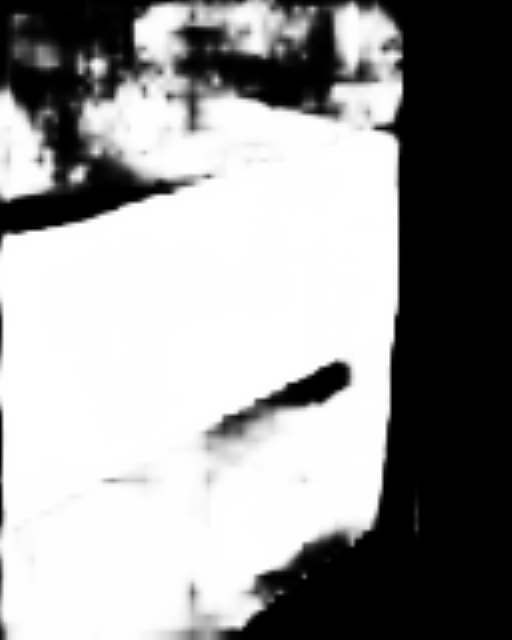}&
    \includegraphics[width=\wvisual, height=\hvisualtall]{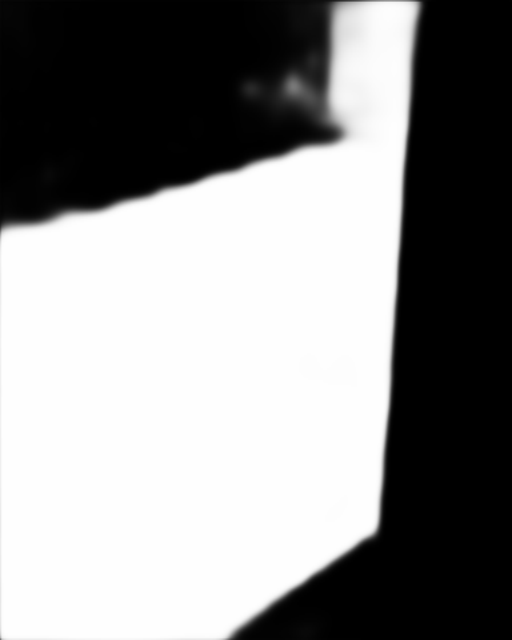}&
    \includegraphics[width=\wvisual, height=\hvisualtall]{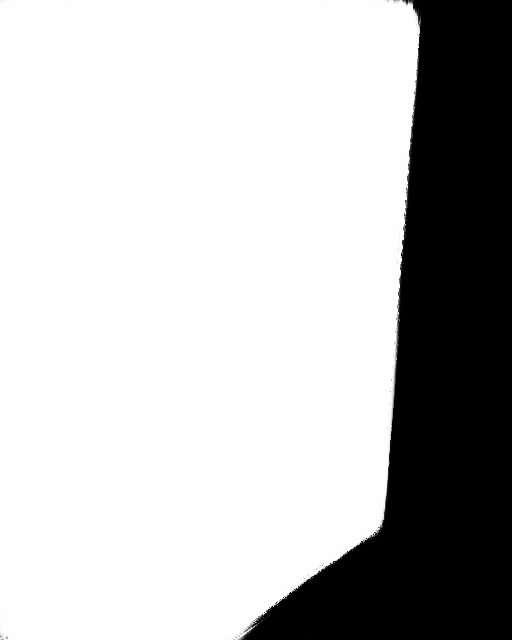}&
    \includegraphics[width=\wvisual, height=\hvisualtall]{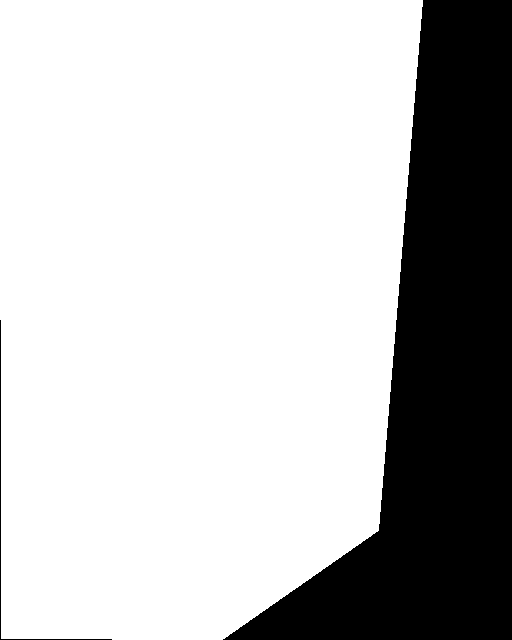} \\

    \includegraphics[width=\wvisual, height=\hvisualtall]{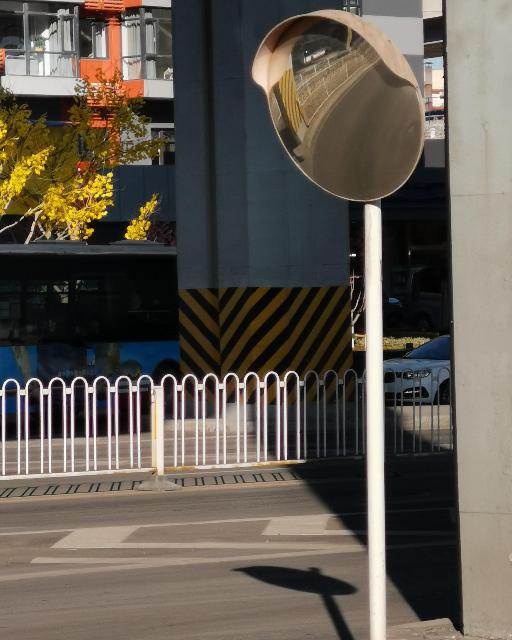}&
    \includegraphics[width=\wvisual, height=\hvisualtall]{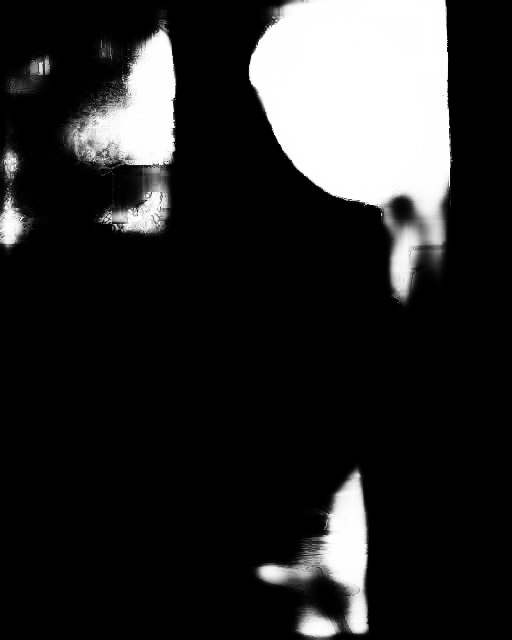}&
    \includegraphics[width=\wvisual, height=\hvisualtall]{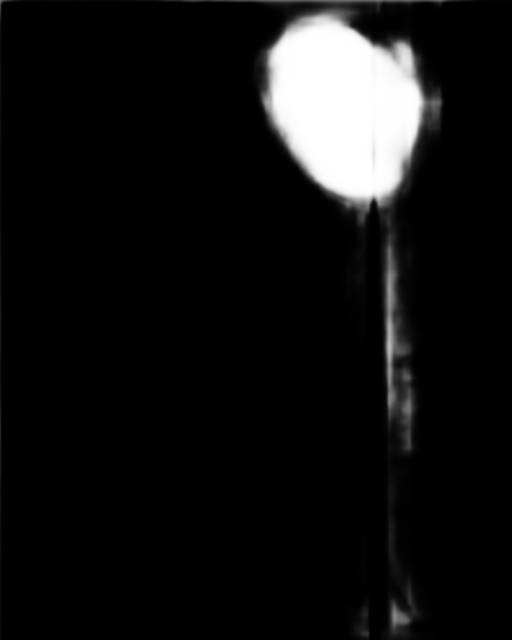}&
    \includegraphics[width=\wvisual, height=\hvisualtall]{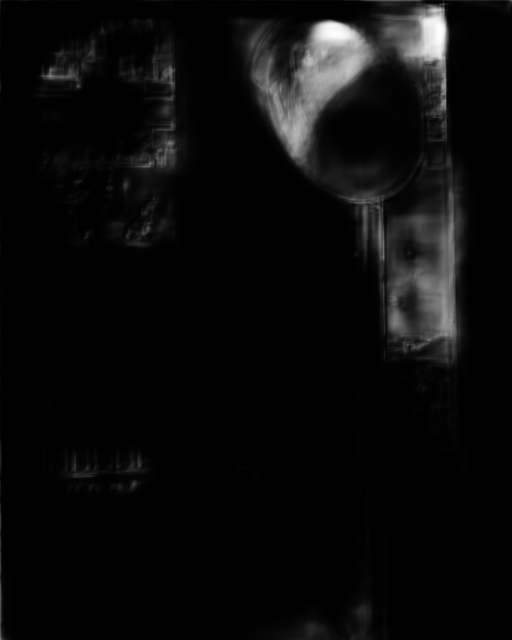}&
    \includegraphics[width=\wvisual, height=\hvisualtall]{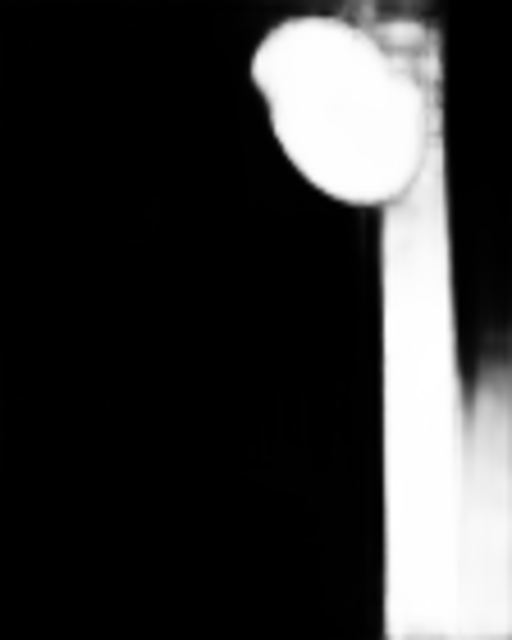}&
    \includegraphics[width=\wvisual, height=\hvisualtall]{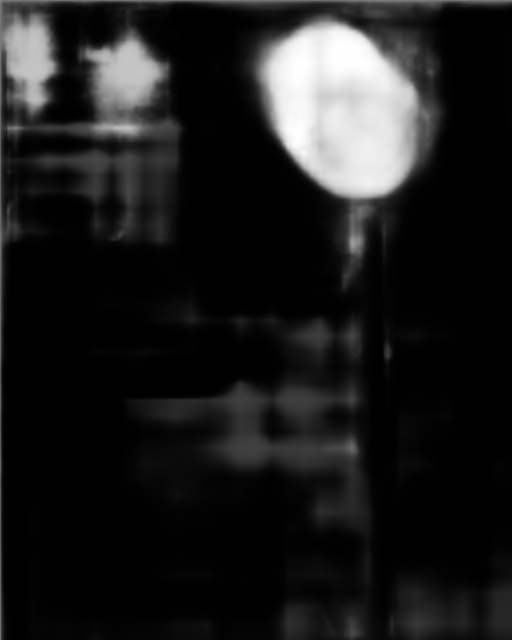}&
    \includegraphics[width=\wvisual, height=\hvisualtall]{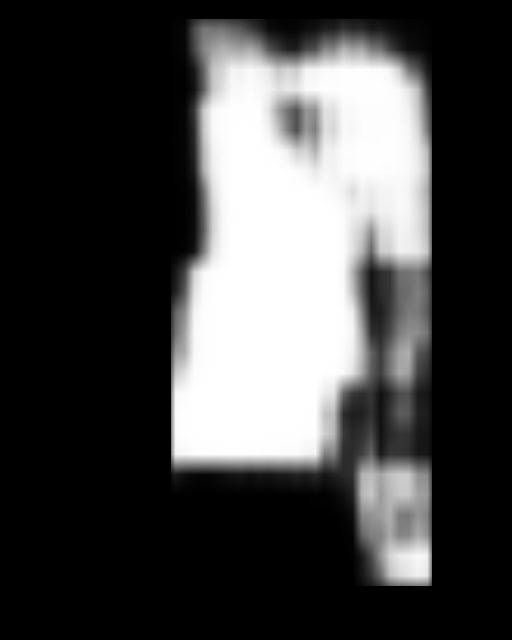}&
    \includegraphics[width=\wvisual, height=\hvisualtall]{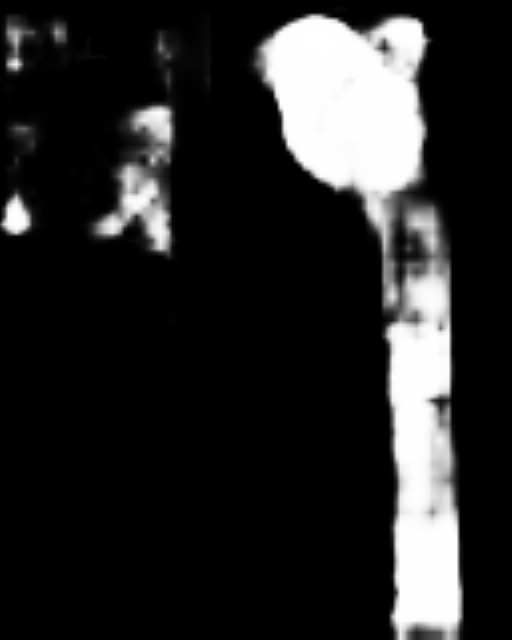}&
    \includegraphics[width=\wvisual, height=\hvisualtall]{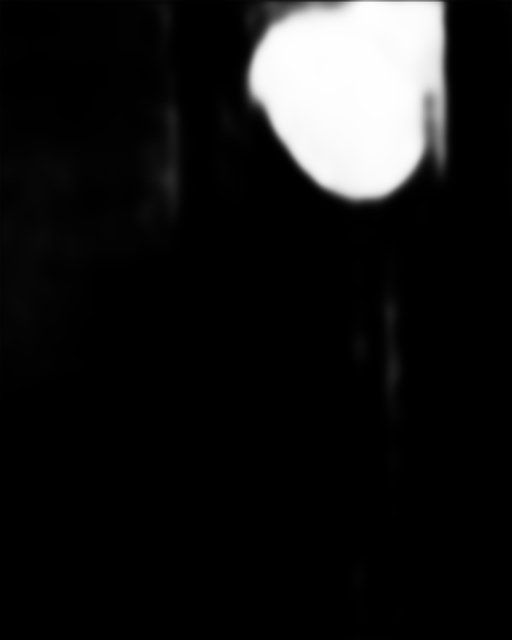}&
    \includegraphics[width=\wvisual, height=\hvisualtall]{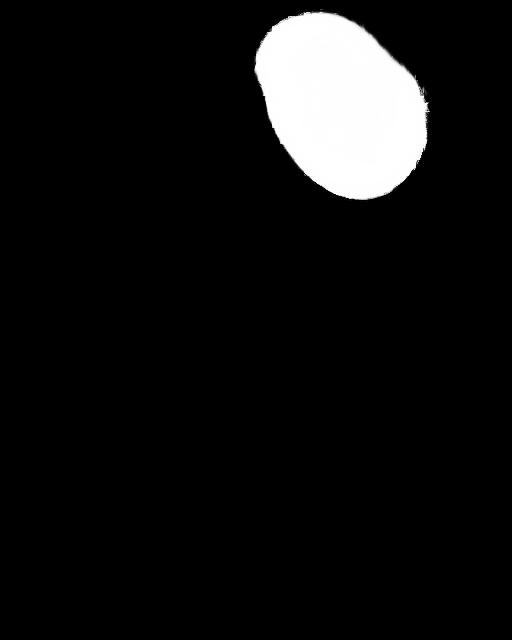}&
    \includegraphics[width=\wvisual, height=\hvisualtall]{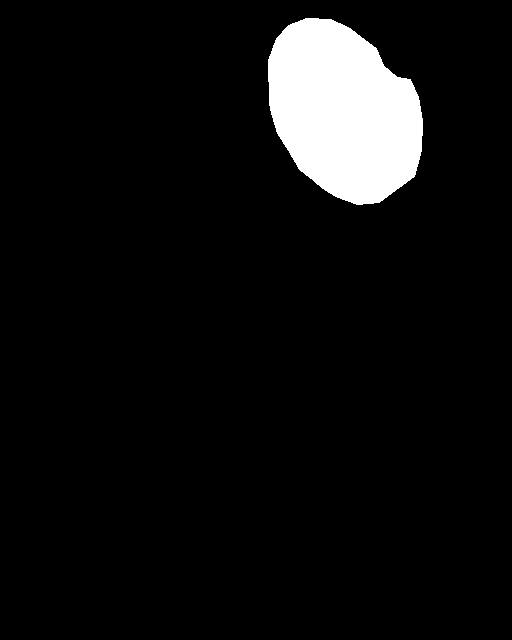} \\

    \includegraphics[width=\wvisual, height=\hvisualtall]{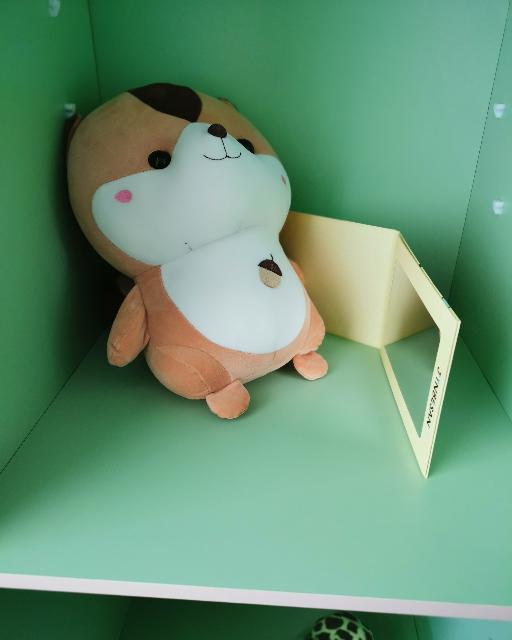}&
    \includegraphics[width=\wvisual, height=\hvisualtall]{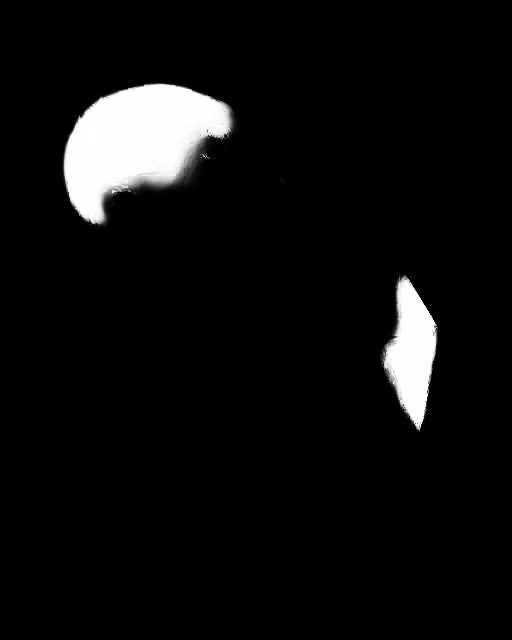}&
    \includegraphics[width=\wvisual, height=\hvisualtall]{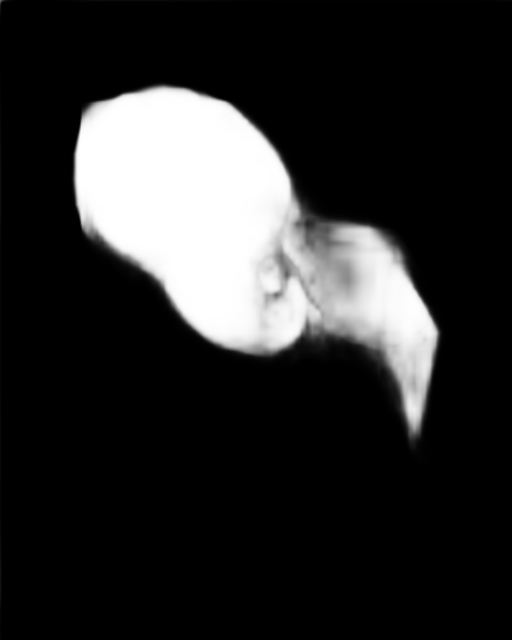}&
    \includegraphics[width=\wvisual, height=\hvisualtall]{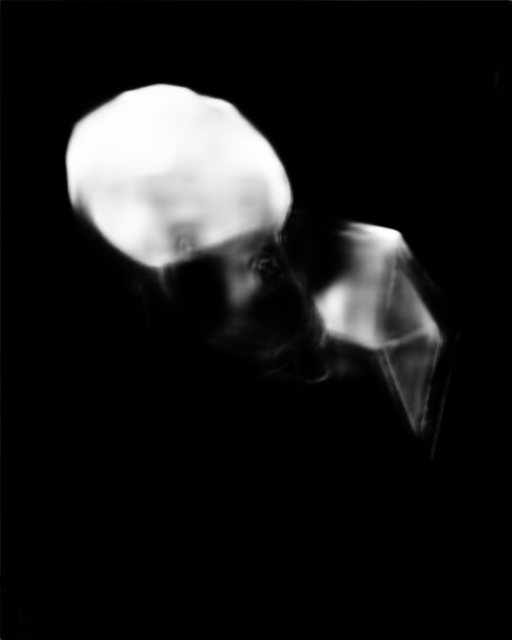}&
    \includegraphics[width=\wvisual, height=\hvisualtall]{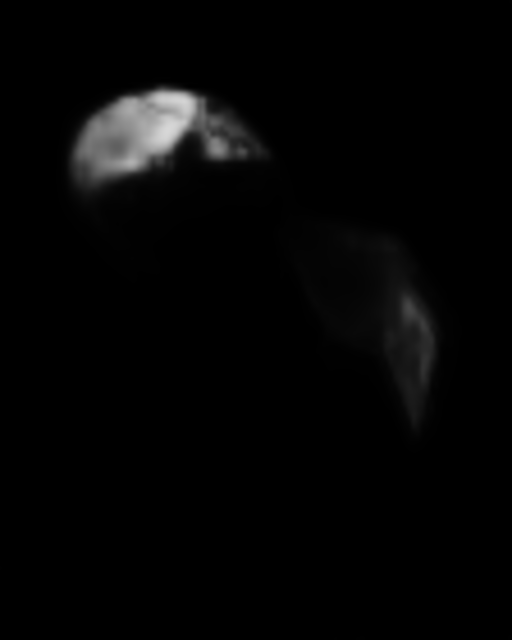}&
    \includegraphics[width=\wvisual, height=\hvisualtall]{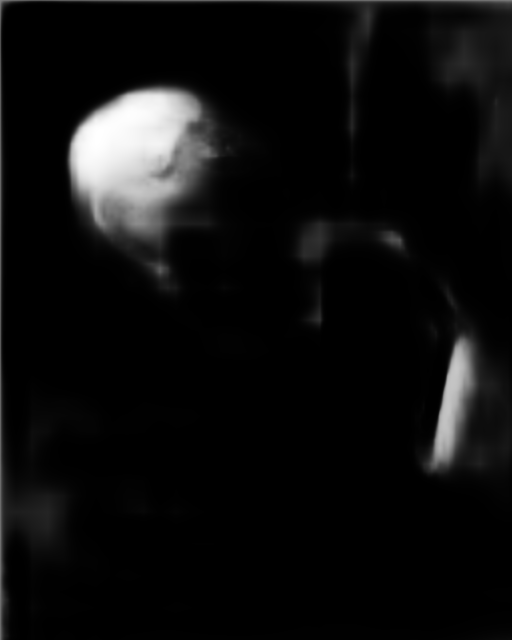}&
    \includegraphics[width=\wvisual, height=\hvisualtall]{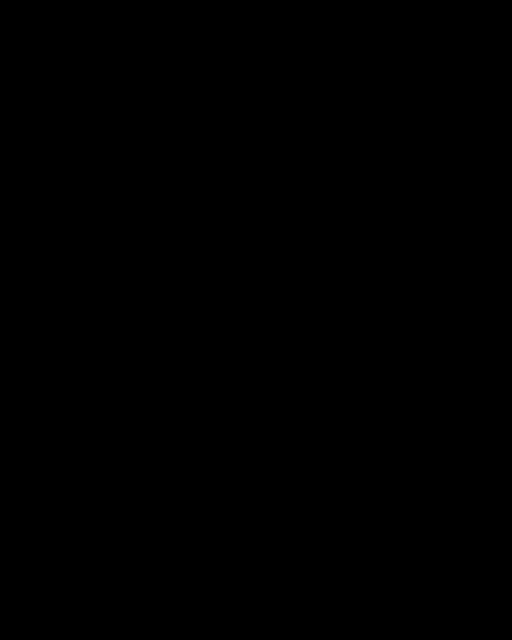}&
    \includegraphics[width=\wvisual, height=\hvisualtall]{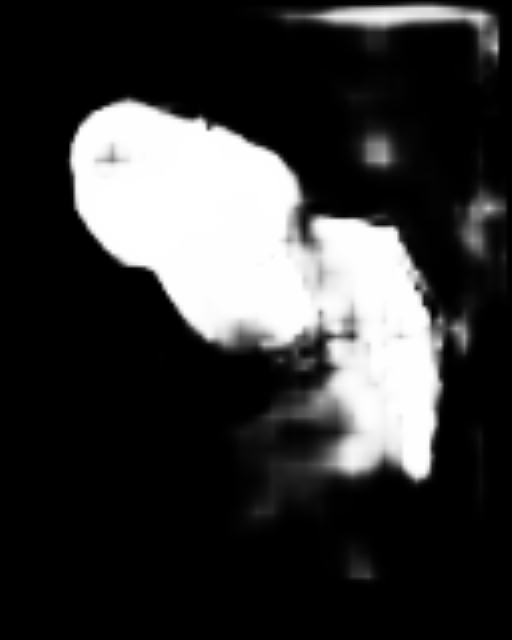}&
    \includegraphics[width=\wvisual, height=\hvisualtall]{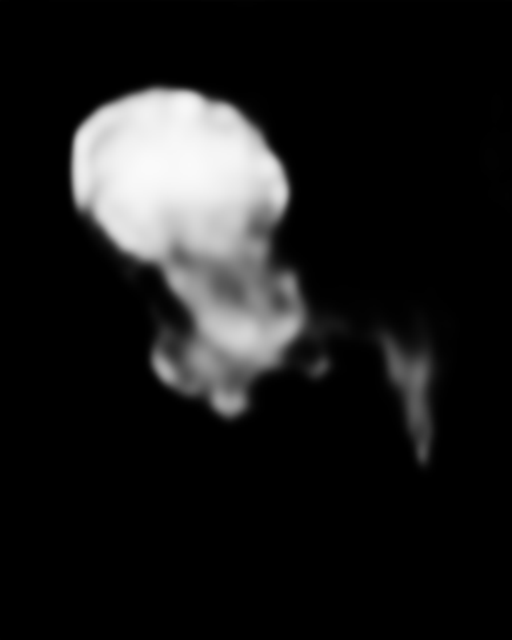}&
    \includegraphics[width=\wvisual, height=\hvisualtall]{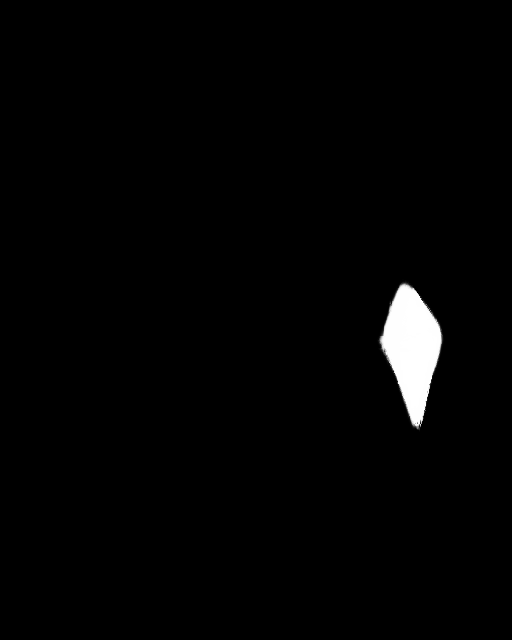}&
    \includegraphics[width=\wvisual, height=\hvisualtall]{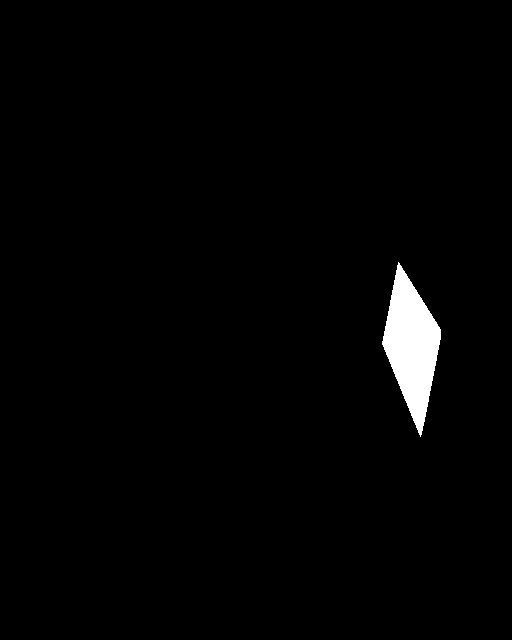} \\

    \includegraphics[width=\wvisual, height=\hvisualtall]{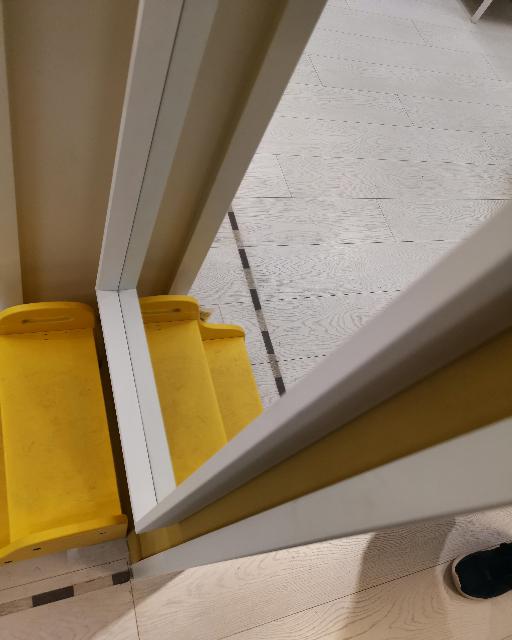}&
    \includegraphics[width=\wvisual, height=\hvisualtall]{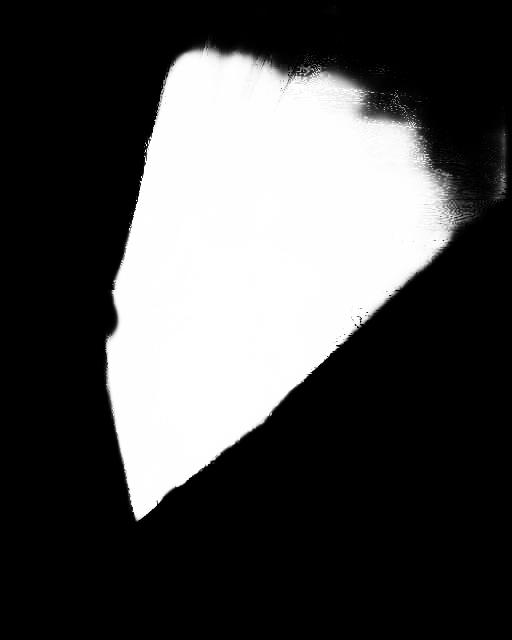}&
    \includegraphics[width=\wvisual, height=\hvisualtall]{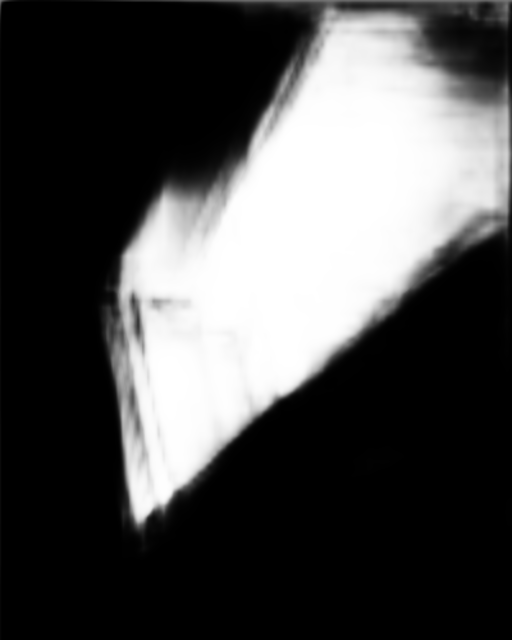}&
    \includegraphics[width=\wvisual, height=\hvisualtall]{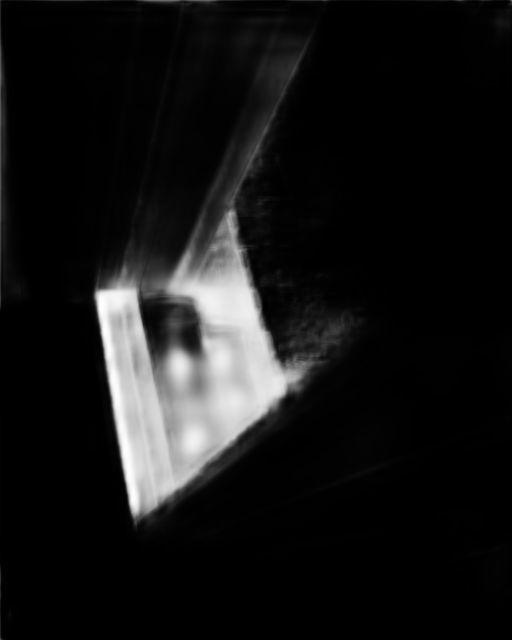}&
    \includegraphics[width=\wvisual, height=\hvisualtall]{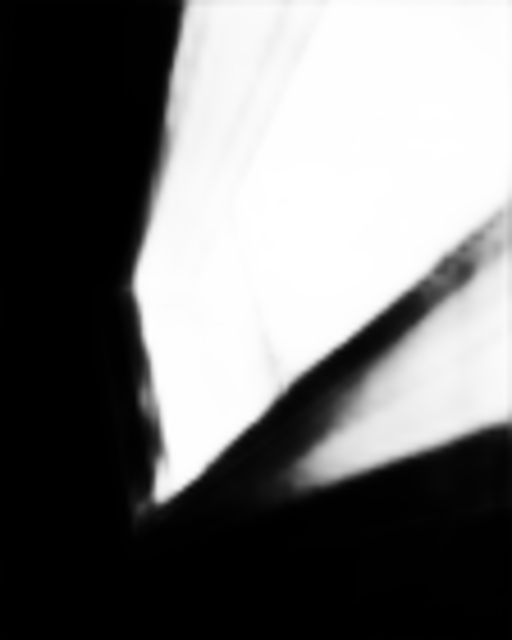}&
    \includegraphics[width=\wvisual, height=\hvisualtall]{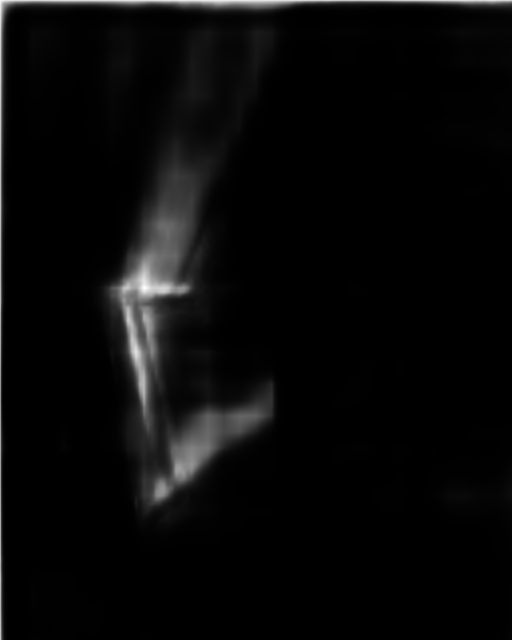}&
    \includegraphics[width=\wvisual, height=\hvisualtall]{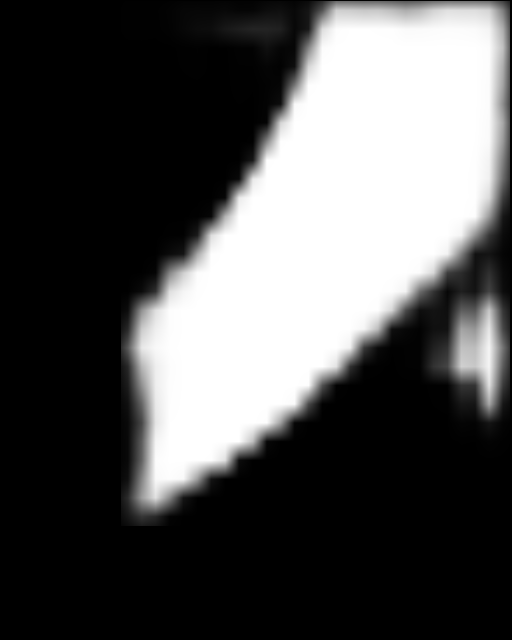}&
    \includegraphics[width=\wvisual, height=\hvisualtall]{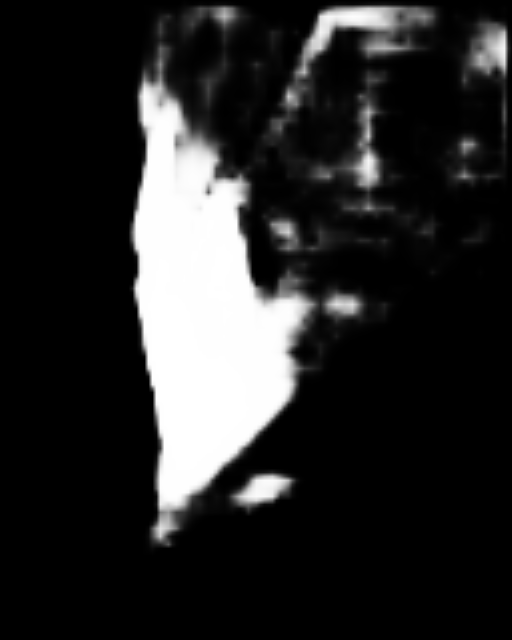}&
    \includegraphics[width=\wvisual, height=\hvisualtall]{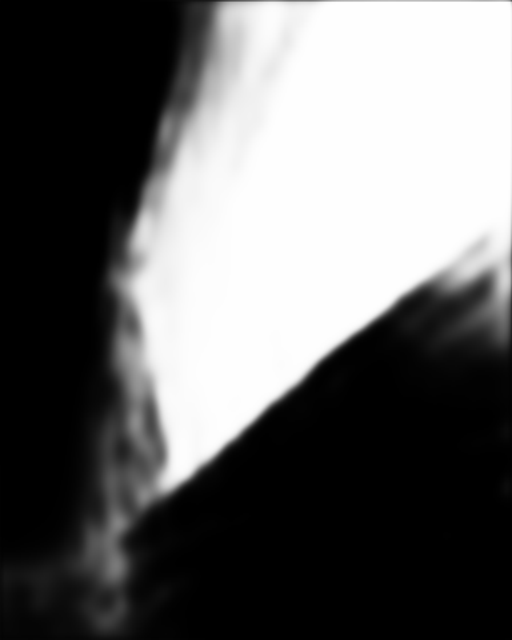}&
    \includegraphics[width=\wvisual, height=\hvisualtall]{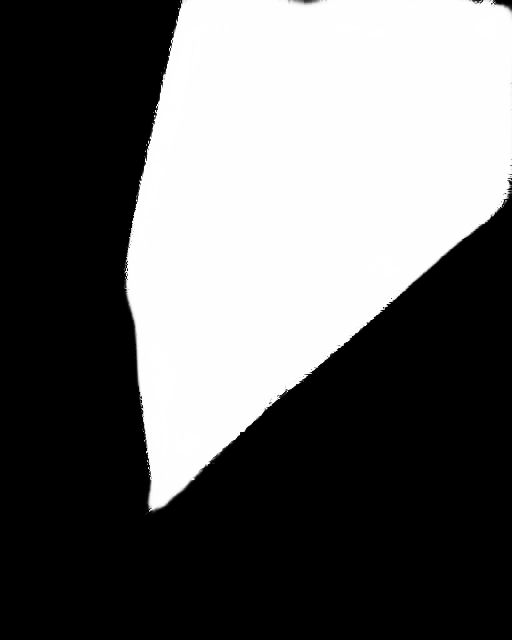}&
    \includegraphics[width=\wvisual, height=\hvisualtall]{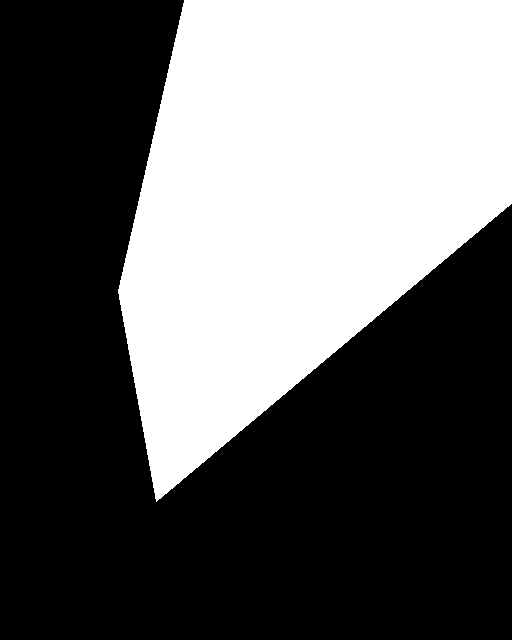} \\

    \includegraphics[width=\wvisual, height=\hvisualtall]{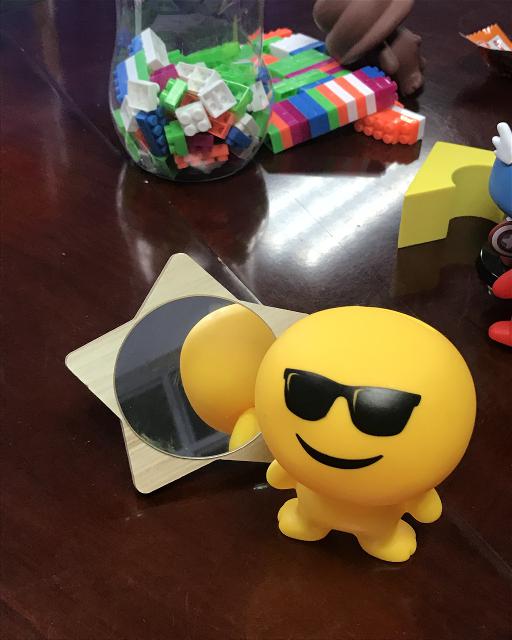}&
    \includegraphics[width=\wvisual, height=\hvisualtall]{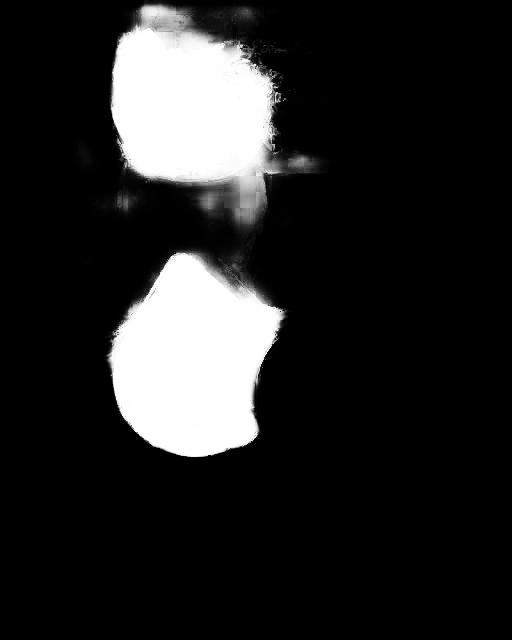}&
    \includegraphics[width=\wvisual, height=\hvisualtall]{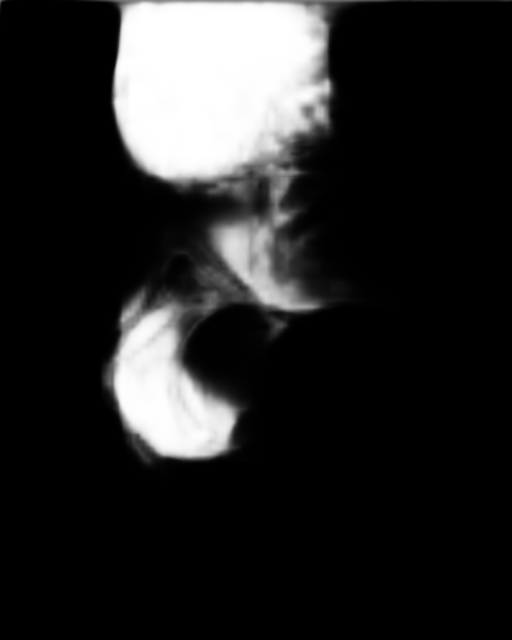}&
    \includegraphics[width=\wvisual, height=\hvisualtall]{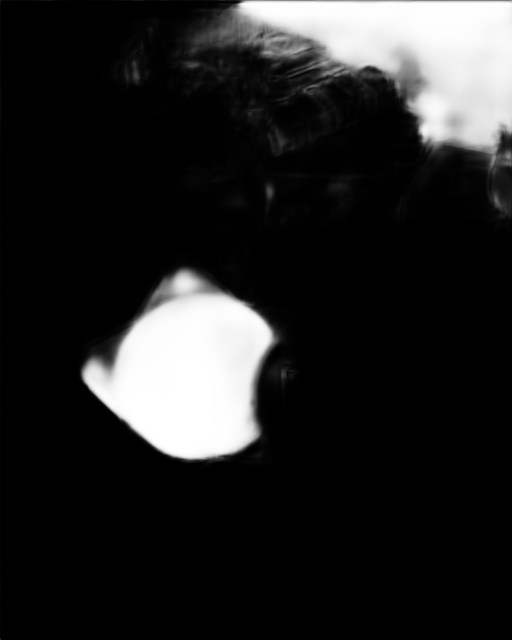}&
    \includegraphics[width=\wvisual, height=\hvisualtall]{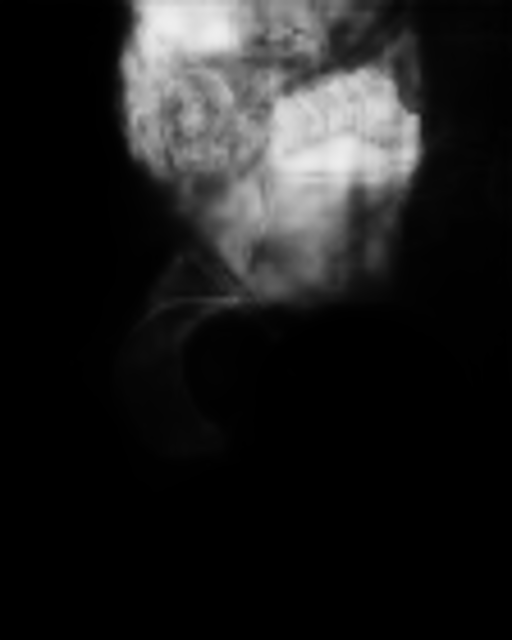}&
    \includegraphics[width=\wvisual, height=\hvisualtall]{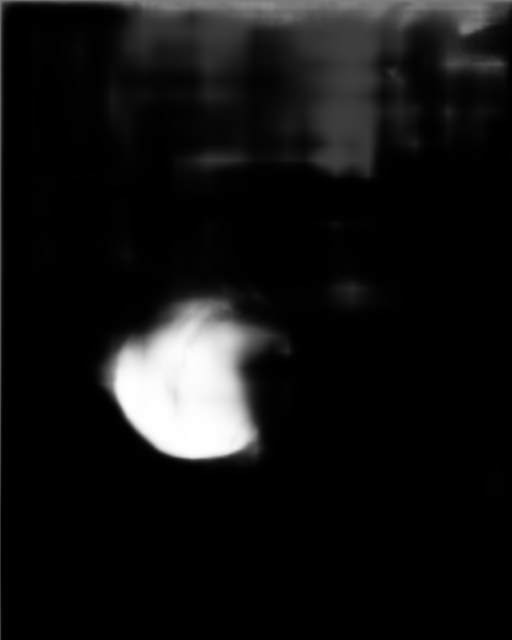}&
    \includegraphics[width=\wvisual, height=\hvisualtall]{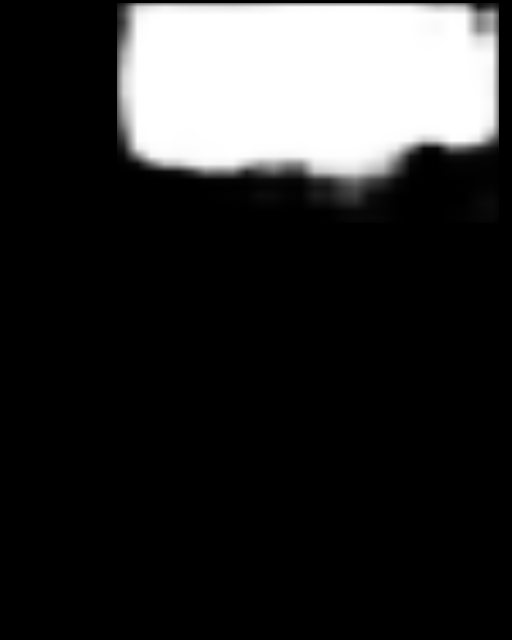}&
    \includegraphics[width=\wvisual, height=\hvisualtall]{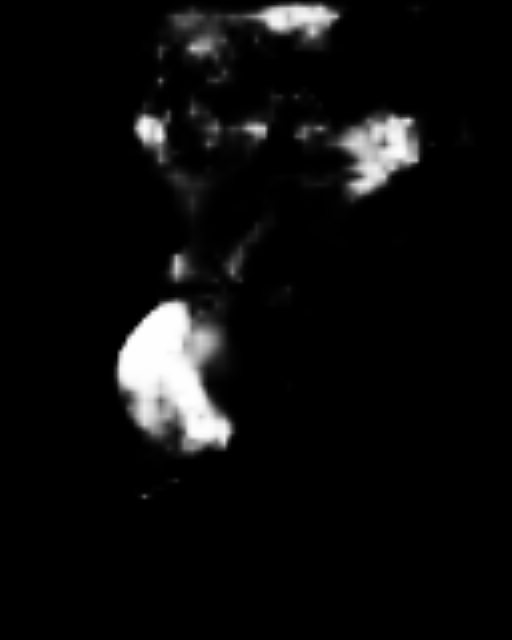}&
    \includegraphics[width=\wvisual, height=\hvisualtall]{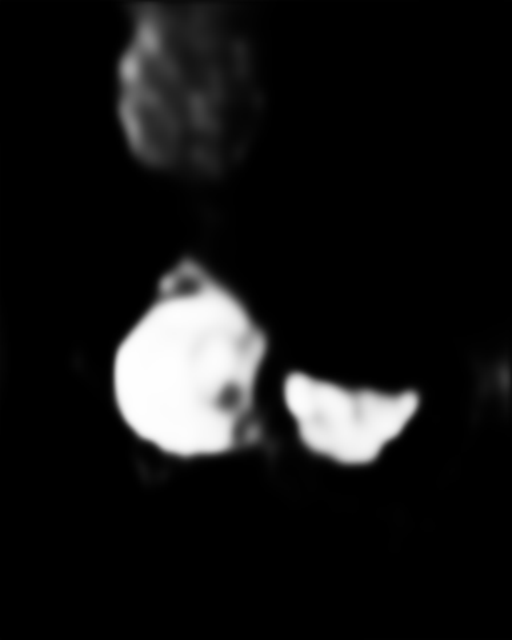}&
    \includegraphics[width=\wvisual, height=\hvisualtall]{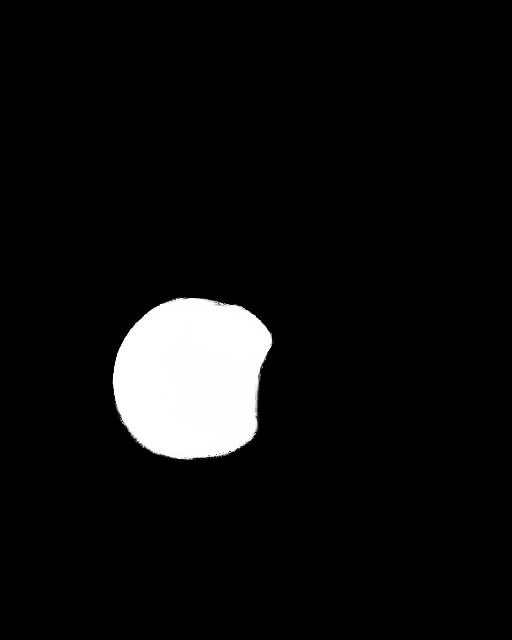}&
    \includegraphics[width=\wvisual, height=\hvisualtall]{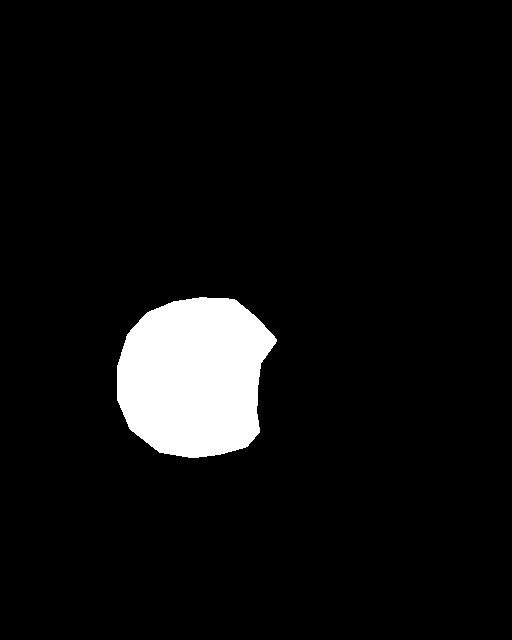} \\

    \footnotesize{Image}  & \footnotesize{BDRAR\cite{Zhu_2018_ECCV}} & \footnotesize{DSC\cite{Hu_2018_CVPR}} & \footnotesize{RAS\cite{Chen_2018_ECCV}} & \footnotesize{PiCANet\cite{liu2018picanet}} & \footnotesize{DSS\cite{Hou_2017_CVPR}} & {\scriptsize Mask RCNN\cite{he2017mask}} & \footnotesize{ICNet\cite{Zhao_2018_ECCV}} &
    \footnotesize{PSPNet\cite{zhao2017pyramid}}  & \footnotesize{MirrorNet}  & \footnotesize{GT} \\
    \end{tabular}
    \vspace{-3mm}
  \caption{Visual comparison of MirrorNet to the state-of-the-art methods on the proposed MSD test set.}
  \label{fig:visual}\vspace{-5mm}
\end{figure*}

\subsection{Comparison to the State-of-the-arts}

{\bf Evaluation on the MSD test set.} Table \ref{tab:comparison} reports the mirror segmentation performance on the proposed MSD test set. We can see that our method achieves the best performance with a large margin on all five metrics: \rf{intersection over union} (IoU), \rf{pixel accuracy} (Acc), F-measure ($F_\beta$), mean absolute \rf{error} (MAE), and balance error rate (BER).
Figure~\ref{fig:visual} shows visual comparisons. We can see that our method can effectively locate and segment small mirrors (4\emph{th}, 5\emph{th} and 7\emph{th} rows). While the state-of-the-arts typically under-segment the large mirrors with high contrasts among their contents, our method successfully detects the mirror regions as a whole (\eg, 1\emph{st} and 3\emph{rd} rows). Our method can also accurately \rf{delineate} the mirror region boundaries, where there are ambiguities caused by nearby objects and their reflections in the mirror (2\emph{nd} row). In general, our method can segment mirrors of different sizes with accurate boundaries. This is mainly contributed by the proposed multi-scale contextual contrasted feature learning.

\def\wade20k{0.166\linewidth}
\def\hade20k{.45in}
\begin{figure}[tb]
\setlength{\tabcolsep}{0.4pt}
  \centering
  \begin{tabular}{ccccccc}
    \includegraphics[width=\wade20k, height=\hade20k]{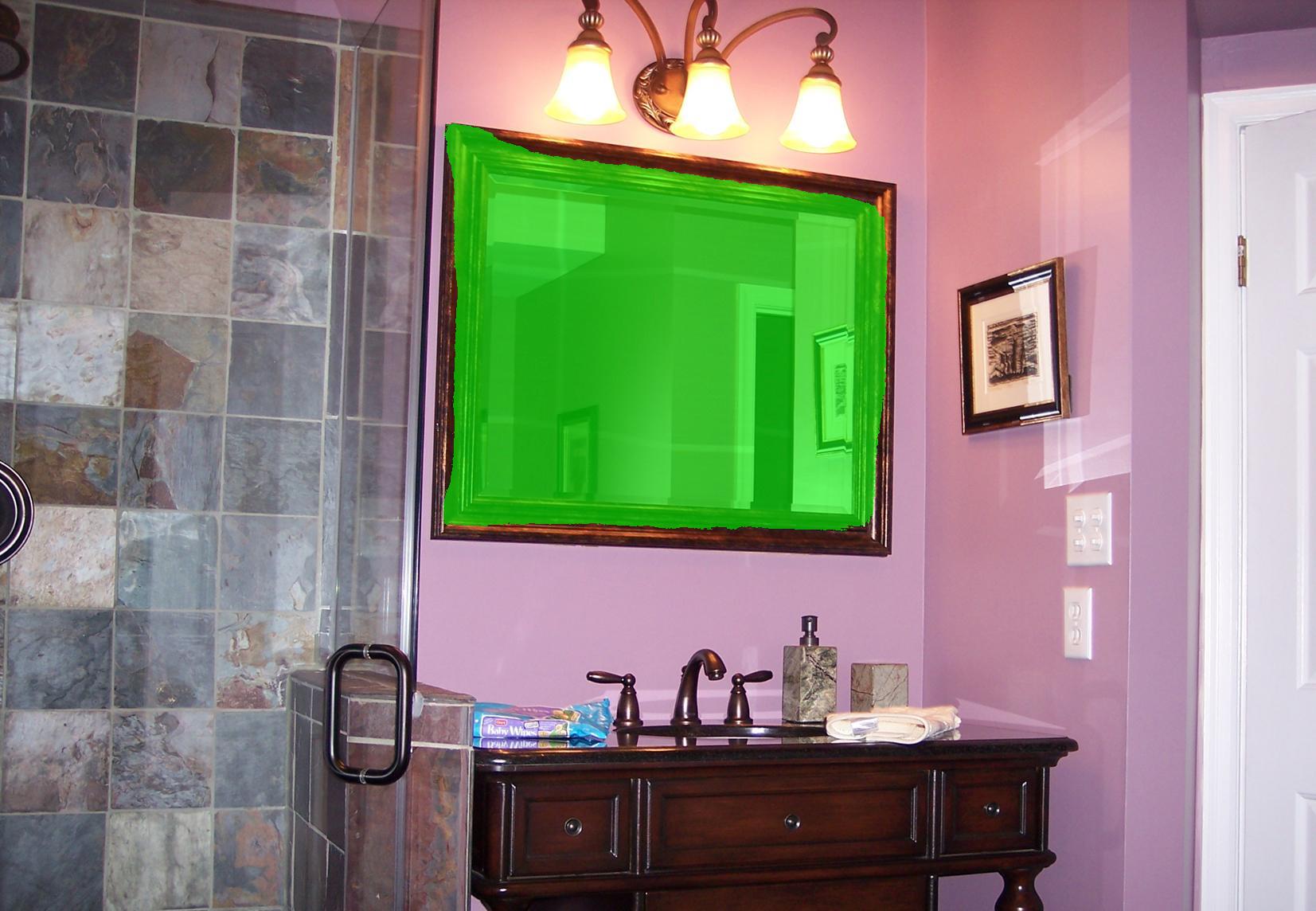}&
    \includegraphics[width=\wade20k, height=\hade20k]{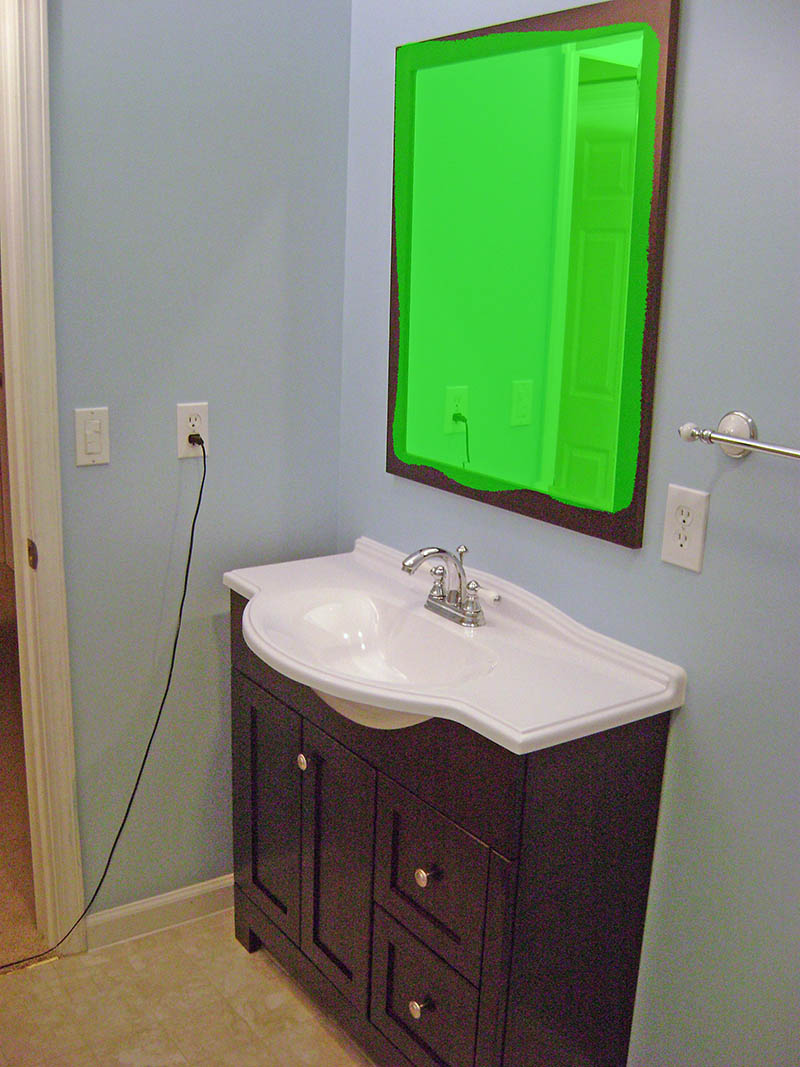}&
    \includegraphics[width=\wade20k, height=\hade20k]{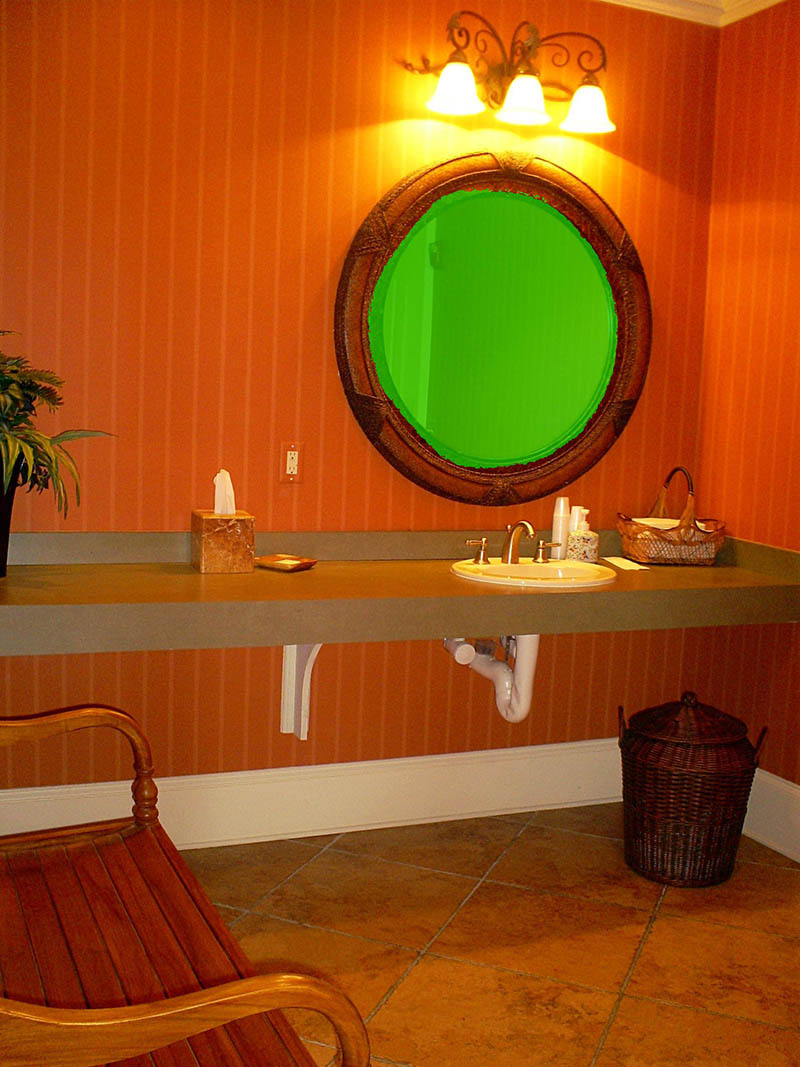}&
    \includegraphics[width=\wade20k, height=\hade20k]{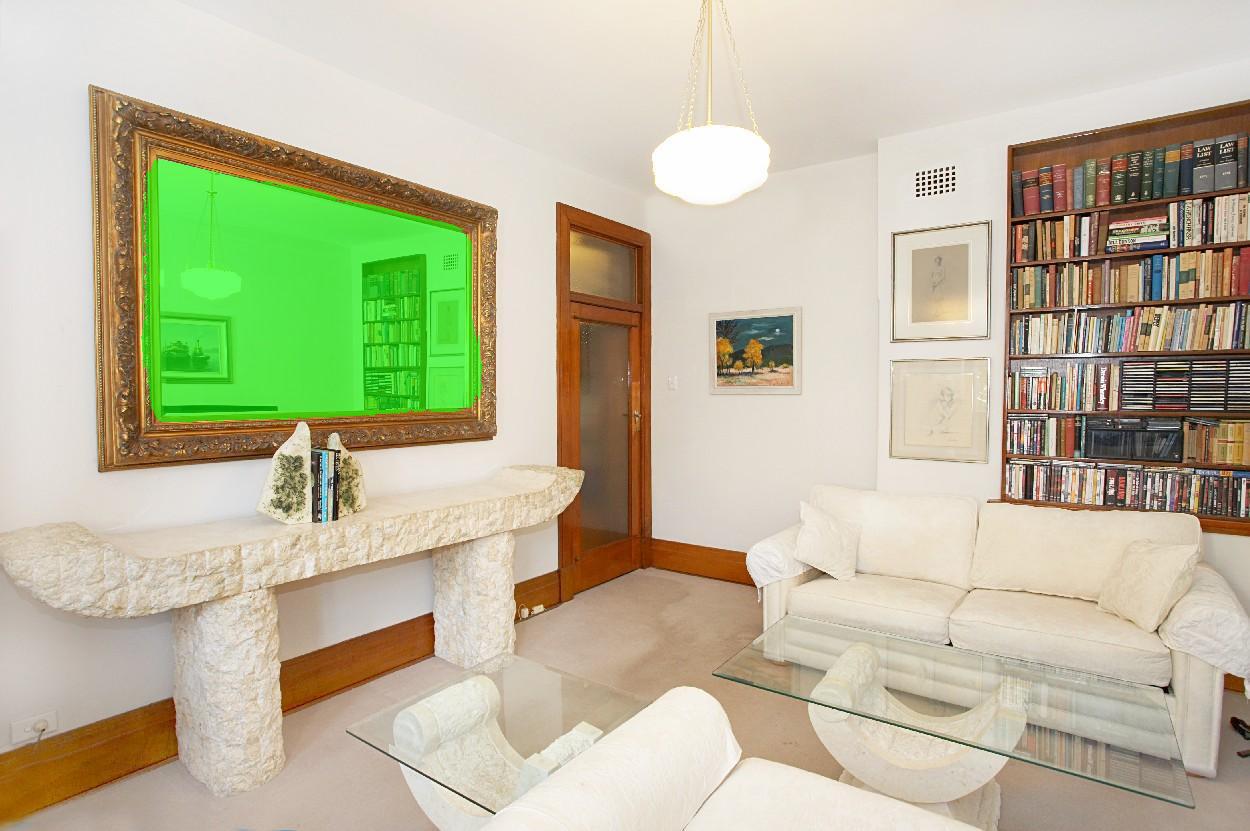}&
    \includegraphics[width=\wade20k, height=\hade20k]{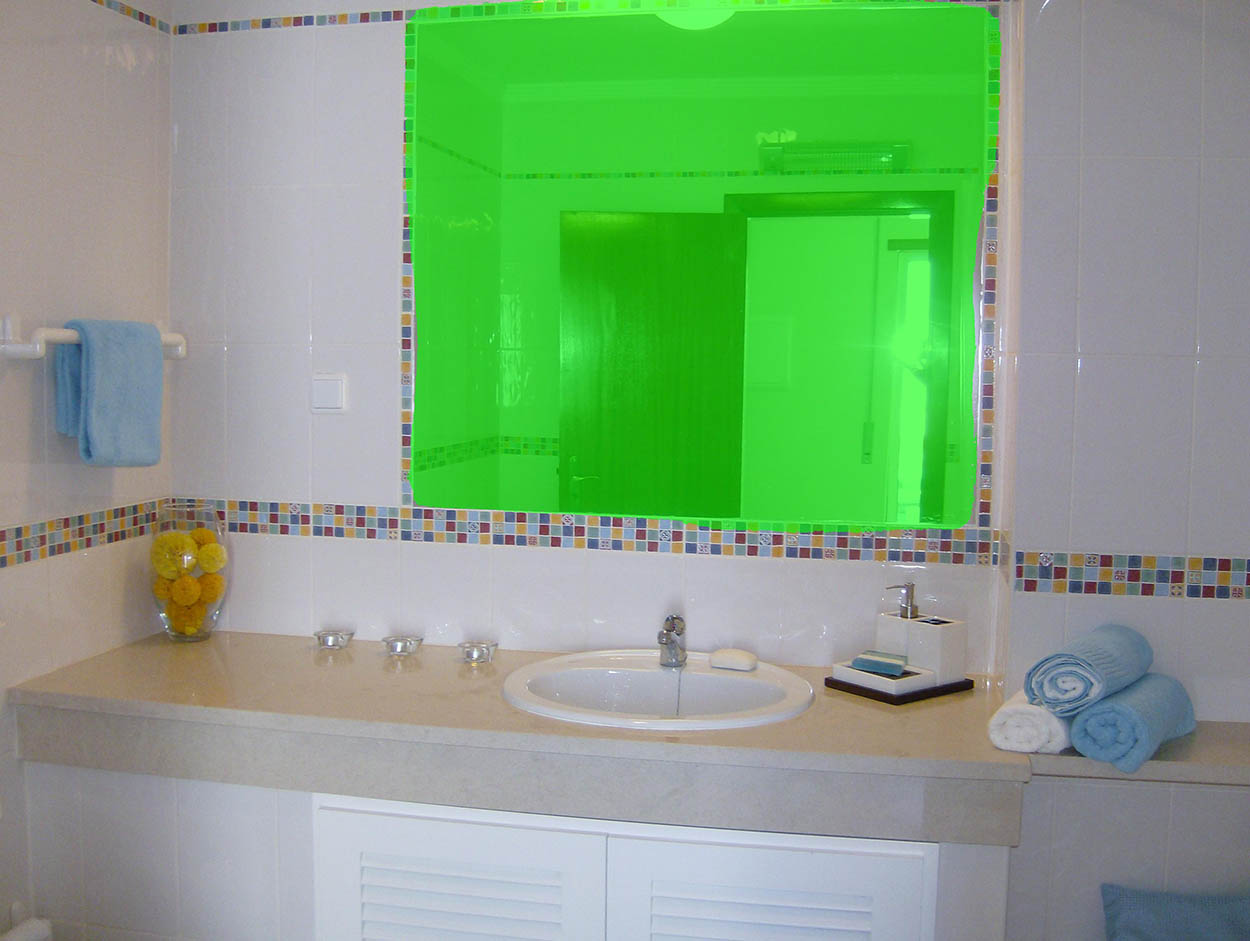}&
    \includegraphics[width=\wade20k, height=\hade20k]{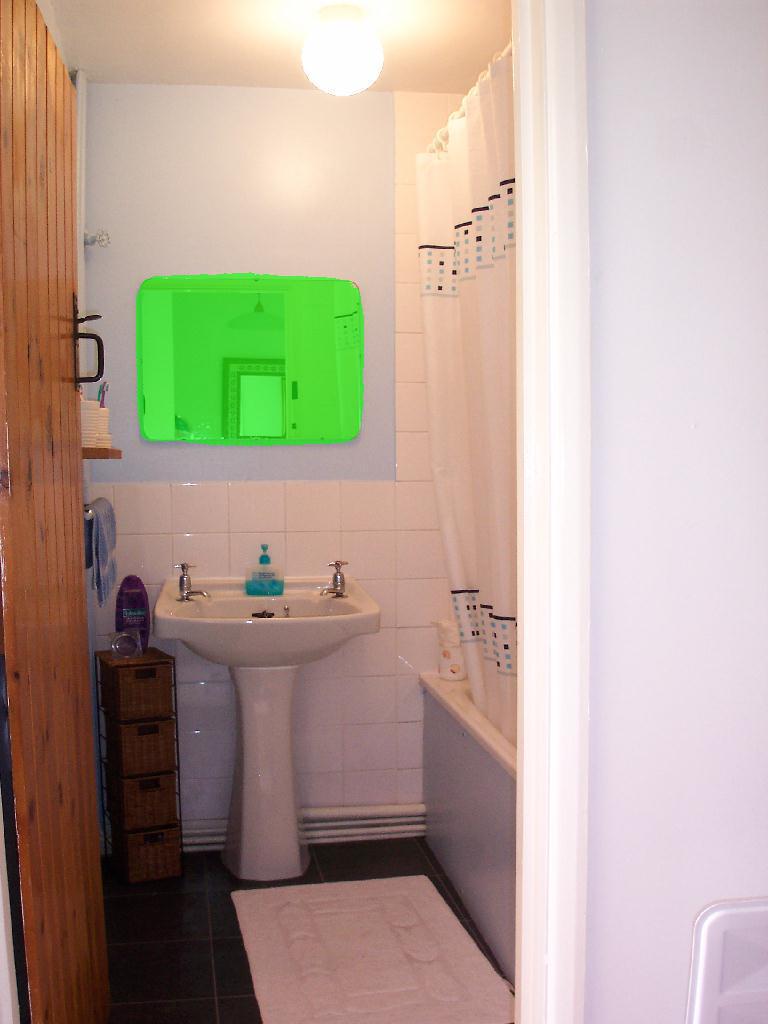}\\

    \includegraphics[width=\wade20k, height=\hade20k]{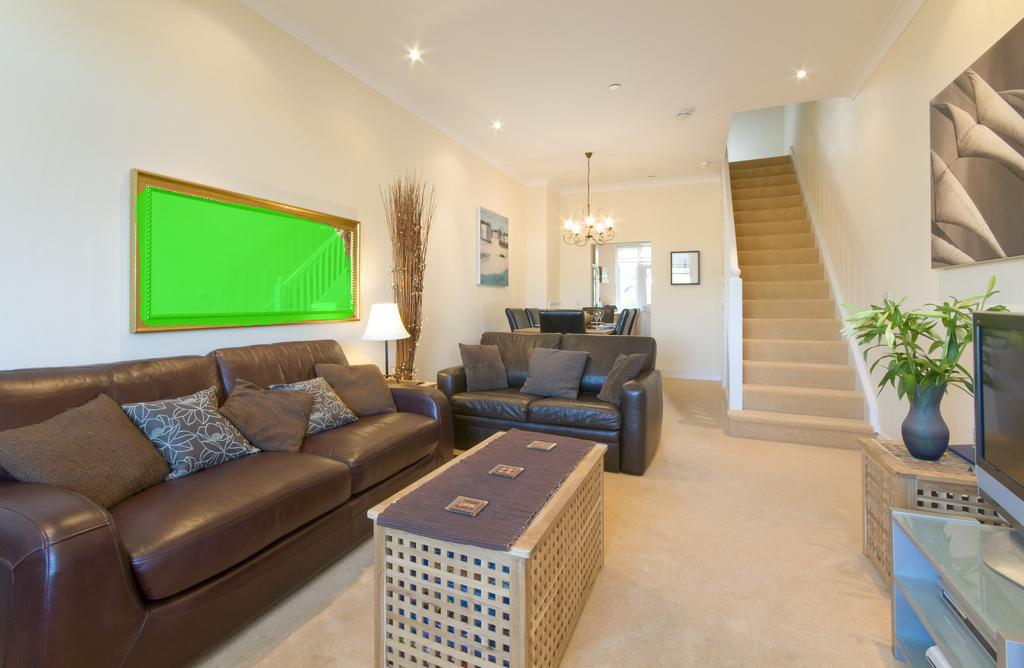}&
    \includegraphics[width=\wade20k, height=\hade20k]{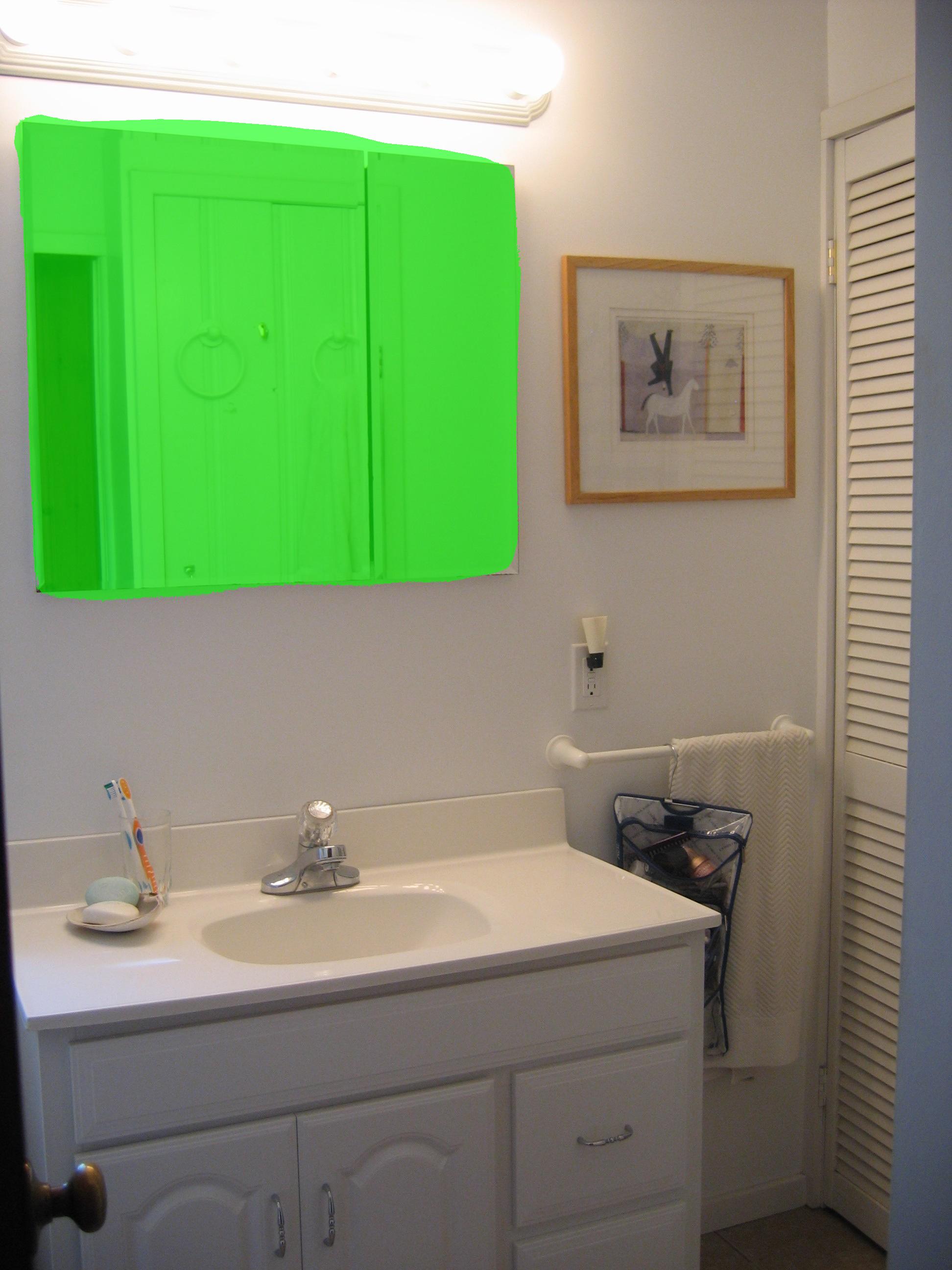}&
    \includegraphics[width=\wade20k, height=\hade20k]{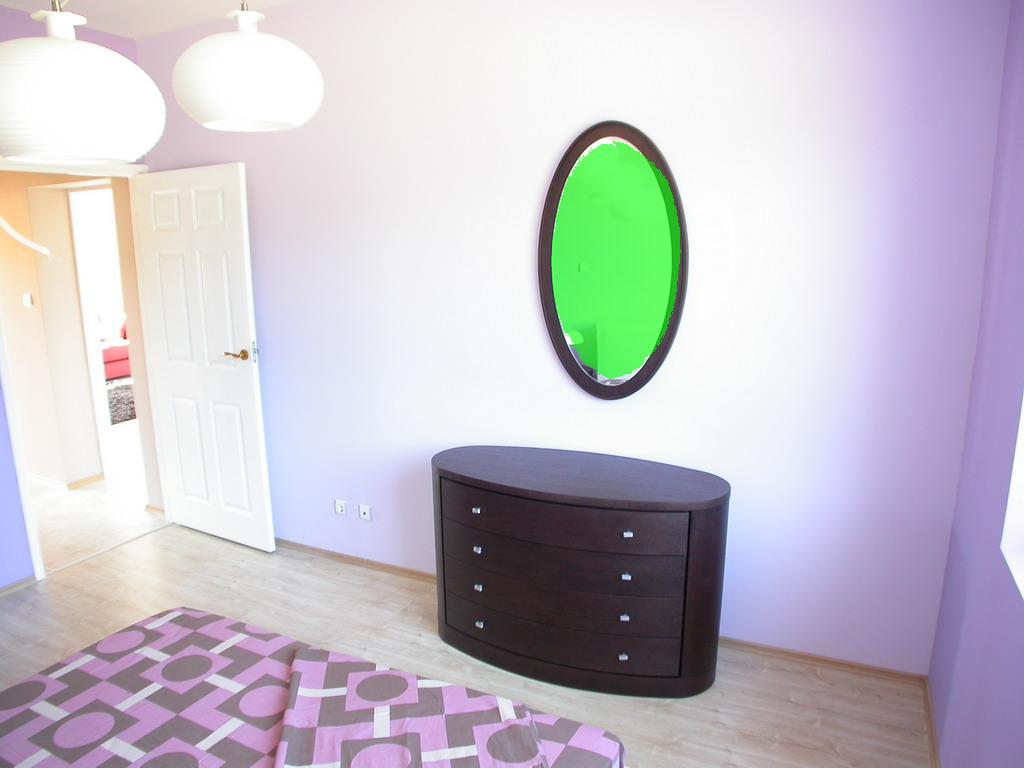}&
    \includegraphics[width=\wade20k, height=\hade20k]{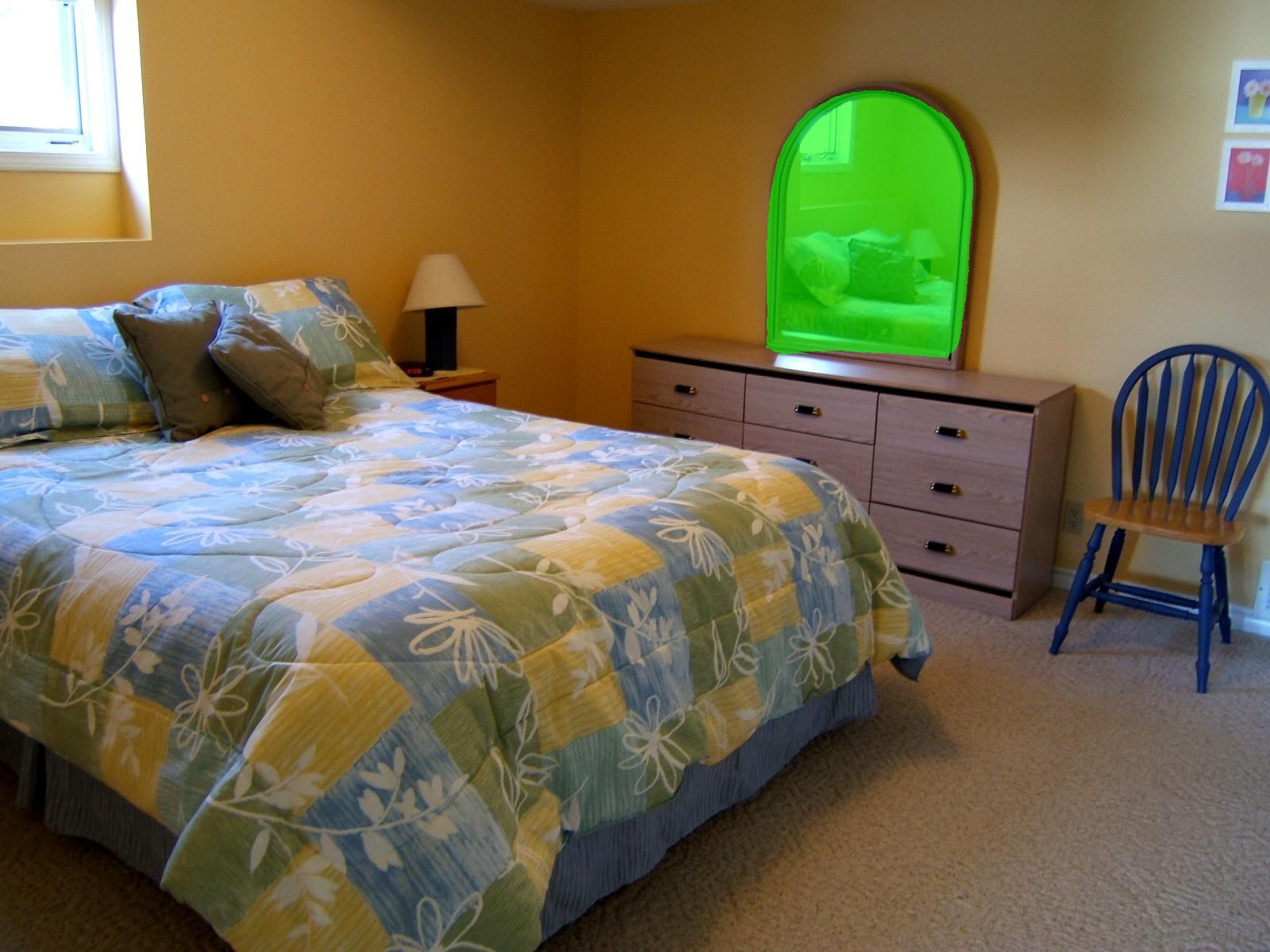}&
    \includegraphics[width=\wade20k, height=\hade20k]{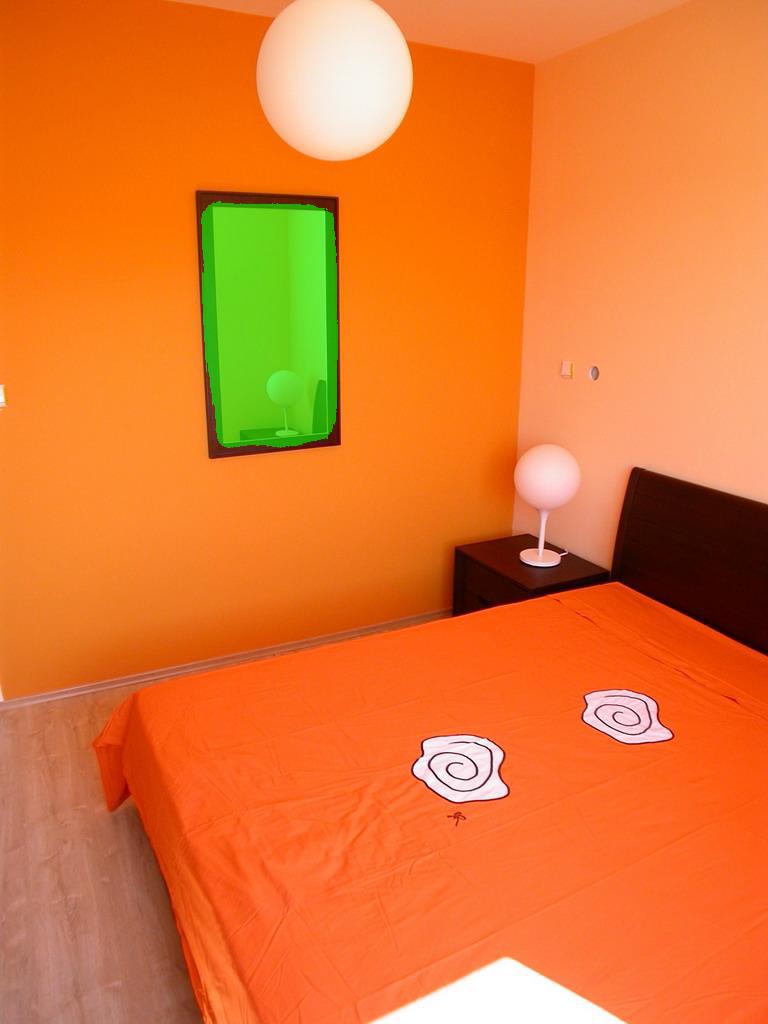}&
    \includegraphics[width=\wade20k, height=\hade20k]{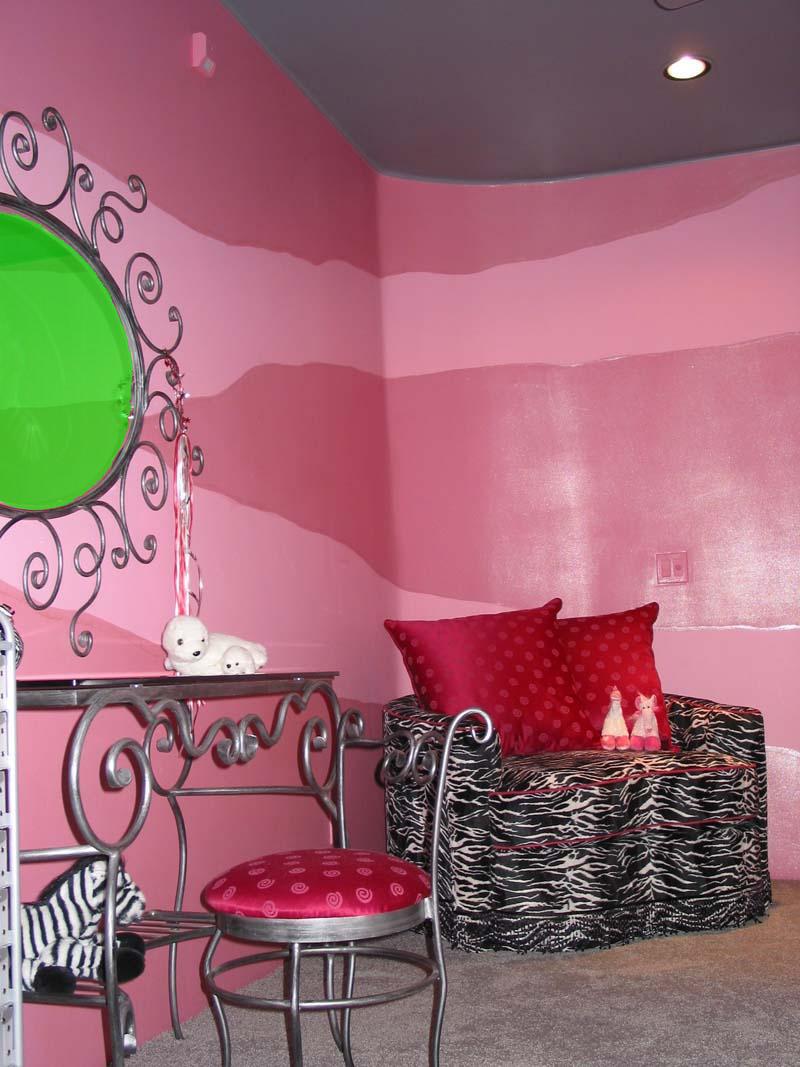}\\
    \\\vspace{-12mm}
    \end{tabular}
  \caption{Some mirror segmentation results of MirrorNet on the ADE20K dataset~\cite{Zhou_2017_CVPR}.}
  \label{fig:ade20k}
  \vspace{-5mm}
\end{figure}

{\bf More mirror segmentation results.}
Figure \ref{fig:ade20k} shows some mirror segmentation results from our MirrorNet on the ADE20K dataset~\cite{Zhou_2017_CVPR}, which demonstrate the effectiveness of MirrorNet.
Figure \ref{fig:more} shows mirror segmentation results of some challenging images downloaded from the Internet. These images contain not only mirrors but also other mirror-like objects, such as paintings (2\emph{nd}, 3\emph{rd} and 6\emph{th} rows), windows (5\emph{th} row), and door (4\emph{th} row).
%
We can see that the existing methods are distracted by these mirror-like objects. However, MirrorNet can distinguish mirrors from paintings/windows (\eg, 2\emph{nd}, 3\emph{rd} and 5\emph{th} rows), as the content within a mirror region is usually semantically consistent with the rest of the image while the content within a painting/window region is often different.
MirrorNet is designed to learn different levels of contextual contrast features between the mirror region and outside.
For example, the mirror region in the 2\emph{nd} row reflects the indoor scene, which is similar to the surroundings of the mirror, while the paintings contain very different scenes. Such differences can be learned by the CCFE module. We understand that there are limitations with this assumption, which can be an interesting future work. In addition, MirrorNet can distinguish the mirror from the door in the 4\emph{th} row. A possible reason is that the bottom of the door region is continuous and thus the door region is not considered as a mirror.

\begin{table}[tbp]
  \begin{small}
  \centering
    \begin{tabular}{p{4cm}|p{1cm}<{\centering}p{0.9cm}<{\centering}p{0.8cm}<{\centering}}
    \hline
      Networks & IoU$\uparrow$ & Acc$\uparrow$ & BER$\downarrow$ \\
    \hline
    basic + BCE loss                                               & 74.03 & 0.821 & 10.60 \\
    basic + lov\'{a}sz-hinge loss \cite{Berman_2018_CVPR}          & 75.36 & 0.820 & 10.44 \\
    basic + CCFE w/o contrasts                                     & 78.59 & 0.851 & 8.55 \\
    basic + CCFE w/ 1B4C                                           & 76.36 & 0.882 & 8.01 \\
    basic + CCFE w/ 4B1C                                           & 78.53 & 0.853 & 9.08 \\
    MirrorNet                                                           & \bf{78.95} & \bf{0.933} & \bf{6.39} \\
    \hline
    \end{tabular}
    \end{small}
    \vspace{-3mm}
      \caption{Component analysis. ``basic'' denotes our network with all CCFE modules removed, ``CCFE w/o contrasts'' denotes using multi-scale dilated convolutions without computing their feature \rf{contrasts}. ``1B4C'' denotes using 1 CCFE block with 4 parallel scales of contrasts, while ``4B1C'' denotes using 4 CCFE blocks with 1 scale of contrasts. Our proposed CCFE module contains 4 blocks and each of them contains 4 scales of contrast extraction.}
  \label{tab:ablation}
  \vspace{-2mm}
\end{table}

\def\wmore{0.19\linewidth}
\def\hmoreone{.3in}
\def\hmoretwo{.5in}
\def\hmorethree{.35in}
\def\hmorefour{.4in}
\def\hmorefive{.45in}
\def\hmoresix{.45in}
\def\hmoreseven{.5in}
\begin{figure}[tbp]
\setlength{\tabcolsep}{1.6pt}
  \centering
  \begin{tabular}{ccccc}
    \includegraphics[width=\wmore, height=\hmoreone]{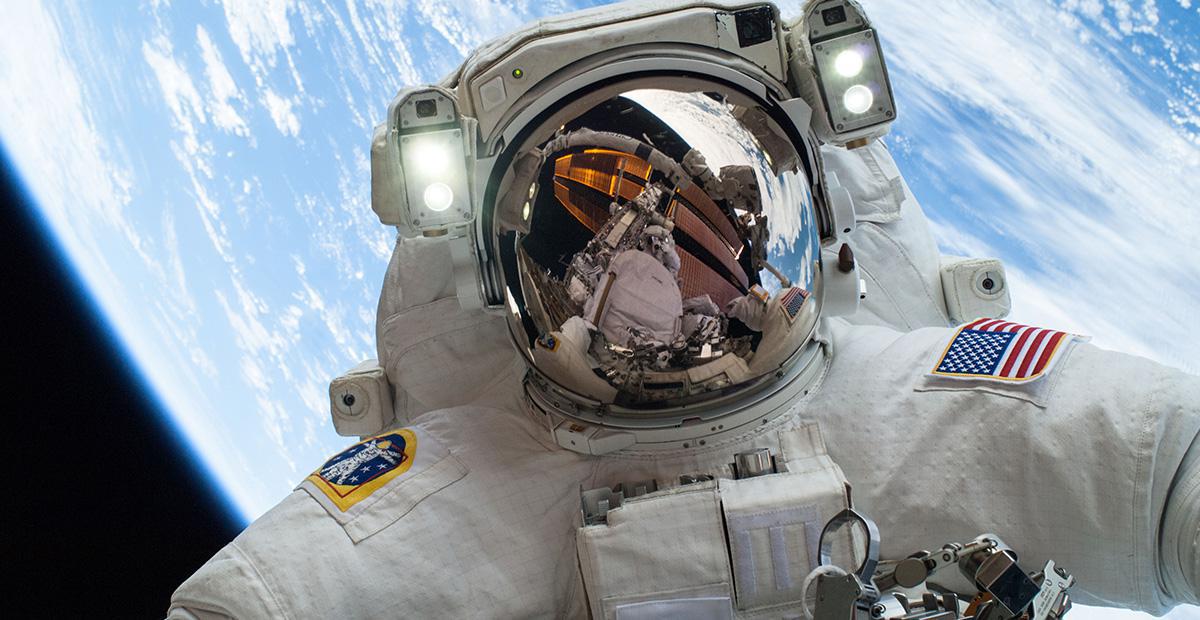}&
    \includegraphics[width=\wmore, height=\hmoreone]{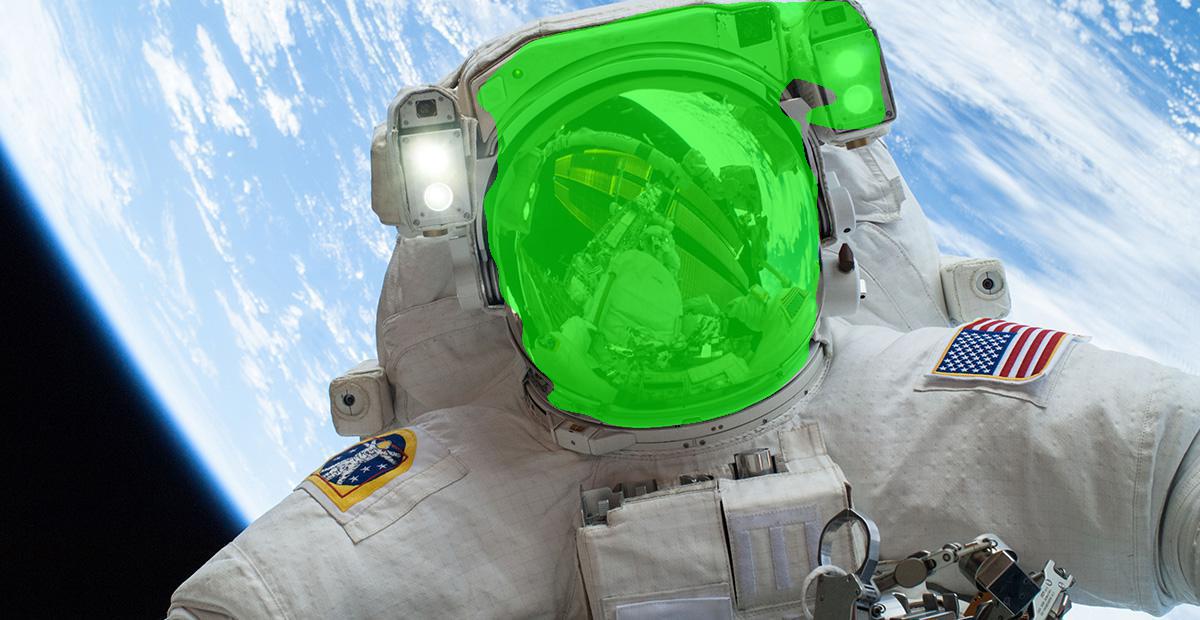}&
    \includegraphics[width=\wmore, height=\hmoreone]{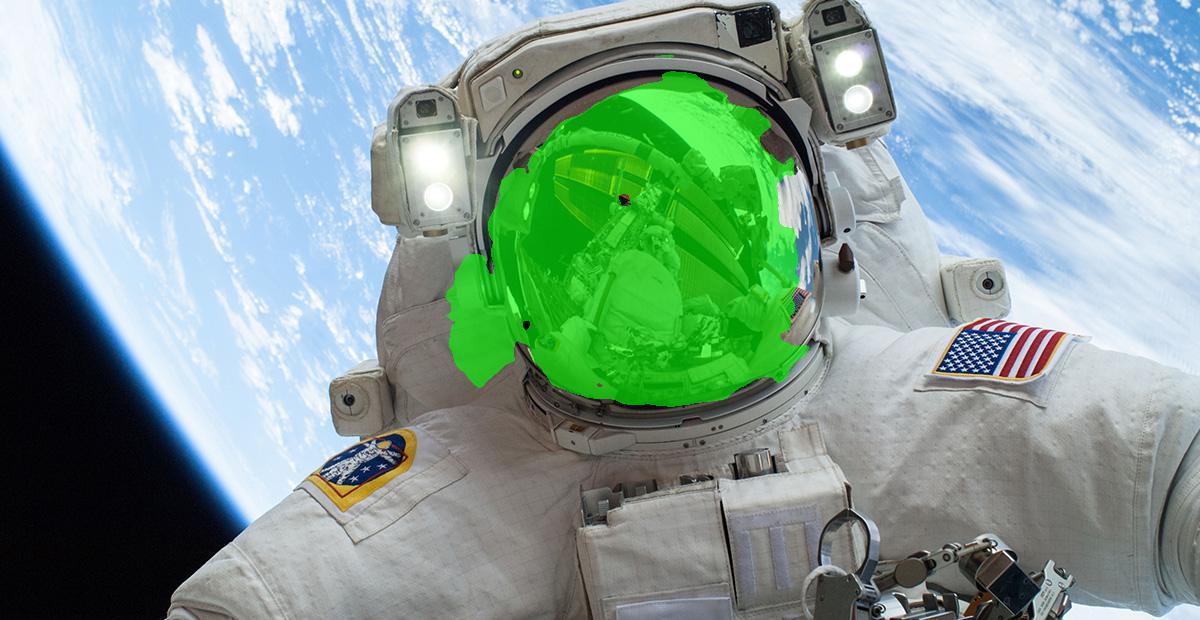}&
    \includegraphics[width=\wmore, height=\hmoreone]{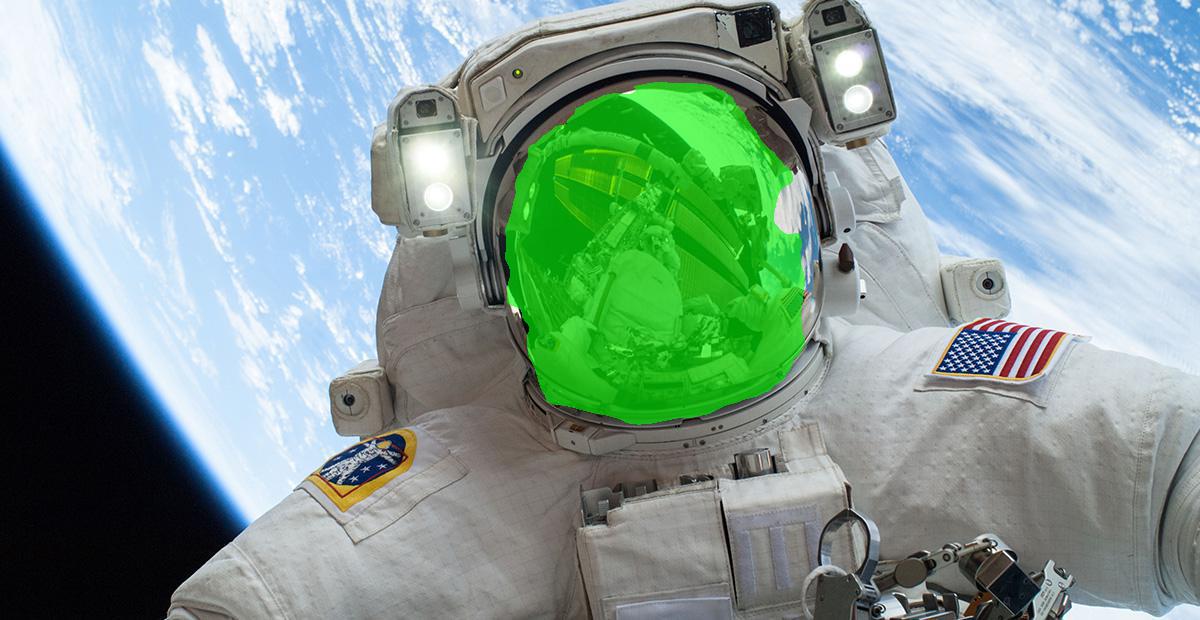}&
    \includegraphics[width=\wmore, height=\hmoreone]{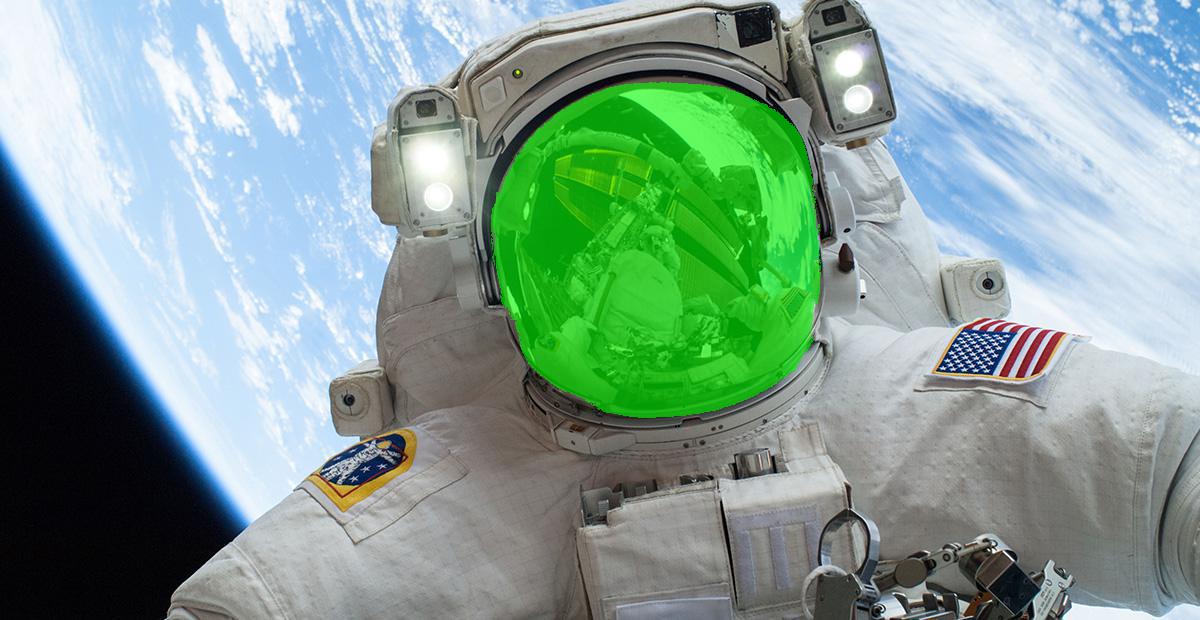} \\

    \includegraphics[width=\wmore, height=\hmorethree]{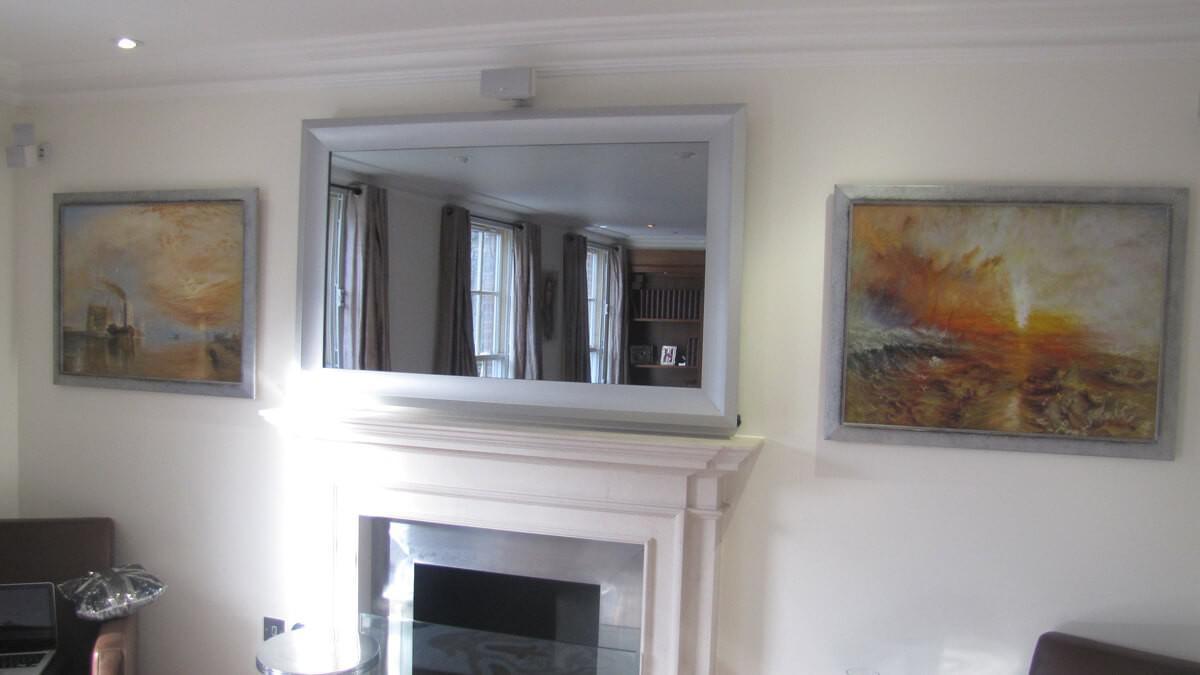}&
    \includegraphics[width=\wmore, height=\hmorethree]{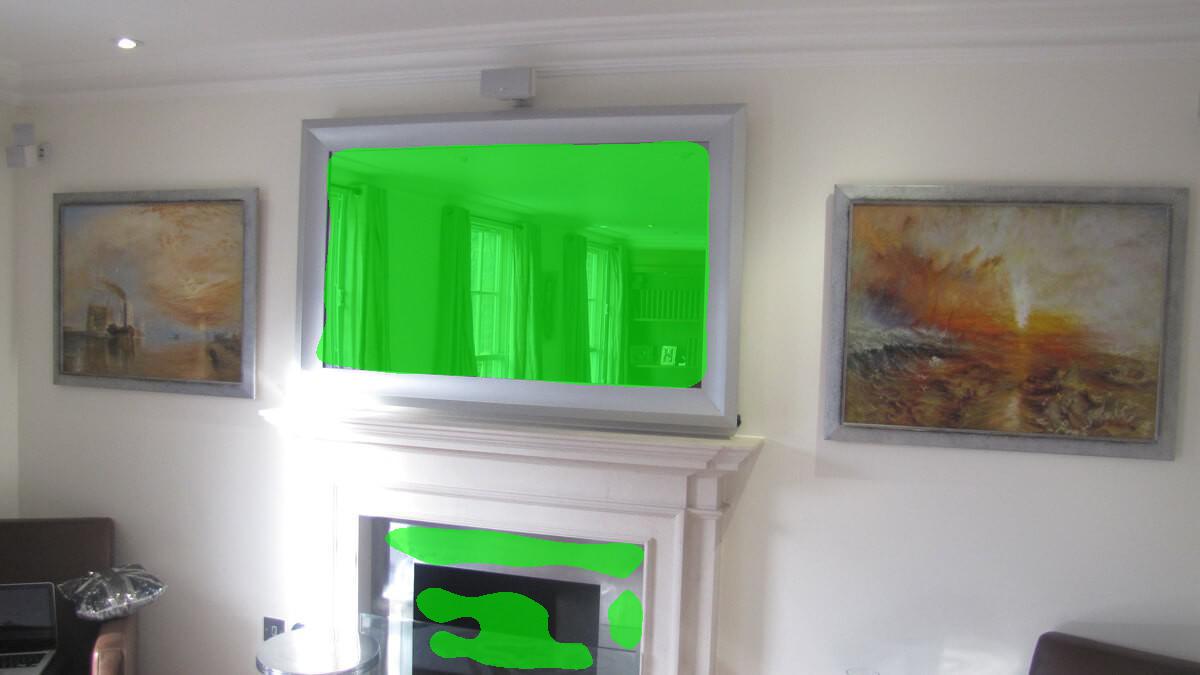}&
    \includegraphics[width=\wmore, height=\hmorethree]{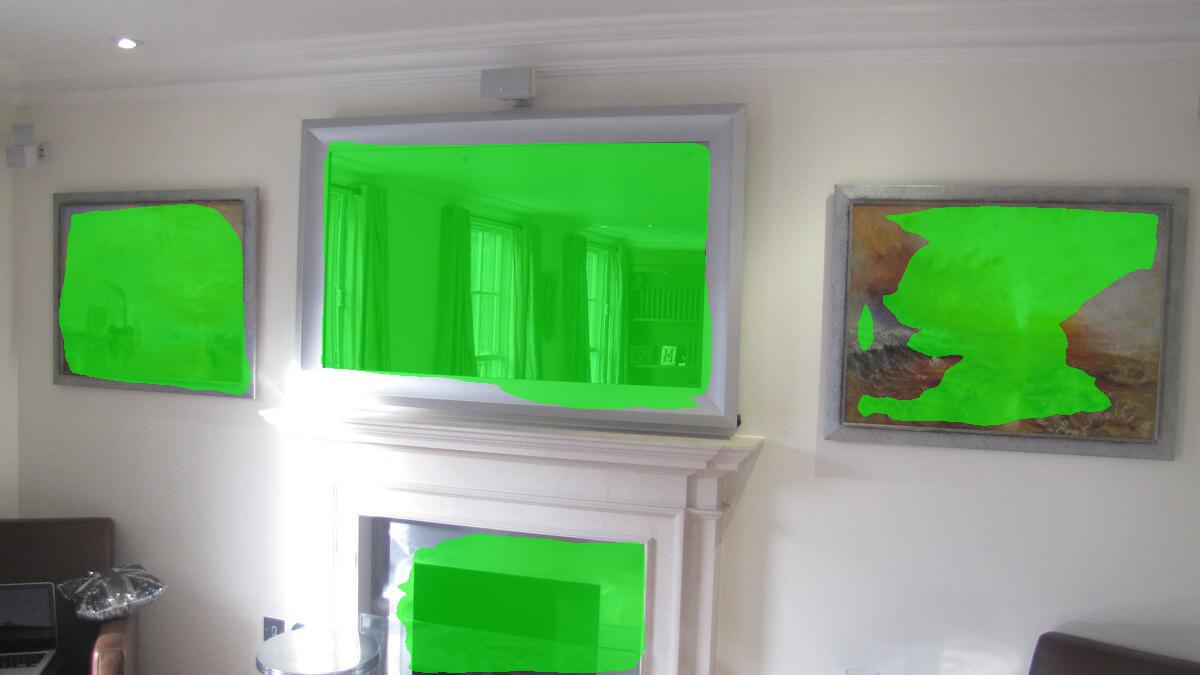}&
    \includegraphics[width=\wmore, height=\hmorethree]{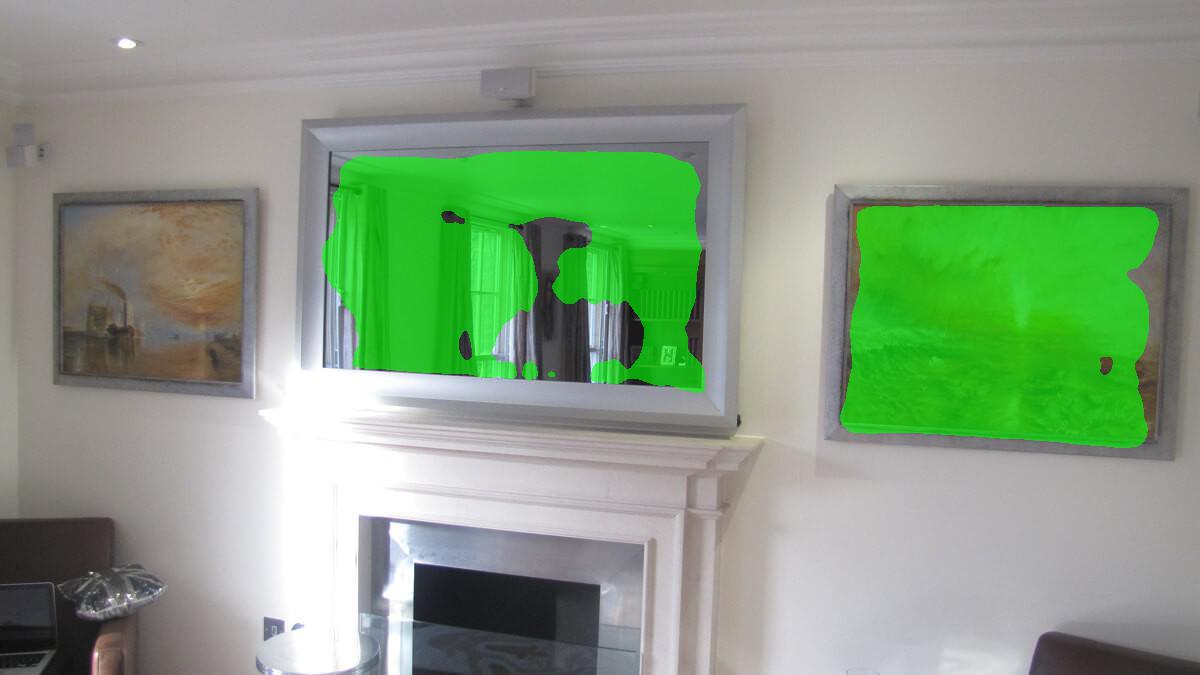}&
    \includegraphics[width=\wmore, height=\hmorethree]{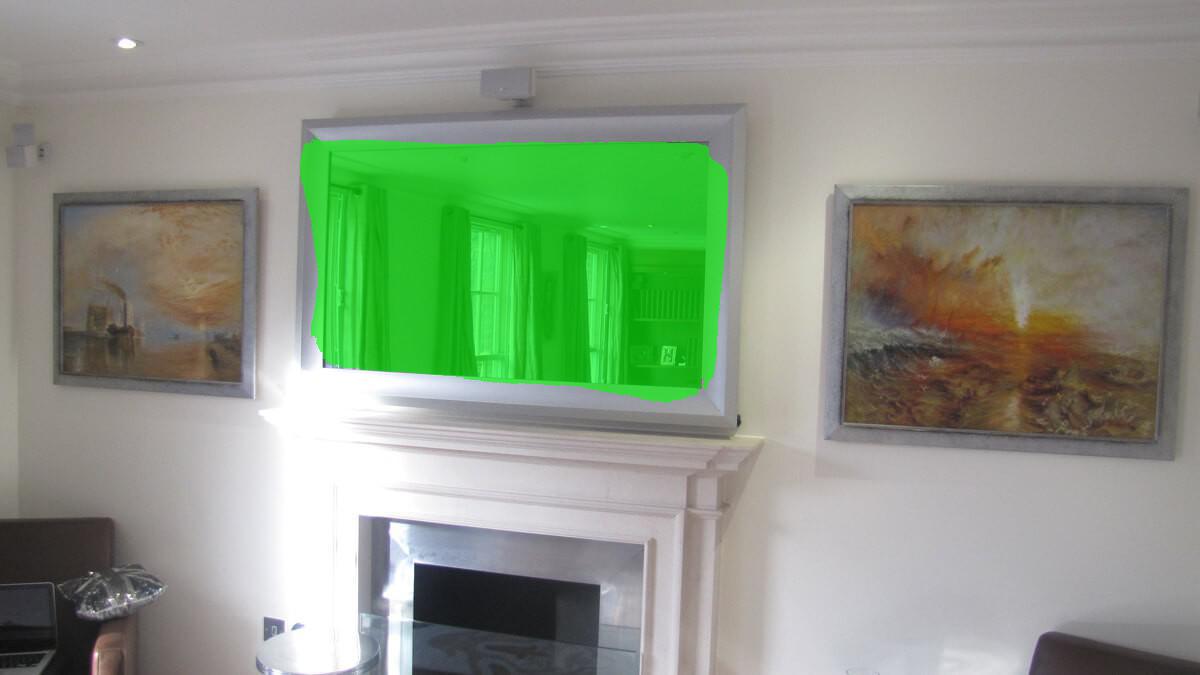} \\

    \includegraphics[width=\wmore, height=\hmorefour]{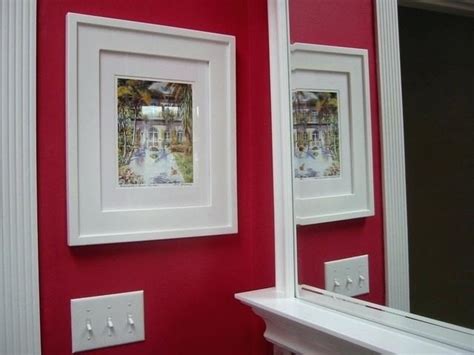}&
    \includegraphics[width=\wmore, height=\hmorefour]{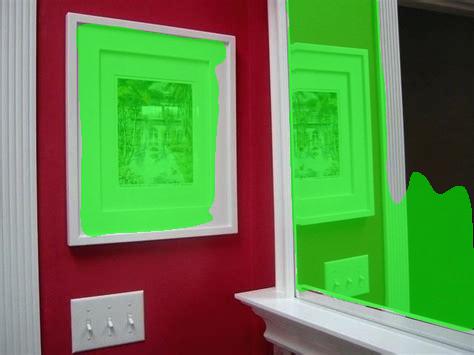}&
    \includegraphics[width=\wmore, height=\hmorefour]{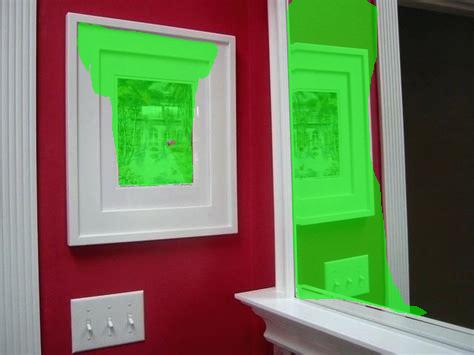}&
    \includegraphics[width=\wmore, height=\hmorefour]{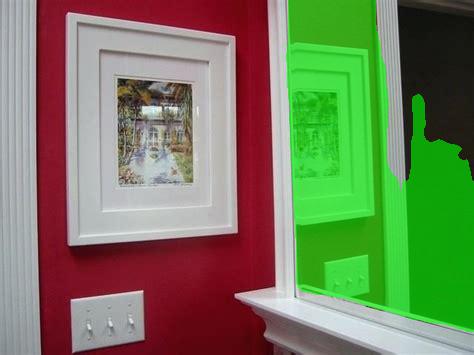}&
    \includegraphics[width=\wmore, height=\hmorefour]{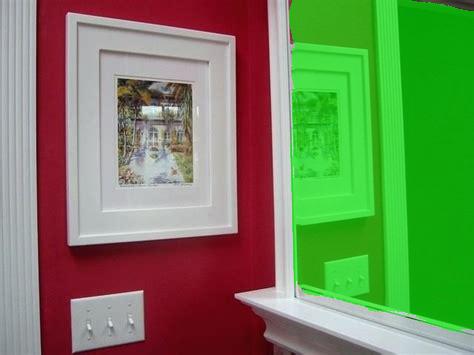} \\

    \includegraphics[width=\wmore, height=\hmorefive]{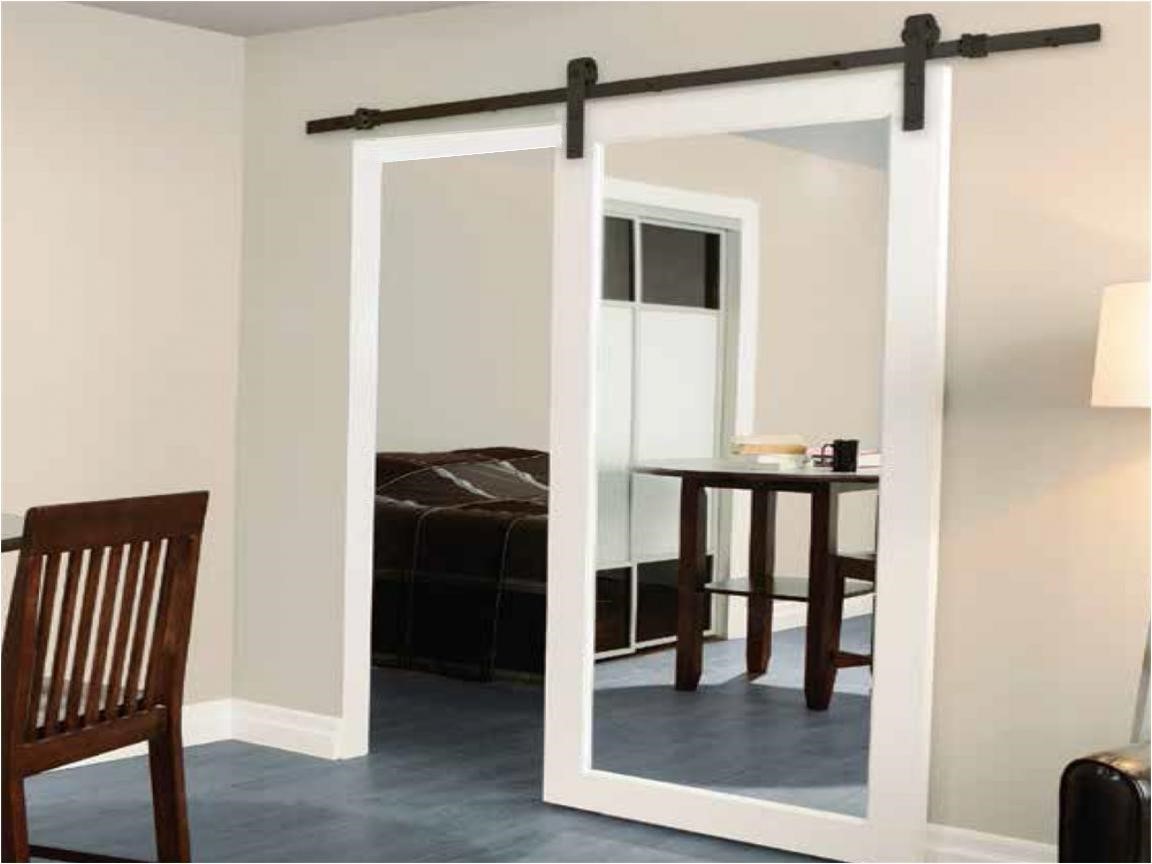}&
    \includegraphics[width=\wmore, height=\hmorefive]{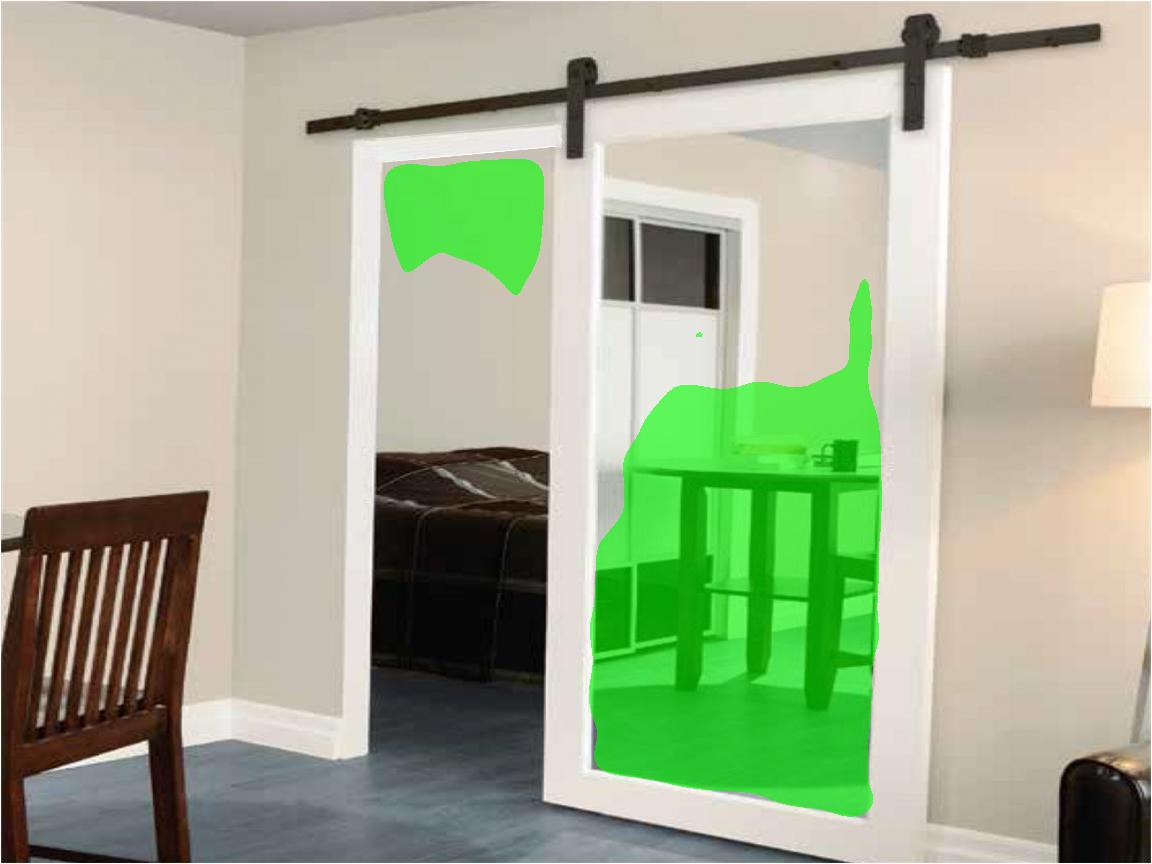}&
    \includegraphics[width=\wmore, height=\hmorefive]{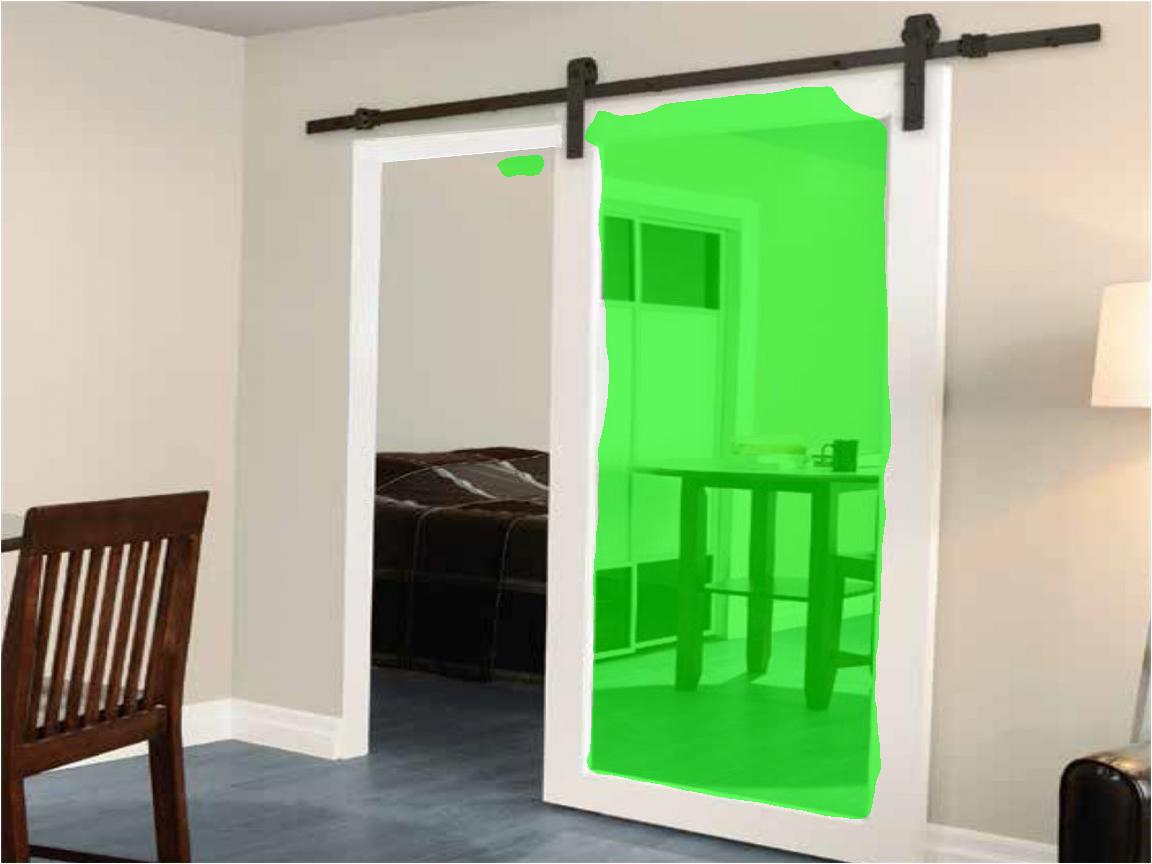}&
    \includegraphics[width=\wmore, height=\hmorefive]{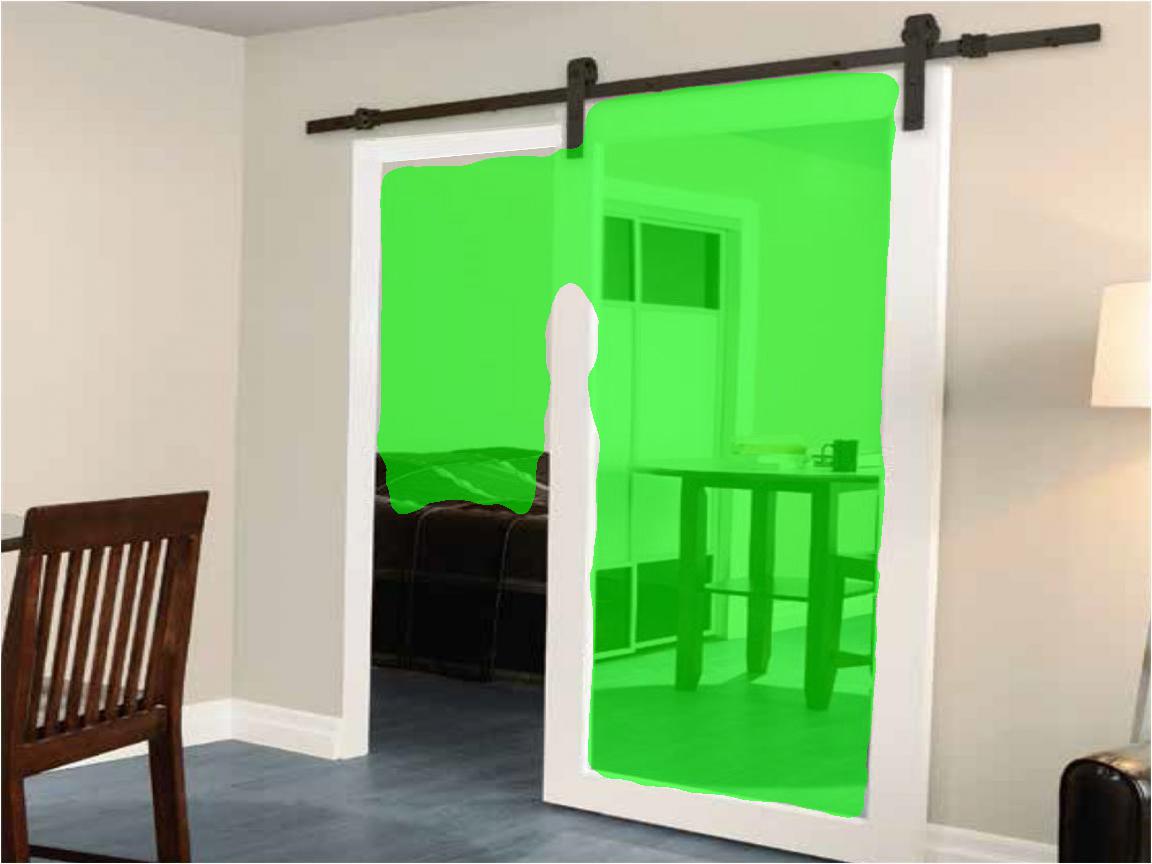}&
    \includegraphics[width=\wmore, height=\hmorefive]{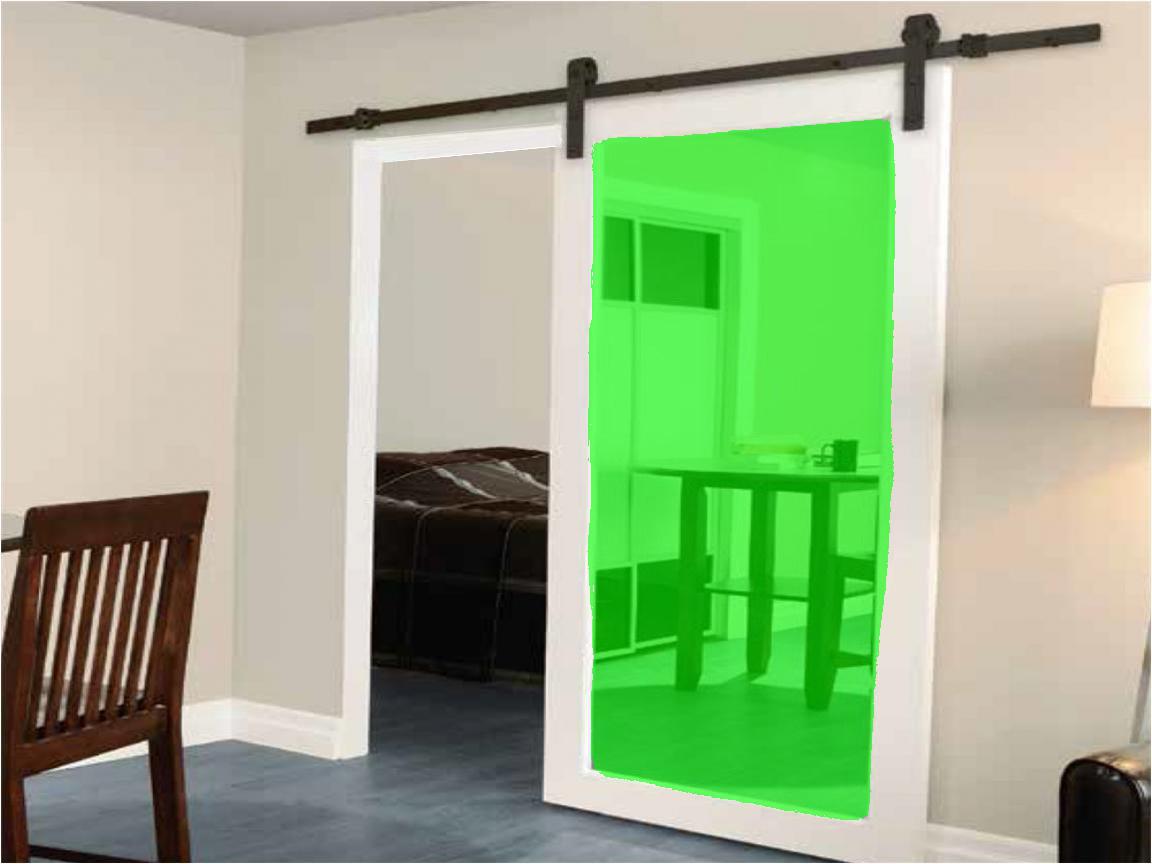} \\

    \includegraphics[width=\wmore, height=\hmoresix]{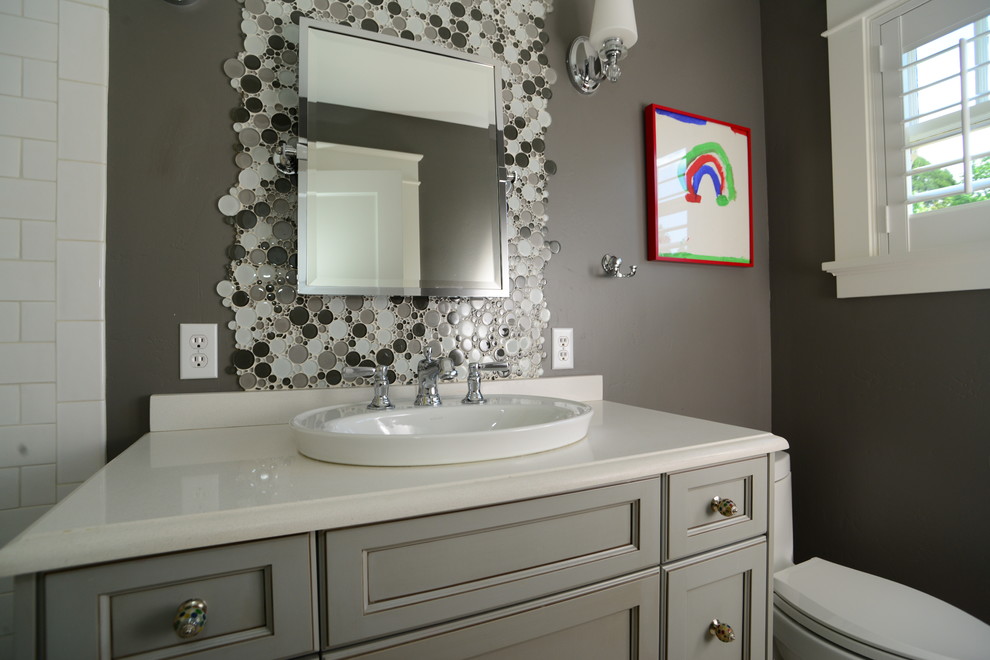}&
    \includegraphics[width=\wmore, height=\hmoresix]{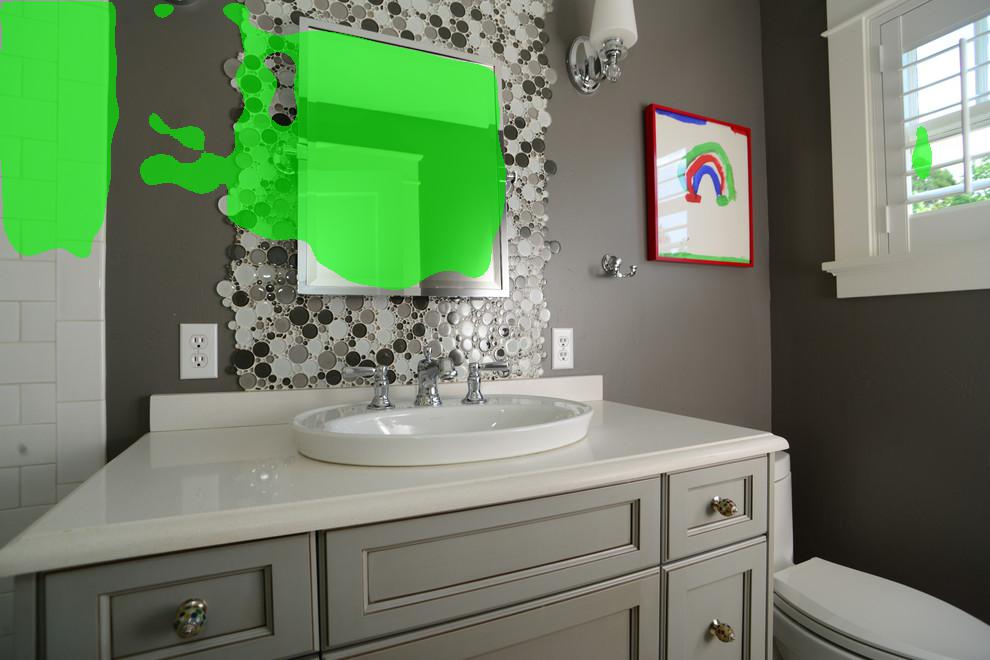}&
    \includegraphics[width=\wmore, height=\hmoresix]{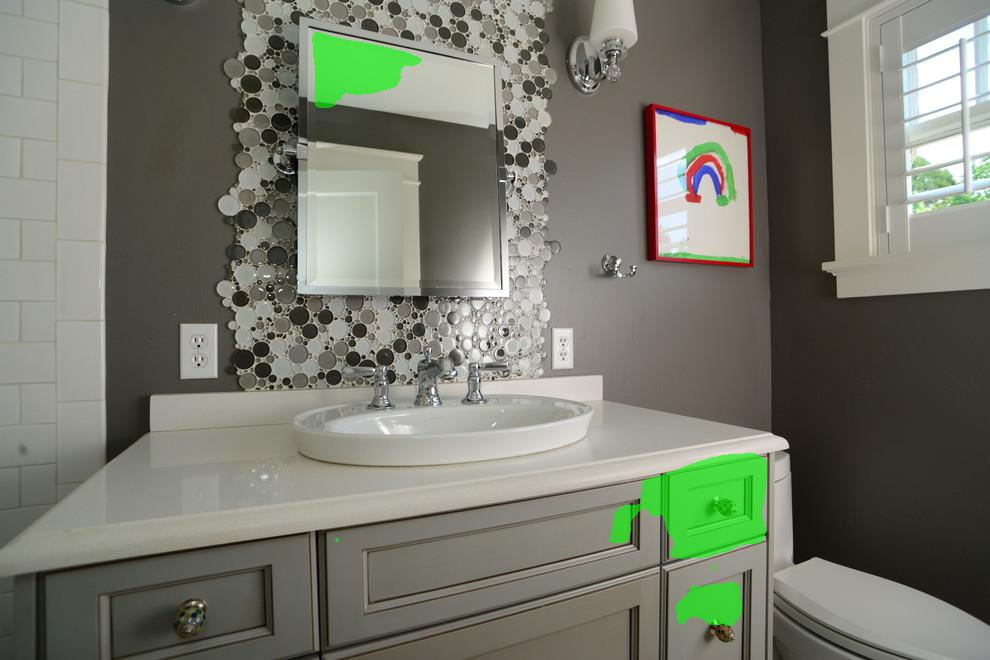}&
    \includegraphics[width=\wmore, height=\hmoresix]{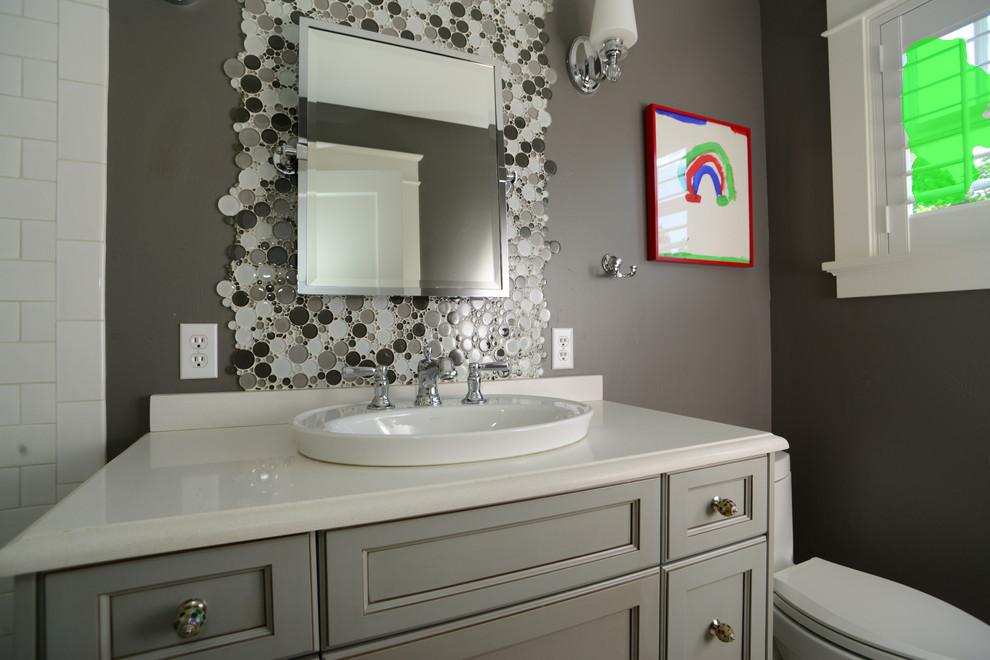}&
    \includegraphics[width=\wmore, height=\hmoresix]{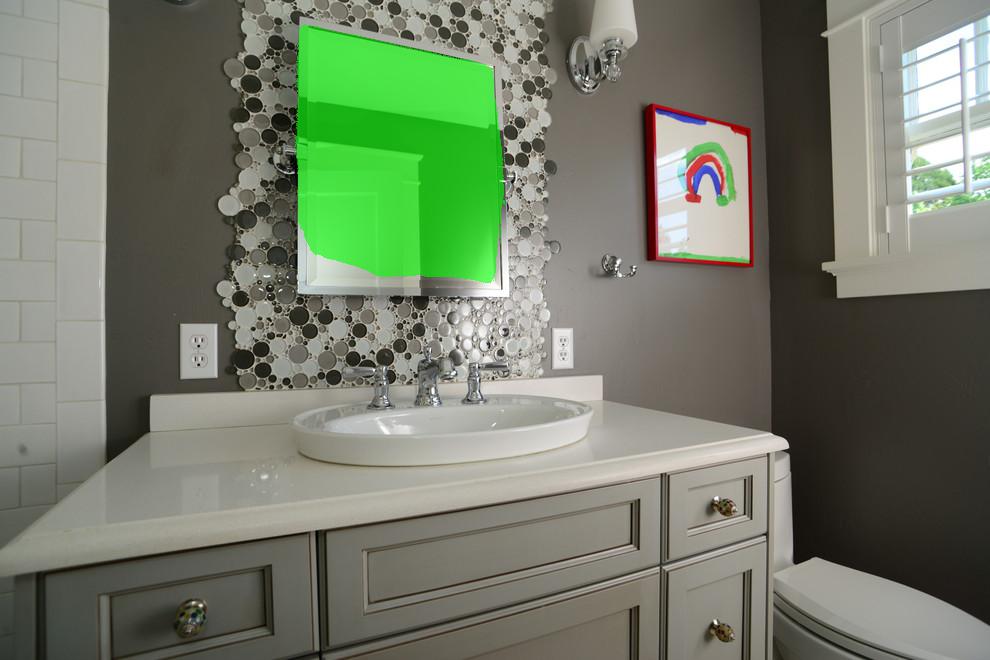} \\

    \includegraphics[width=\wmore, height=\hmoreseven]{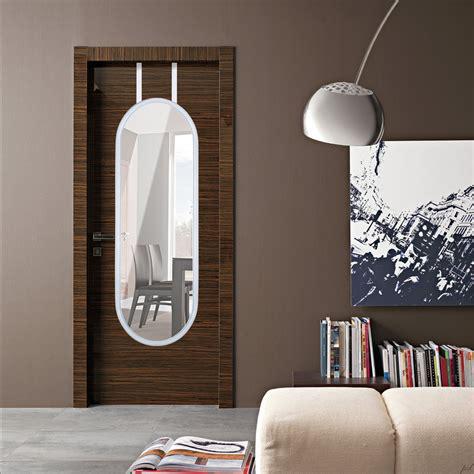}&
    \includegraphics[width=\wmore, height=\hmoreseven]{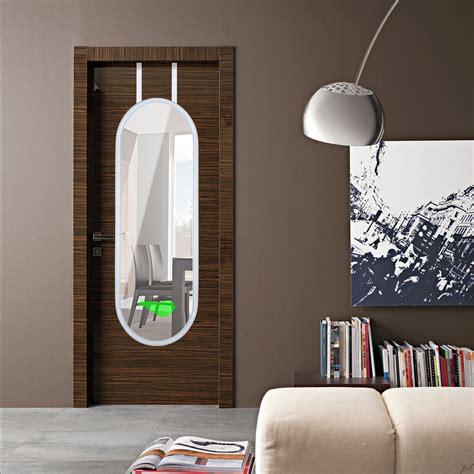}&
    \includegraphics[width=\wmore, height=\hmoreseven]{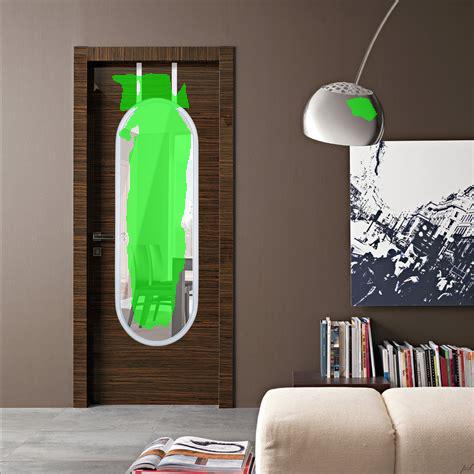}&
    \includegraphics[width=\wmore, height=\hmoreseven]{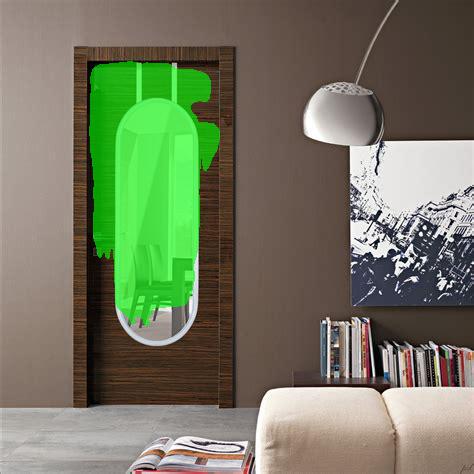}&
    \includegraphics[width=\wmore, height=\hmoreseven]{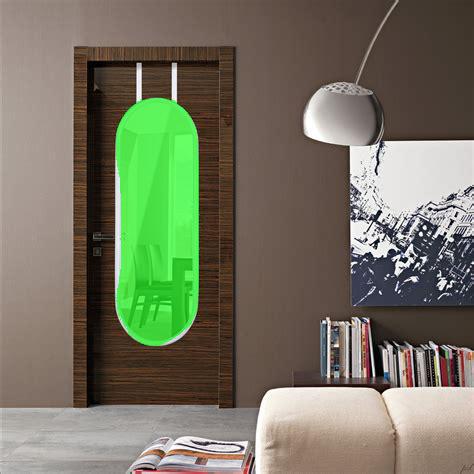} \\

    \footnotesize{Image} & \footnotesize{PSPNet\cite{zhao2017pyramid}} & \footnotesize{DSC\cite{Hu_2018_CVPR}} &
    \footnotesize{PiCANet\cite{liu2018picanet}} & \footnotesize{MirrorNet} \\\vspace{-8mm}
    \end{tabular}
  \caption{More mirror segmentation results on challenging images obtained from the Internet.}
  \label{fig:more}\vspace{-4mm}
\end{figure}

\subsection{Component Analysis}
Table~\ref{tab:ablation} demonstrates the effectiveness of the lov\'{a}sz-hinge loss \cite{Berman_2018_CVPR} and the proposed CCFE module. We can see that the lov\'{a}sz-hinge loss \cite{Berman_2018_CVPR} performs better than the binary cross entropy (BCE) loss in our task, due to its scale-invariant property. In addition, while multi-scale dilated convolutions (\ie, ``CCFE w/o contrasts'') benefit the segmentation performance, we can see that using only one CCFE block with 4 parallel scales of contrast extraction (``basic + CCFE w/ 1B4C'') can improve both the pixel accuracy and BER. In contrast, using four CCFE blocks with one single scale of contrast extraction mainly improves the IoU. Our proposed multi-scale contextual contrasted feature learning takes advantage of both. Figure~\ref{fig:internal} shows a visual example, in which we can see that our method successfully learns the global contextual contrasted features for addressing the mirror under-segmentation problem.

\section{Conclusion and Future Work}
In this paper, we have presented a novel method to segment mirrors from an input image. Specifically, we have constructed the first large-scale mirror dataset (MSD). It contains 4,018 images with mirrors and corresponding masks. We have also proposed a novel network to leverage multi-scale contextual contrasts for mirror detection. We have conducted extensive experiments to verify the superiority of the proposed network against state-of-the-art methods developed for other relevant problems, on both the proposed MSD test set, the ADE20K dataset~\cite{Zhou_2017_CVPR}, and some challenging images obtained from the Internet.

\def\winternal{0.22\linewidth}
\def\hinternal{.68in}
\begin{figure}[tbp]
\setlength{\tabcolsep}{2pt}
  \centering
  \begin{tabular}{cccc}
    \includegraphics[width=\winternal, height=\hinternal]{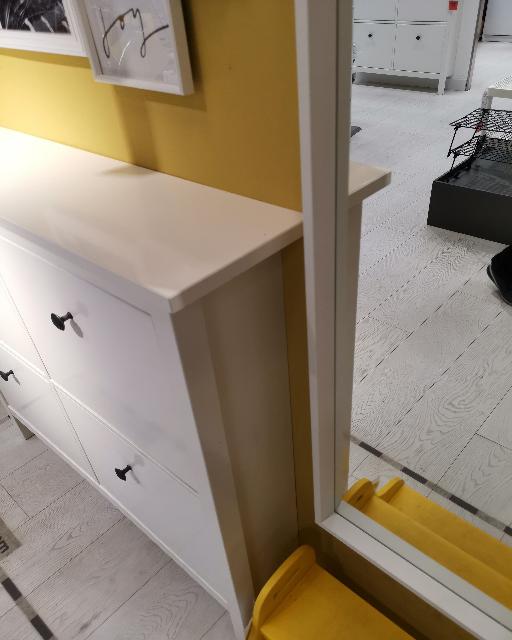}&
    \includegraphics[width=\winternal, height=\hinternal]{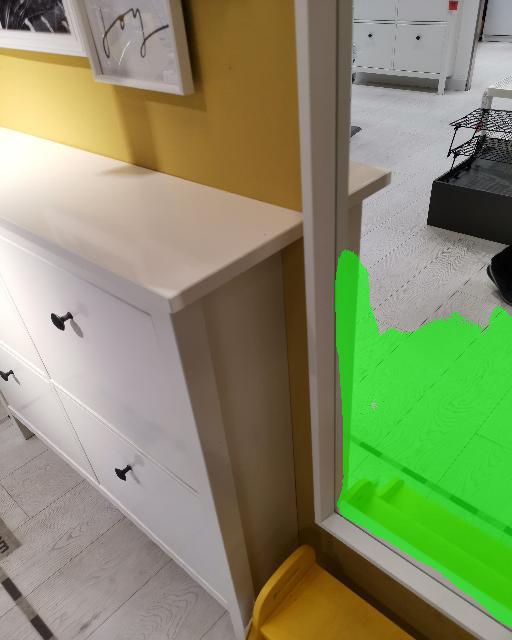}&
    \includegraphics[width=\winternal, height=\hinternal]{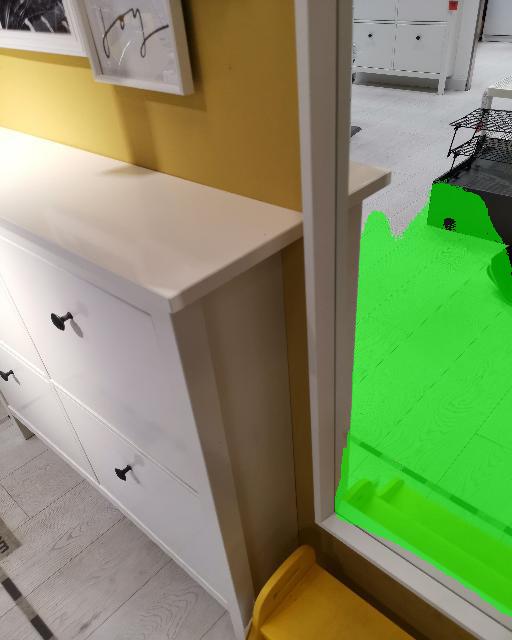}&
    \includegraphics[width=\winternal, height=\hinternal]{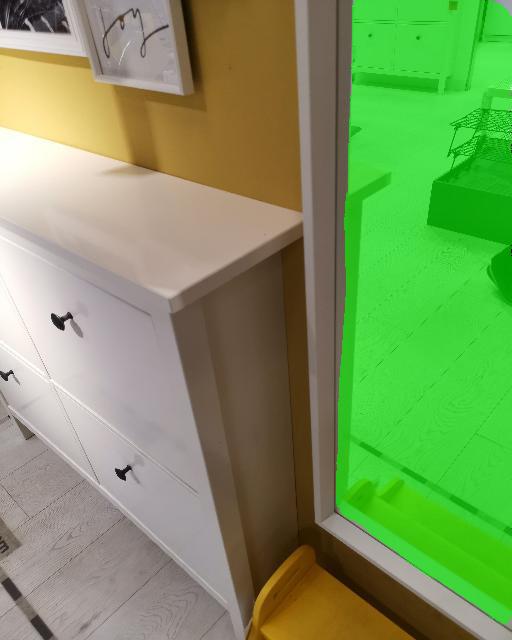} \\
    \small{image} & \small{basic} & \tabincell{c}{\small{basic + CCFE}\\\small{w/o contrasts}} &
    \tabincell{c}{\small{basic + CCFE}\\\small{w/ \rf{contrasts}}} \\\vspace{-8mm}

    \end{tabular}
  \caption{Visual example of the component analysis.}
  \label{fig:internal}\vspace{-3mm}
\end{figure}

\def\wfailure{0.43\linewidth}
\def\hfailure{.6in}
\begin{figure}
\setlength{\tabcolsep}{2pt}
  \centering
  \begin{tabular}{cc}

    \includegraphics[width=\wfailure, height=\hfailure]{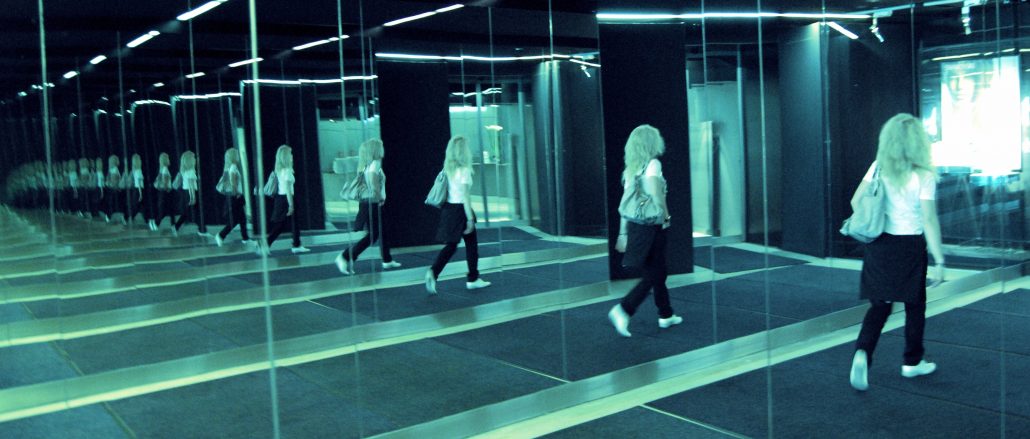}&
    \includegraphics[width=\wfailure, height=\hfailure]{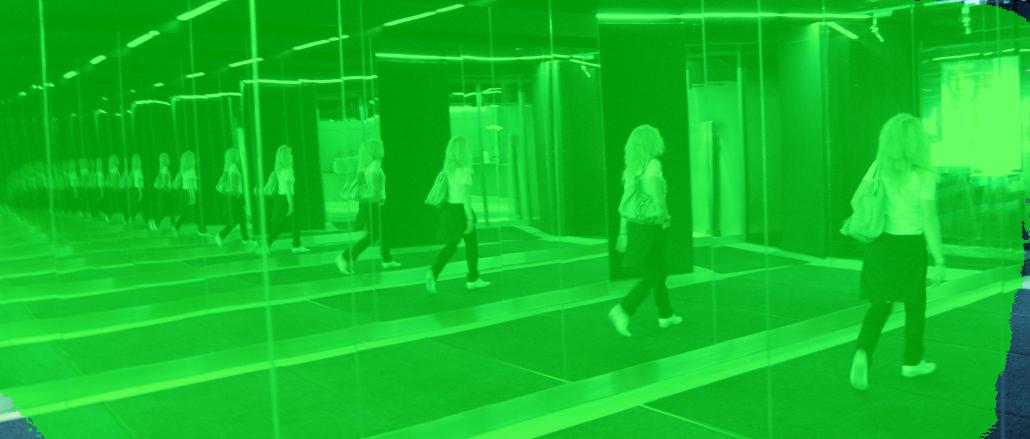} \\

    \includegraphics[width=\wfailure, height=\hfailure]{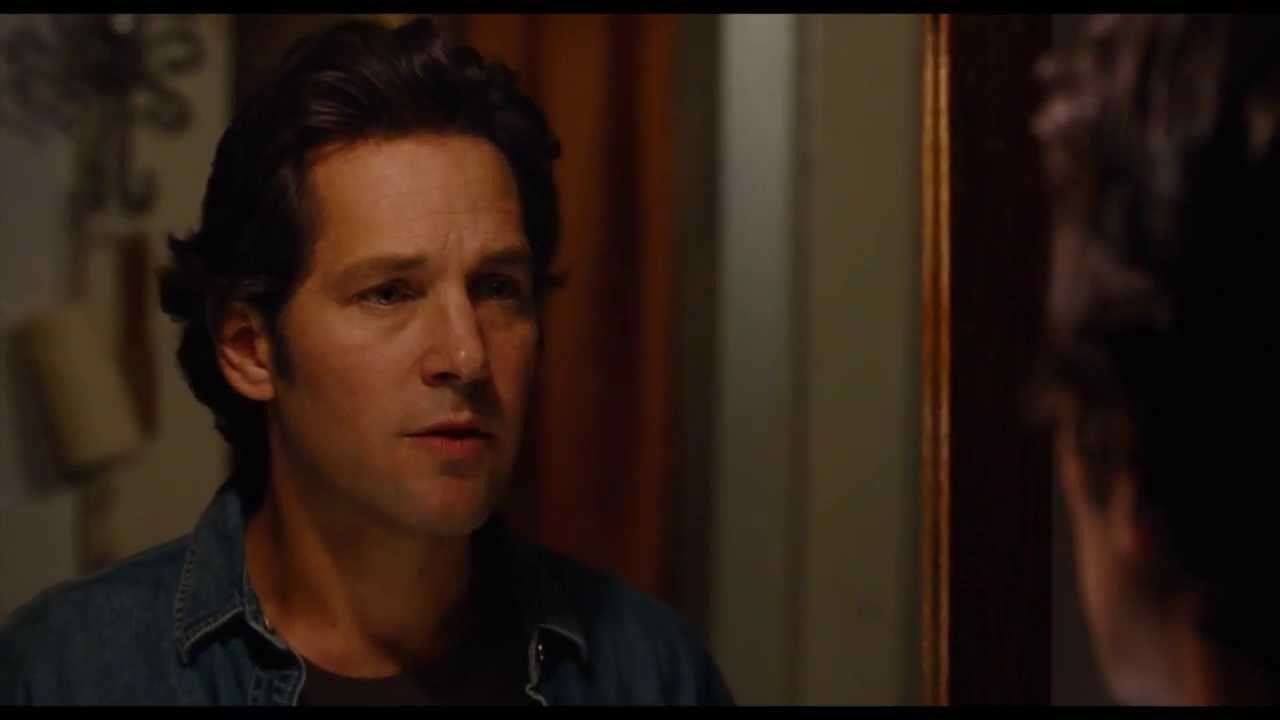}&
    \includegraphics[width=\wfailure, height=\hfailure]{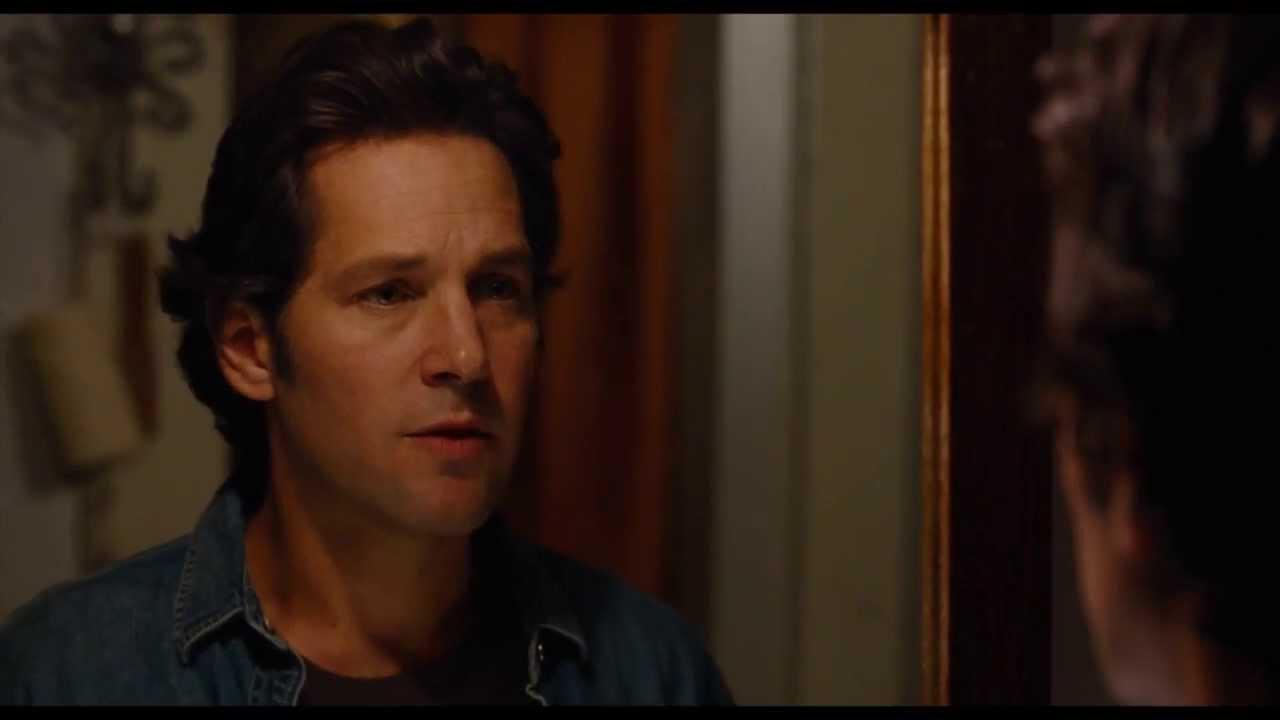} \\

    \small{input images} & \small{our results} \\\vspace{-6mm}

    \end{tabular}
  \caption{Failure cases. Our mirror segmentation method can fail in extreme scenarios where insufficient contextual contrasts can be extracted.}
  \label{fig:failure}\vspace{-2.5mm}
\end{figure}

Our method does have limitations. As it relies on modeling the contextual contrasts presented in the input images, it tends to fail in some extreme scenes where insufficient contextual contrasts between the mirrors and their surroundings can be perceived, as shown in Figure~\ref{fig:failure}.

As a first attempt to address the automatic mirror segmentation problem, we focus in this paper on segmenting mirrors that appear in our daily life scenes. However, in some cities, the glass walls of skyscrapers may often exhibit mirror-like effects and reflect the surrounding objects/scenes. There are also very large mirrors that may appear outside some stores. As a future work, we are interested to extend our method to detect this kind of mirrors that appear in city streets, which may benefit outdoor vision tasks such as autonomous driving and drone navigation.

\textbf{Acknowledgements.} This work was supported in part by the NNSF of China under Grants 91748104, U1811463, 61632006, 61425002, and 61751203, by the Open Project Program of the State Key Lab of CAD\&CG (Grant A1901), Zhejiang University, the Open Research Fund of Beijing Key Laboratory of Big Data Technology for Food Safety (No. BTBD-2018KF), and a SRG grant from City University of Hong Kong (Ref: 7004889).

{\small
\bibliographystyle{ieee_fullname}
\bibliography{paper}
}

\end{document}